\documentclass[10pt,twocolumn,letterpaper]{article}

\usepackage[pagenumbers]{cvpr}      %

\makeatletter
\@namedef{ver@everyshi.sty}{}
\makeatother
\usepackage{tikz}
\usepackage{pgfplots}

\definecolor{cvprblue}{rgb}{0.21,0.49,0.74}
\definecolor{lightgray2}{rgb}{0.88,0.88,0.88}

\usepackage[pagebackref,breaklinks,colorlinks,allcolors=cvprblue]{hyperref}

\usepackage{csquotes}
\usepackage{colortbl}
\usepackage{xcolor}
\usepackage[export]{adjustbox}

\usepackage{amsmath,amsfonts,bm}

\def\eqref#1{equation~\ref{#1}}

\def\1{\bm{1}}

\DeclareMathAlphabet{\mathsfit}{\encodingdefault}{\sfdefault}{m}{sl}
\SetMathAlphabet{\mathsfit}{bold}{\encodingdefault}{\sfdefault}{bx}{n}

\newcommand{\bx}{\mathbf{x}}

\newcommand{\bz}{\mathbf{z}}

\newcommand{\bepsilon}{{\boldsymbol{\epsilon}}}

\usepackage{amsmath,amsfonts,bm,amssymb,mathtools,amsthm}

\usepackage{algorithm}
\usepackage{algcompatible}

\usepackage{booktabs}
\usepackage{multirow}
\usepackage{makecell}
\usepackage{pifont}
\usepackage{abstract}

\usepackage{graphicx}
\usepackage{tikz}
\usepackage{microtype}

\newcommand{\trf}[1]{{\textbf{\color{red}{#1}}}} %
\newcommand{\tbd}[1]{{\color{blue}{\underline{#1}}}} %

\definecolor{tabfirst}{RGB}{250, 210, 207} %
\definecolor{tabsecond}{RGB}{206, 234, 214} %
\definecolor{tabthird}{RGB}{210, 227, 252} %

\newcommand{\textshape}[2]{%
\linespread{0.1}
\begin{tikzpicture}
  \node[inner sep=0pt,outer sep=0pt] (outer) {%
    \begin{tikzpicture}  %
      \node[shape=rectangle, minimum width=#1, minimum height=#1/4*3, inner sep=1pt, align=center, text width=#1-8pt, draw, fill=white] (text) {%
        \fontsize{6}{7.2}\selectfont #2 
      };
      \draw[gray] (text.south west) ++(2pt,2pt) rectangle (text.north east) ++(-2pt,-2pt); %
    \end{tikzpicture}
  };
\end{tikzpicture}
}

\newcommand{\redrectangle}[5]{%
  \begin{tikzpicture}
    \node[inner sep=0pt] (image) at (0,0) {\includegraphics[width=#2]{#1}};
    \node[shape=rectangle, minimum width=2cm, minimum height=1cm, draw=red, line width=2pt, fill=none, anchor=north west] (square) at ([xshift=#4*#2,yshift=#5*#2]image.north west) {}; %
  \end{tikzpicture}
}

\newcommand{\redrectanglee}[5]{%
  \begin{tikzpicture}
    \node[inner sep=0pt] (image) at (0,0) {\includegraphics[width=#2]{#1}};
    \node[shape=rectangle, minimum width=0.8cm, minimum height=0.6cm, draw=red, line width=2pt, fill=none, anchor=north west] (square) at ([xshift=#4*#2,yshift=#5*#2]image.north west) {}; %
  \end{tikzpicture}
}

\newcommand{\redrectangleee}[5]{%
  \begin{tikzpicture}
    \node[inner sep=0pt] (image) at (0,0) {\includegraphics[width=#2]{#1}};
    \node[shape=rectangle, minimum width=0.4cm, minimum height=0.3cm, draw=red, line width=2pt, fill=none, anchor=north west] (square) at ([xshift=#4*#2,yshift=#5*#2]image.north west) {}; %
  \end{tikzpicture}
}

\newcommand{\whiterectangleee}[5]{%
  \begin{tikzpicture}
    \node[inner sep=0pt] (image) at (0,0) {\includegraphics[width=#2]{#1}};
    \node[shape=rectangle, minimum width=0.4cm, minimum height=0.3cm, draw=white, line width=2pt, fill=none, anchor=north west] (square) at ([xshift=#4*#2,yshift=#5*#2]image.north west) {}; %
  \end{tikzpicture}
}

\def\paperID{24} %
\def\confName{CVPR}
\def\confYear{2025}

\title{The Power of Context: How Multimodality Improves Image Super-Resolution}

\author{Kangfu Mei\thanks{This work was done during an internship at Google.}$^{\,\,\,\,1,2}$, Hossein Talebi$^{1}$, Mojtaba Ardakani$^{1}$,\\ Vishal M. Patel$^{2}$, Peyman Milanfar$^{1}$,  Mauricio Delbracio$^{1}$\\
 $^{1}$ Google, $^{2}$ Johns Hopkins University \\
 Project Page: \url{https://mmsr.kfmei.com/}}

\newlength{\imagewidth}

\begin{document}
\twocolumn[{%
    \vspace{-1\baselineskip}
	\maketitle
	\renewcommand\twocolumn[1][]{#1}%
	\vspace{-4\baselineskip}
	\begin{center}
	\centering
    \settowidth{\imagewidth}{\includegraphics{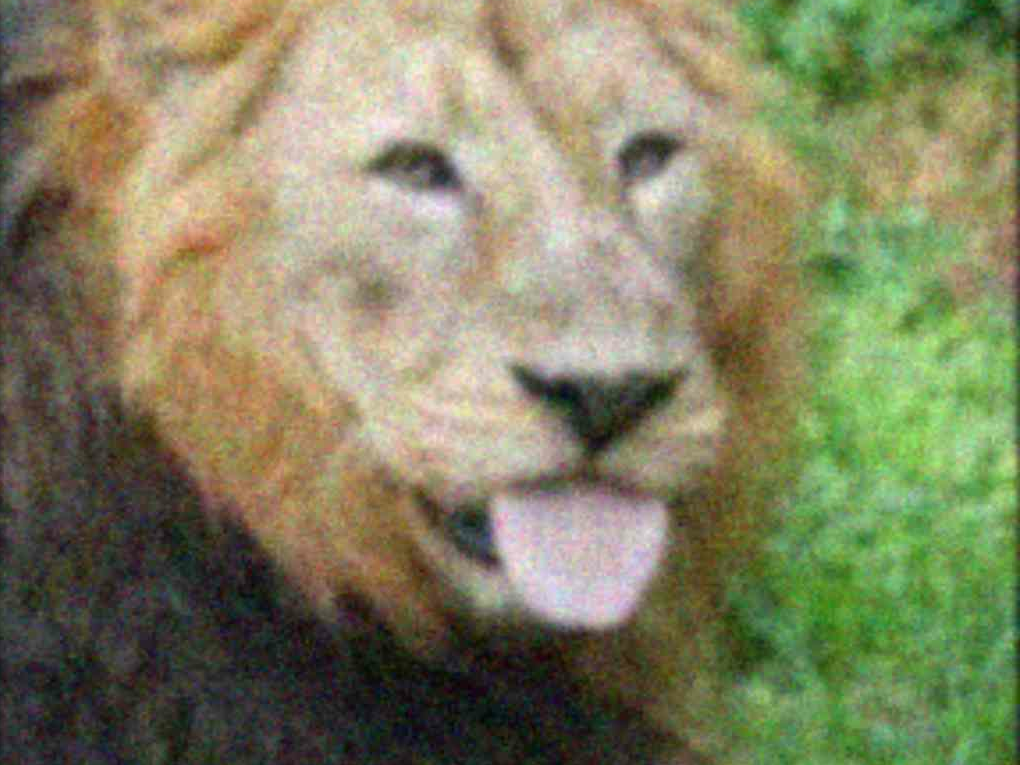}}
    \def\xwidth{0.13\linewidth}
    \def\ywidth{0.12\linewidth}
    \setlength{\tabcolsep}{1pt}
    \renewcommand\arraystretch{0.8}
    \begin{tabular}[t]{c c c c c c c c c}\\
    \multicolumn{3}{c}{\cellcolor{tabfirst} \scriptsize Inputs} & & \multicolumn{3}{c}{\cellcolor{tabsecond} \scriptsize Outputs} & & \cellcolor{tabthird} \scriptsize Reference \\
    \cellcolor{tabfirst} \redrectanglee{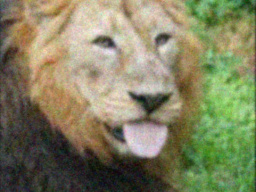}{\xwidth}{0.3cm}{0.4}{-0.4} &
    \cellcolor{tabfirst}
    \includegraphics[width=\xwidth, clip=true, trim = 0.45\imagewidth{} 0.22\imagewidth{} 0.25\imagewidth{} 0.558\imagewidth{}]{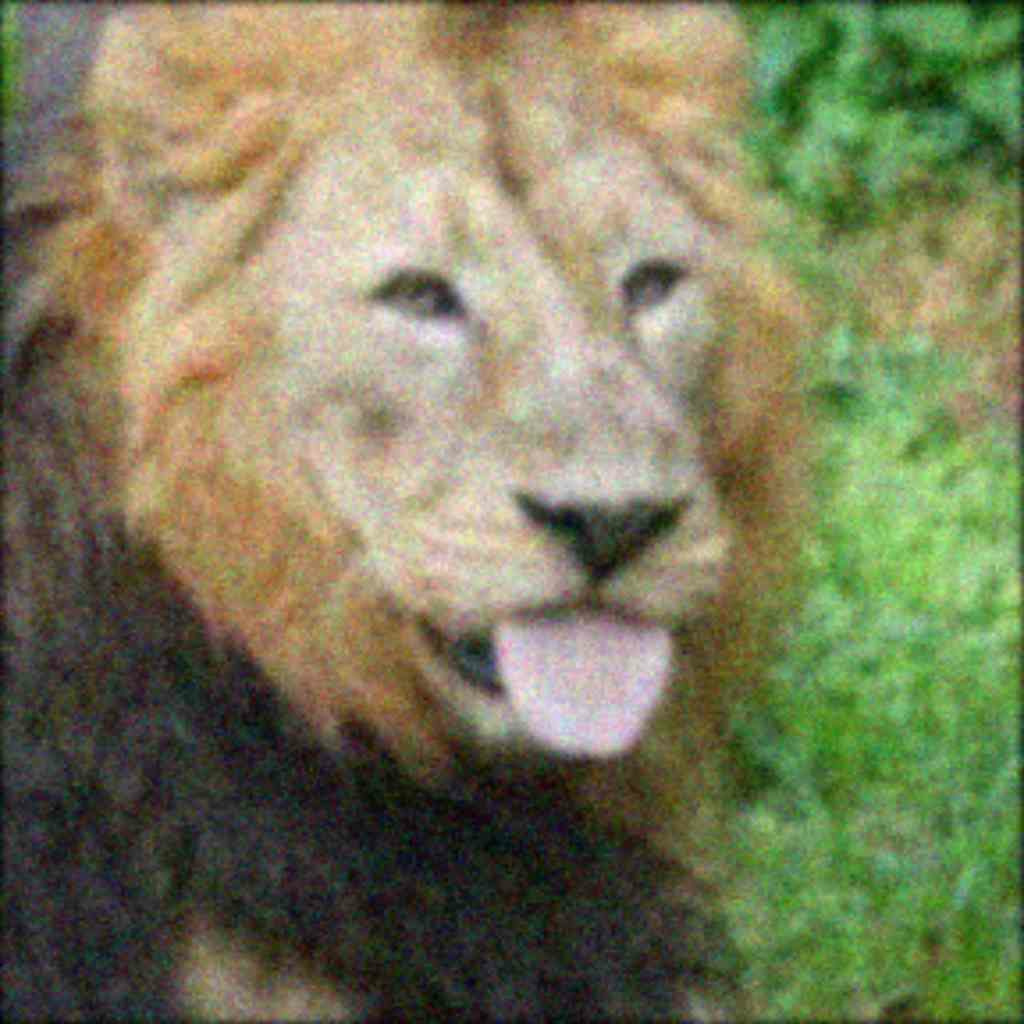} &
    \cellcolor{tabfirst} \textshape{\xwidth}{A \textbf{close-up} of a male \textcolor{pink}{\textbf{lion}} with a dark mane, light tan face, and pink tongue \textbf{sticking out} \dots} & &
    \cellcolor{tabsecond} \redrectanglee{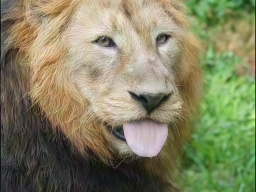}{\xwidth}{0.3cm}{0.4}{-0.4} &
    \cellcolor{tabsecond} \redrectanglee{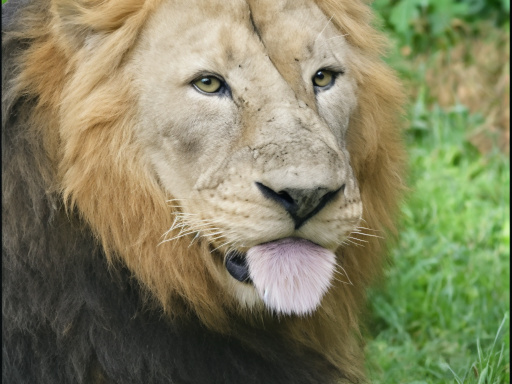}{\xwidth}{0.3cm}{0.4}{-0.4} &
    \cellcolor{tabsecond} \redrectanglee{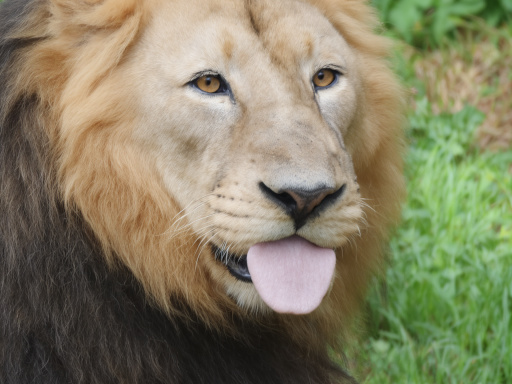}{\xwidth}{0.3cm}{0.4}{-0.4} & &
    \cellcolor{tabthird} \redrectanglee{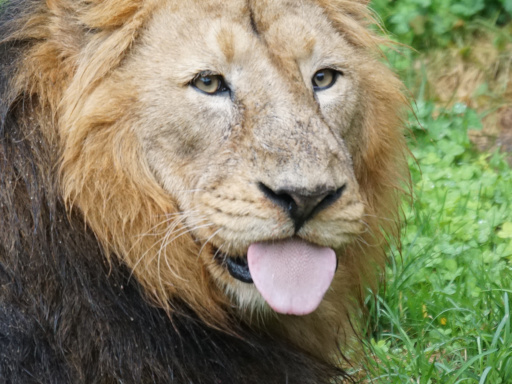}{\xwidth}{0.3cm}{0.4}{-0.4} \\
    \cellcolor{tabfirst} \scriptsize LR & \cellcolor{tabfirst} \scriptsize LR (Zoomed) & \cellcolor{tabfirst} \scriptsize Caption & & \cellcolor{tabsecond} \scriptsize PASD & \cellcolor{tabsecond} \scriptsize SeeSR  &  \cellcolor{tabsecond} \scriptsize MMSR (Ours) & & \cellcolor{tabthird} \scriptsize HR \\
    \cellcolor{tabfirst} \redrectanglee{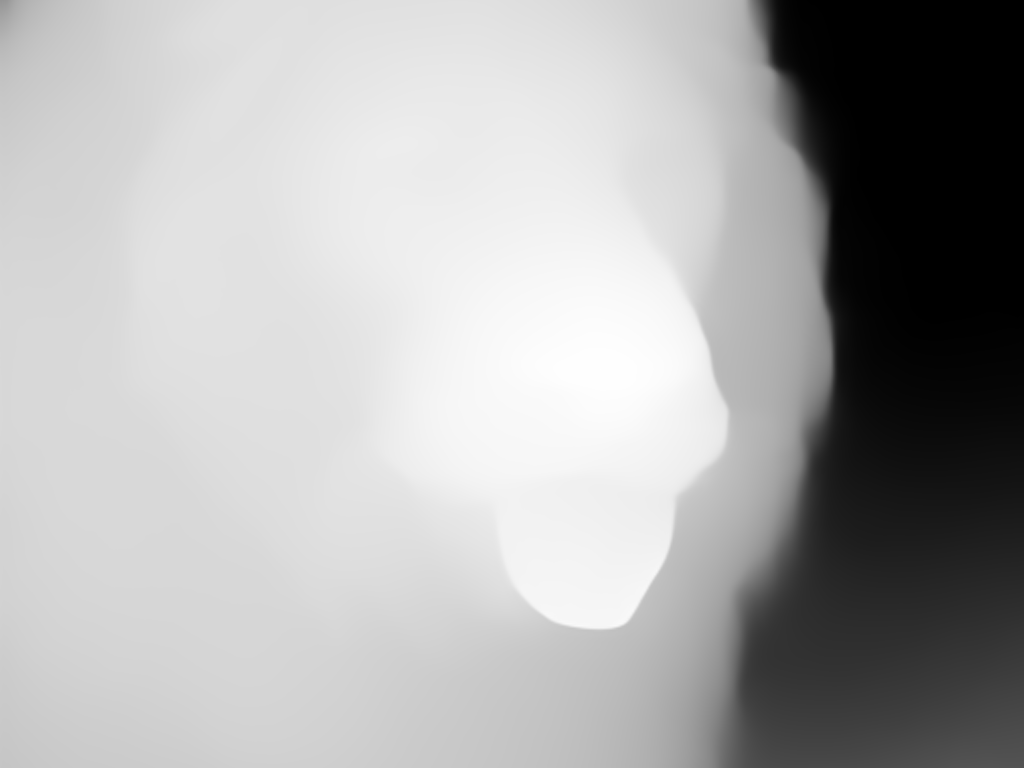}{\xwidth}{0.3cm}{0.4}{-0.4} &
    \cellcolor{tabfirst} \redrectanglee{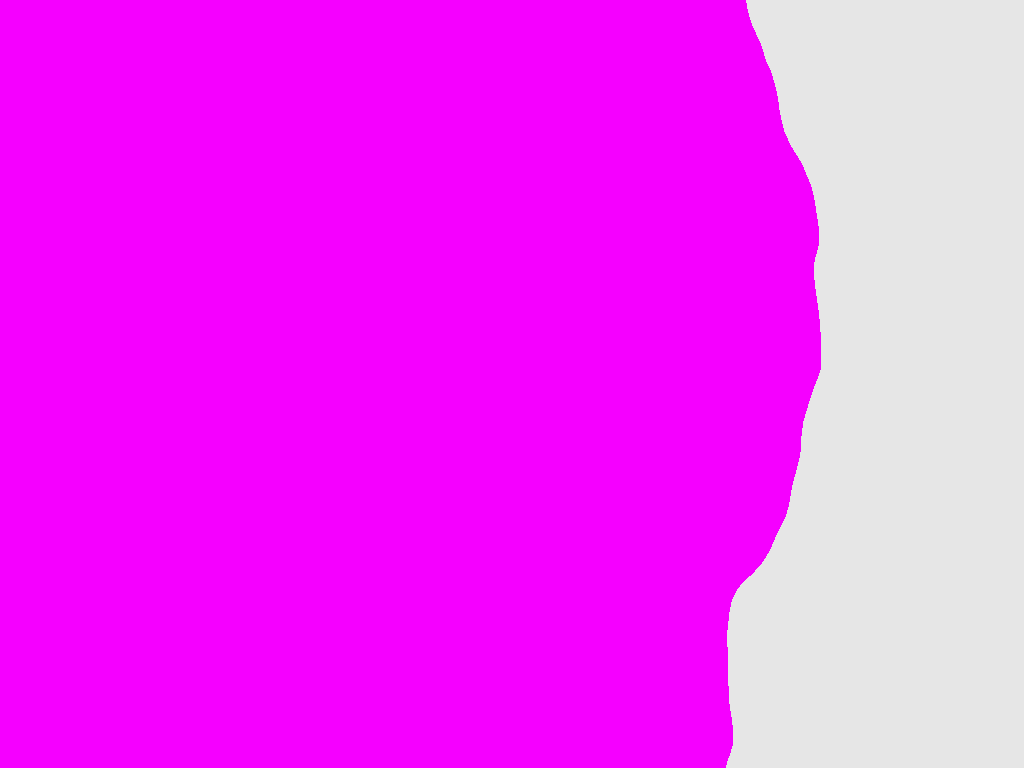}{\xwidth}{0.3cm}{0.4}{-0.4} &
    \cellcolor{tabfirst} \redrectanglee{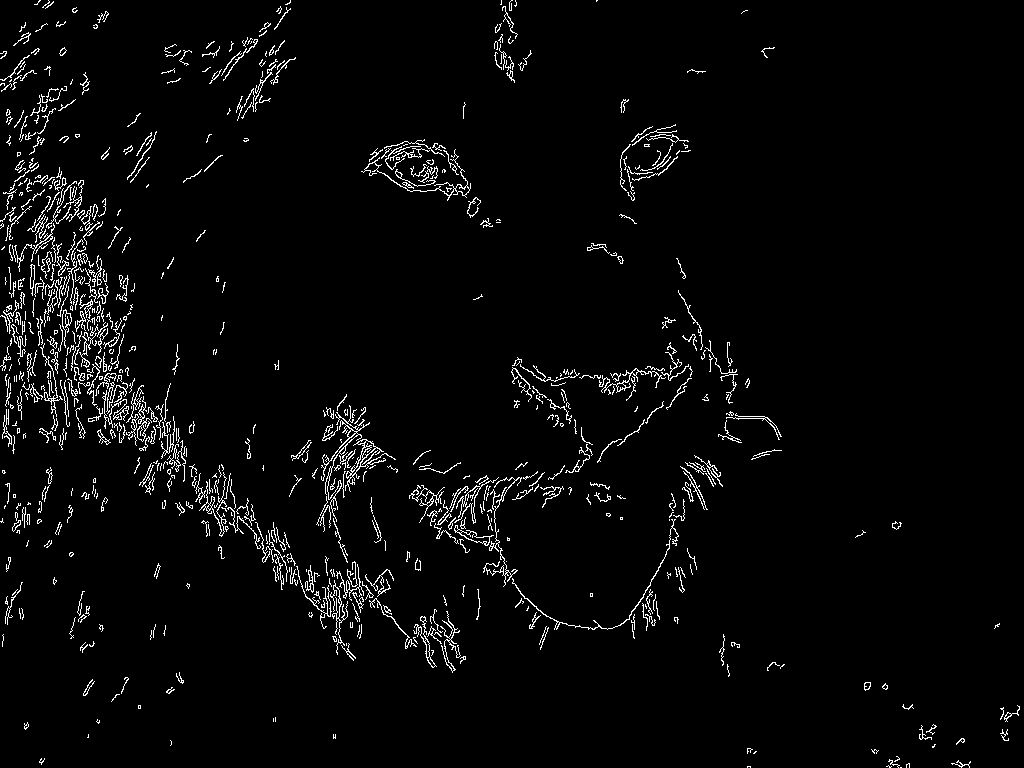}{\xwidth}{0.3cm}{0.4}{-0.4} & & 
    \cellcolor{tabsecond}
    \includegraphics[width=\xwidth, clip=true, trim = 0.45\imagewidth{} 0.22\imagewidth{} 0.25\imagewidth{} 0.558\imagewidth{}]{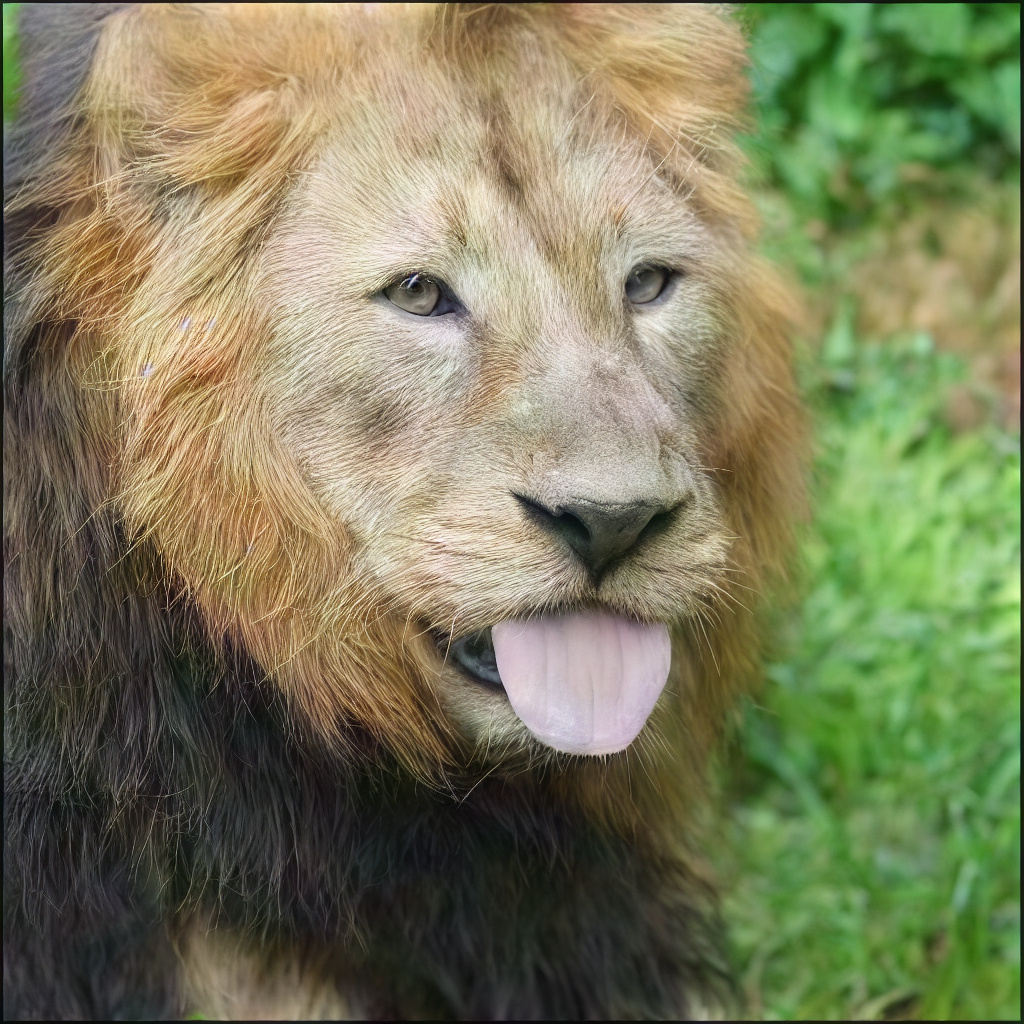} &
    \cellcolor{tabsecond}  
    \includegraphics[width=\xwidth, clip=true, trim = 0.45\imagewidth{} 0.22\imagewidth{} 0.25\imagewidth{} 0.558\imagewidth{}]{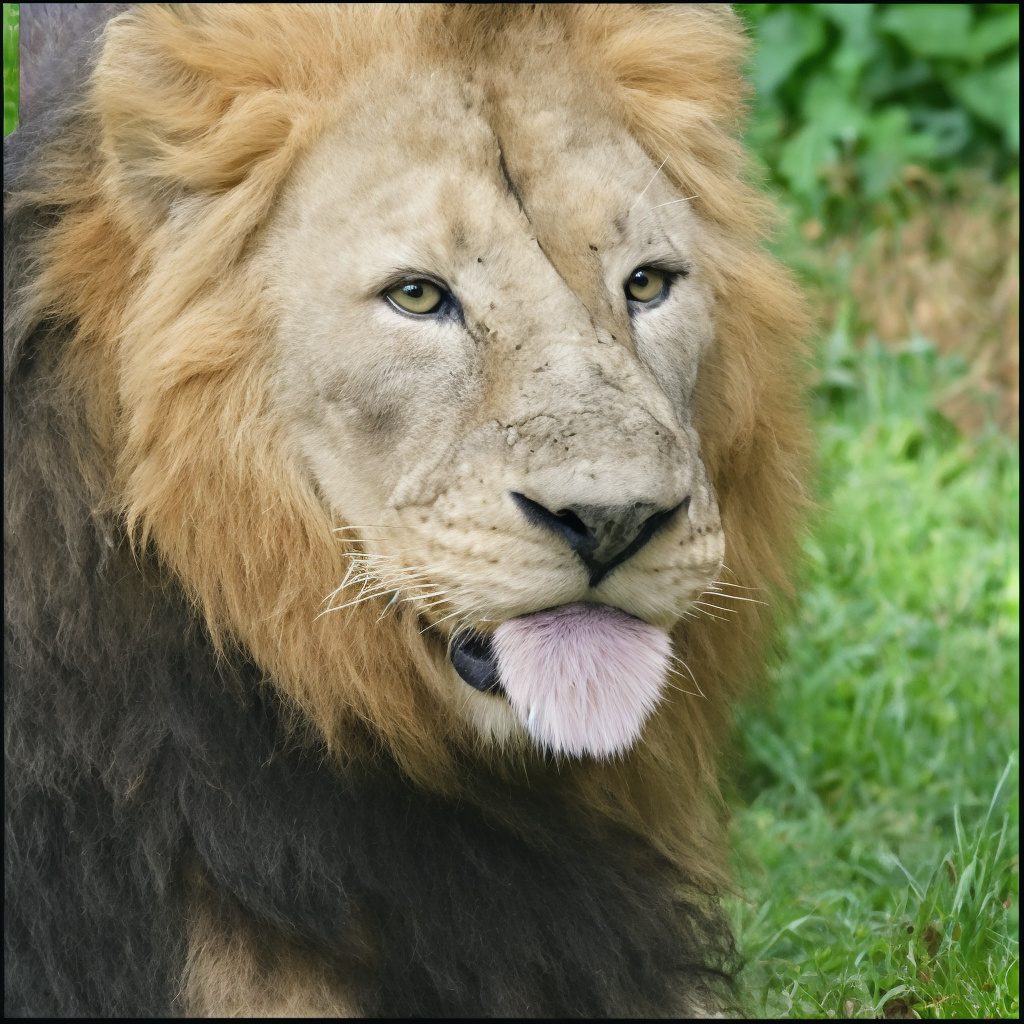} &
    \cellcolor{tabsecond}
    \includegraphics[width=\xwidth, clip=true, trim = 0.45\imagewidth{} 0.22\imagewidth{} 0.25\imagewidth{} 0.558\imagewidth{}]{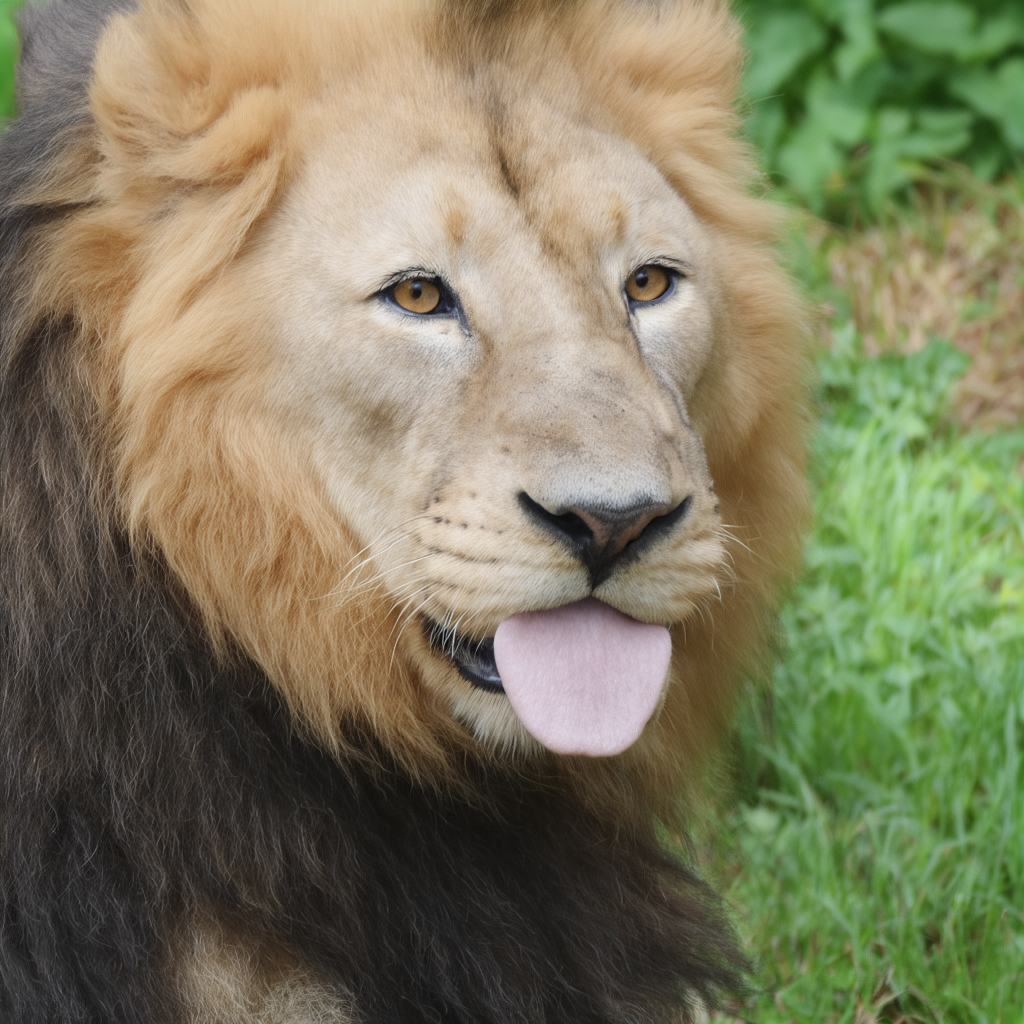} & &
    \cellcolor{tabthird}  
    \includegraphics[width=\xwidth, clip=true, trim = 0.45\imagewidth{} 0.22\imagewidth{} 0.25\imagewidth{} 0.558\imagewidth{}]{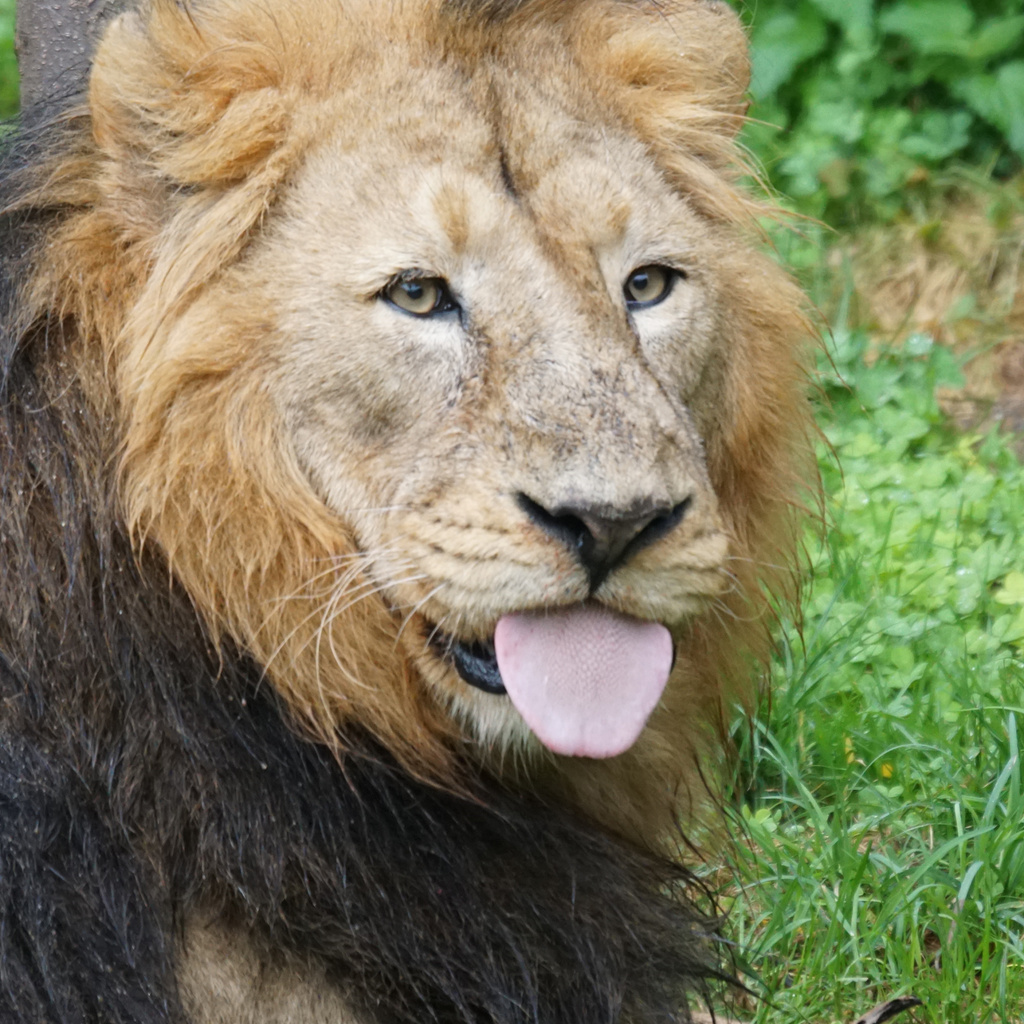}
     \\
    \cellcolor{tabfirst} \scriptsize Depth & \cellcolor{tabfirst} \scriptsize Segmentation &  \cellcolor{tabfirst} \scriptsize Edge & & \cellcolor{tabsecond} \scriptsize PASD (Zoomed) & \cellcolor{tabsecond} \scriptsize SeeSR (Zoomed) &  \cellcolor{tabsecond} \scriptsize MMSR (Zoomed) & & \cellcolor{tabthird} \scriptsize HR (Zoomed) \\
    \\
    \end{tabular}
\vskip-15pt	\captionof{figure}{
Our Multimodal Super-Resolution (MMSR) method leverages the rich context of multimodal guidance, including image captions, depth maps, semantic segmentation maps, and edges inferred from LR. MMSR surpasses state-of-the-art methods by producing more realistic results and suppressing artifacts that, while plausible, are inconsistent with the information present in the LR input.
}
\label{fig:teaser}
\end{center}
}]
\saythanks

\begin{abstract}
\vspace{-3pt}
Single-image super-resolution (SISR) remains challenging due to the inherent difficulty of recovering fine-grained details and preserving perceptual quality from low-resolution inputs.
Existing methods often rely on limited image priors, leading to suboptimal results.
We propose a novel approach that leverages the rich contextual information available in multiple modalities --including depth, segmentation, edges, and text prompts-- to learn a powerful generative prior for SISR within a diffusion model framework.
We introduce a flexible network architecture that effectively fuses multimodal information, accommodating an arbitrary number of input modalities without requiring significant modifications to the diffusion process.
Crucially, we mitigate hallucinations, often introduced by text prompts, by using spatial information from other modalities to guide regional text-based conditioning.
Each modality's guidance strength can also be controlled independently, allowing steering outputs toward different directions, such as increasing bokeh through depth or adjusting object prominence via segmentation.
Extensive experiments demonstrate that our model surpasses state-of-the-art generative SISR methods, achieving superior visual quality and fidelity.
\end{abstract}    
\section{Introduction}
Single image super-resolution (SISR) aims to generate high-resolution images from low-resolution inputs while preserving semantic identity and texture details. While it is not essential to treat SISR as a regression problem, past methods~\citep{dong2015image, wang2018esrgan, li2018multi, mei2022deep} typically use a deep neural network to learn a direct mapping from low-resolution images to high-resolution images.
Even though these regression-based methods achieve good scores on paired metrics like PSNR and SSIM, they have failed to produce results with high quality comparable to natural images, a task at which recent generative models have excelled~\cite{blau2018perception,delbracio2023inversion,chung2023prompt}.
The advent of powerful generative models, such as autoregressive models~\citep{ramesh2021zero, chen2020generative} and diffusion models~\citep{rombach2022high, podell2023sdxl, saharia2022photorealistic}, has revolutionized image generation tasks, including text-to-image synthesis. This has inspired recent efforts to leverage these pre-trained generative models for downstream tasks like SISR~\cite{meng2021sdedit, rombach2022high, brooks2023instructpix2pix}.
For instance, very recent works~\cite{yu2024scaling, qu2024xpsr} achieve super-resolution by leveraging emerging vision-language models (\eg, Gemini~\cite{team2023gemini}, LLaVA~\cite{liu2024visual}, ChatGPT-4~\cite{achiam2023gpt}) and pretrained text-to-image models to first generate captions from low-resolution images and then use these captions as prompts to generate high-resolution images.

While providing rich textual descriptions can significantly enhance the quality of generated images~\cite{betker2023improving, liu2024playground, esser2024scaling}, relying solely on text for SISR poses challenges.
Recent works~\cite{kamath2023s,liu2023visual, zhang2024visionlanguagemodelsrepresentspace, chen2024spatialvlm} have shown that text prompts cannot represent spatial relationships.
This implies that textual information, such as texture descriptions, can only be applied to the whole image.
Figure~\ref{fig:teaser} provides a representative example.
Previous text-based super-resolution methods~\cite{wu2024seesr, yang2023pixel} use a `lion' caption, which results in a furry tongue.
\emph{
However, while lions have fur, their tongues do not grow hair.}
Can we leverage spatial cues from depth and segmentation data to improve the learned prior and enhance the quality of SISR?

\begin{figure}[!t]
    \centering
    \includegraphics[width=\linewidth]{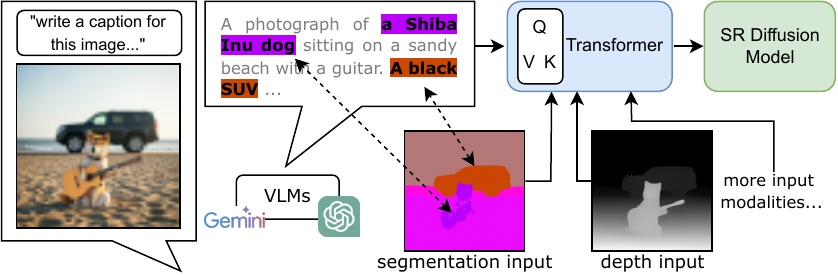}
    \vskip-5pt
    \caption{
    Language models struggle to accurately represent spatial information, leading to coarse and imprecise image super-resolution. To overcome this limitation, we incorporate additional spatial modalities like depth maps and semantic segmentation maps. These modalities provide detailed spatial context, allowing our model to implicitly align language descriptions with individual pixels through a transformer network. This enriched understanding of the image significantly enhances the realism of our super-resolution results and minimizes distortion.}
    \label{fig:teaser2}
    \vspace{-.5\baselineskip}
\end{figure}

Fortunately, we can directly extract additional modalities from the low-resolution image by using various pretrained cross-modal prediction models~\cite{cheng2022masked,yang2024depth}.
In this paper, we introduce a new diffusion model architecture dedicated to multimodal super-resolution, which is conceptually illustrated in Figure~\ref{fig:teaser2}.
By integrating multiple modalities into a single diffusion model, our method overcomes the challenges of recovering fine-grained details and preserving perceptual quality.
Specifically, we propose conditioning diffusion models on modalities including text captions, semantic segmentation maps, depth maps, and edges to implicitly align text captions for correctly prompting different regions.

We demonstrate the proposed multimodal diffusion model on the SISR task.
Our effective multimodal architecture achieves better realism in SISR results than the best text-driven method and largely eliminates hallucinations that do not match the input.
Moreover, we find that the multimodal SISR enables a new controlling feature, where we can explicitly adjust the weights of each modality to steer the generated results in different directions.

Our contributions are summarized as follows:

\begin{itemize}
\item We demonstrate the effectiveness of token-wise encoding for seamlessly injecting multiple modalities into pre-trained text-to-image diffusion models without architectural modifications or significant model size overhead.

\item We propose a novel multimodal latent connector that efficiently fuses information from different modalities, maintaining linear time complexity with respect to the number of modalities.

\item We introduce a new multimodal classifier-free guidance technique that enhances realism at higher guidance rates while mitigating excessive hallucinations and fake details.

\item Our method enables adjustment of the influence of each modality, allowing for fine-grained manipulation of SISR results while preserving realism and quality.
\end{itemize}

\section{Related Work}
\begin{figure*}[!ht]
    \centering
    \includegraphics[width=\linewidth]{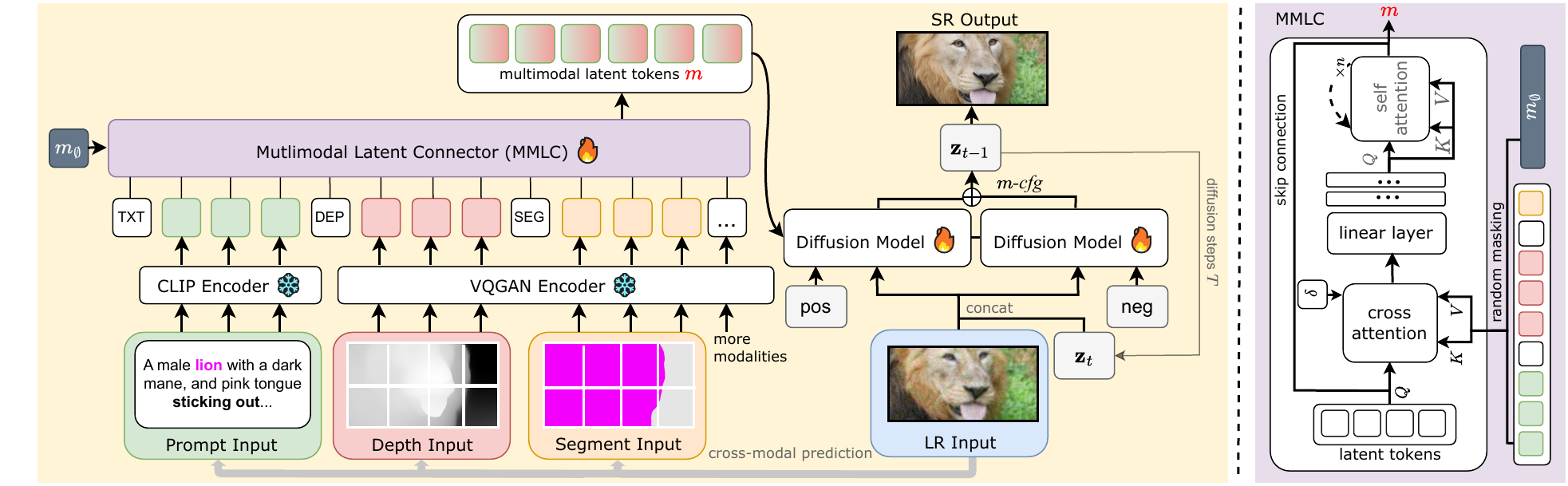}
\caption{
This diagram illustrates our multimodal super-resolution pipeline.  Starting with a low-resolution (LR) image, we extract modalities like depth and semantic segmentation maps. These modalities are encoded into tokens and transformed into multimodal latent tokens ($m$). Our diffusion model uses these tokens and the LR input to generate a high-resolution (SR) output. A multimodal classifier-free guidance (\emph{m-cfg}) refines the SR image for enhanced quality.
}
\label{fig:ibiarch}
\vspace{-1\baselineskip}
\end{figure*}

\noindent \textbf{Generative Prior Powered Super-resolution.}
Recent advances in image restoration leverage the power of foundational models to enhance the quality of degraded images~\cite{lin2023diffbir,yue2024resshift,yang2024dynamic,chen2024low,tsao2024boosting,gandikota2024text,wang2024sinsr,qu2024xpsr,noroozi2024you,mei2024codi,chung2024prompttuning,zhang2024hit, mei2024latent}. This trend is fueled by the capacity of pretrained generative models to capture natural image statistics and transfer this knowledge to the super-resolution task, enabling photorealistic image generation.
Early works in this domain include LDM-SR~\cite{rombach2022high} and StableSR~\cite{wang2024exploiting} for single image super-resolution.  Beyond this, methods like InstructPix2Pix~\cite{brooks2023instructpix2pix} utilize instructions for image editing, while ControlNet~\cite{zhang2023adding} and IP-Adapter~\cite{ye2023ip} facilitate cross-modality image translation. These approaches highlight a common observation: the quality of results in downstream tasks is strongly correlated with the quality of the pretrained models~\cite{mei2024bigger}.

Recent works focus on leveraging powerful text-to-image diffusion models, employing text prompts to fully exploit the learned prior. PASD~\cite{yang2023pixel} uses image content descriptions; SPIRE~\cite{qi2023tip} and PromptIP~\cite{potlapalli2024promptir} use degradation descriptions; and SeeSR~\cite{wu2024seesr} employs a combination of context and degradation descriptions. Compared to earlier methods that directly fine-tuned diffusion models on super-resolution data, these text-prompt-driven approaches not only achieve superior realism but also enable multi-faceted outputs by conditioning on different prompts
~\cite{gandikota2024text}. This capability enriches the typically single-output super-resolution task.

While text prompts offer advantages, they can be ambiguous in representing spatial relationships~\cite{betker2023improving, liu2024playground, esser2024scaling,chen2024spatialvlm}. Our proposed MMSR framework addresses this by integrating multimodal inputs within a novel architecture.
We also demonstrate native super-resolution effects control with this new architecture, enabling control of both overall realism and the effect of each modality.

\vspace{.5\baselineskip}
\noindent\textbf{Vision-language Understanding and Generation.} 
Recent large vision-language models (e.g., Gemini \cite{team2023gemini}, LLaVA \cite{liu2024visual}, GPT-4o \cite{achiam2023gpt}) excel in tasks like image captioning~\cite{wang2023caption}, but translating visual information into diverse modalities for image generation remains challenging. While models like 4M~\cite{4m,4m21} effectively extract high-level visual information (depth, normals, semantics, etc.), leveraging these for generation is a promising direction.
Recent exploration includes GLIGEN
~\cite{li2023gligen} generate images from bounding boxes, and ControlNet~\cite{zhang2023adding} generate images from depth and other modalities.
More recent works~\cite{gu2024kaleido, gu2024dart} explore using discrete hidden-feature tokens to guide generation.
In this paper, we introduce a new approach for multimodal control with discrete vision tokens, effectively integrating depth maps, semantic segmentation maps, and text prompts for superior performance. Experiments on real-world super-resolution demonstrate the effectiveness of our approach.

\section{Method}
Single-image super-resolution aims to recover a high-resolution image $\mathbf{x}$ from its low-resolution counterpart $\mathbf{x}_{LR}$. This is an ill-posed problem, often leading to generative models that produce ``hallucinated'' details – plausible yet inconsistent with the input.
To mitigate this, we introduce auxiliary information, such as depth ($m_\mathrm{dep}$), semantic segmentation ($m_\mathrm{seg}$), and edge maps ($m_\mathrm{edg}$), collectively denoted as $m$. By conditioning the generative process on both the low-resolution image and these auxiliary modalities, we propose a new distribution $p(\mathbf{x} | \mathbf{x}_{LR}, m)$ with reduced uncertainty compared to the original distribution $p(\mathbf{x} | \mathbf{x}_{LR})$.

This reduction in uncertainty can be understood through the lens of information theory. The auxiliary modality $m$ provides additional information about the high-resolution image $\mathbf{x}$ that is not present in the low-resolution input $\mathbf{x}_{LR}$. Since (conditional) mutual information is non-negative:
\setlength{\belowdisplayskip}{4pt} \setlength{\belowdisplayshortskip}{4pt}
\setlength{\abovedisplayskip}{4pt} \setlength{\abovedisplayshortskip}{4pt}
\begin{equation}
I(\mathbf{x} ; m | \mathbf{x}_{LR}) = H(p(\mathbf{x} | \mathbf{x}_{LR})) - H(p(\mathbf{x} | \mathbf{x}_{LR}, m)),
\end{equation}
where $H(\cdot)$ denotes entropy. Consequently, the entropy of the conditional distribution with the auxiliary modality is:
\begin{equation}
H(p(\mathbf{x} | \mathbf{x}_{LR})) \geq H(p(\mathbf{x} | \mathbf{x}_{LR}, m)).
\end{equation}
The same motivation for reducing uncertainty is also shared by recent diffusion guidance works~\cite{nichol2021improved, ho2022classifier}.
While low-entropy sampling does not guarantee sharper images, it often leads to outputs with better visual quality compared to standard sampling.
Motivated by this observation, we introduce a diffusion model~\cite{ho2020denoising} to learn the multimodal distribution $p(\mathbf{x} | \mathbf{x}_{LR}, m)$, effectively incorporating auxiliary information to enhance SISR quality.

Ideally, the auxiliary modalities should provide information complementary to the low-resolution input. In practice, the auxiliary modalities are derived from the high-resolution image during training, ensuring informative conditioning.  While we use modalities derived from the low-resolution input during inference, we demonstrate that this still leads to superior performance, including higher-quality details and improved fidelity to the input.

\subsection{Unified Mutlimodal Diffusion Conditioning}
We introduce a new diffusion network architecture, illustrated in Figure~\ref{fig:ibiarch}, for simultaneously conditioning on multiple modalities.
Unlike recent methods like ControlNet~\cite{zhang2023adding} and IP-Adapter~\cite{ye2023ip}, which duplicate network components for each modality and incur significant computational overhead, our approach leverages a pretrained VQGAN image tokenizer~\cite{esser2021taming}.
This allows us to encode diverse modalities into a unified token representation for conditioning the diffusion model, without introducing additional model parameters or modifying the diffusion network itself. These tokens are concatenated with the text prompt embedding and used for cross-attention within the diffusion model. To efficiently process this long token sequence, we introduce a lightweight multimodal connector. This connector employs a dedicated architecture to achieve linear complexity for cross-attention, significantly reducing the computational burden.
In what follows we present the implementation details.

\vspace{.5em}
\noindent \textbf{Token-wise Multimodal Encoding.}
While VQGAN~\cite{esser2021taming} has proven effective for cross-modal encoding in image understanding and generation~\cite{bai2024sequential,4m, 4m21}, its optimal application for image super-resolution remains an open question.  Unlike tasks focused on understanding or generation, super-resolution requires tokenization to not only capture semantic information but also preserve pixel-wise details crucial for accurate reconstruction, such as spatial relationships within a depth map.
Inspired by recent discussions on using discrete or continuous tokenization for autoregressive image generation~\cite{tang2024hart}, we investigate the impact of quantization on multimodal super-resolution.  Specifically, we analyze how different quantization strategies affect the reconstruction quality of individual modalities and the final generated image.  To this end, we utilize a VQGAN model pre-trained on a large-scale multimodal image dataset.

\begin{figure}[!t]
    \centering
    \def\xwidth{0.24\linewidth}
    \def\ywidth{0.12\linewidth}
    \setlength{\tabcolsep}{1pt}
    \renewcommand\arraystretch{0.6}
    \settowidth{\imagewidth}{\includegraphics{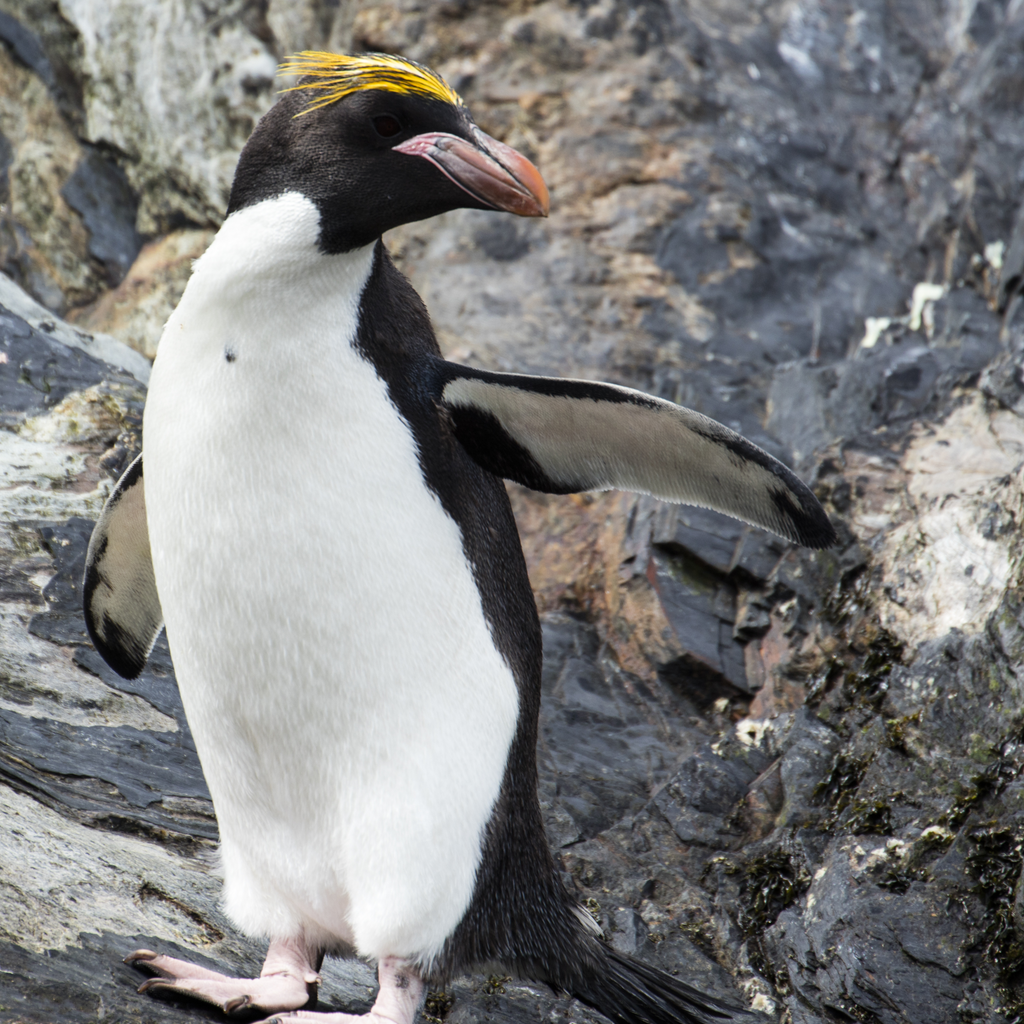}}
    \begin{tabular}[t]{c c c c}\\
    \includegraphics[width=\xwidth, clip=true, trim = 0.1\imagewidth{} 0.3\imagewidth{} 0.1\imagewidth{} 0.3\imagewidth{}]{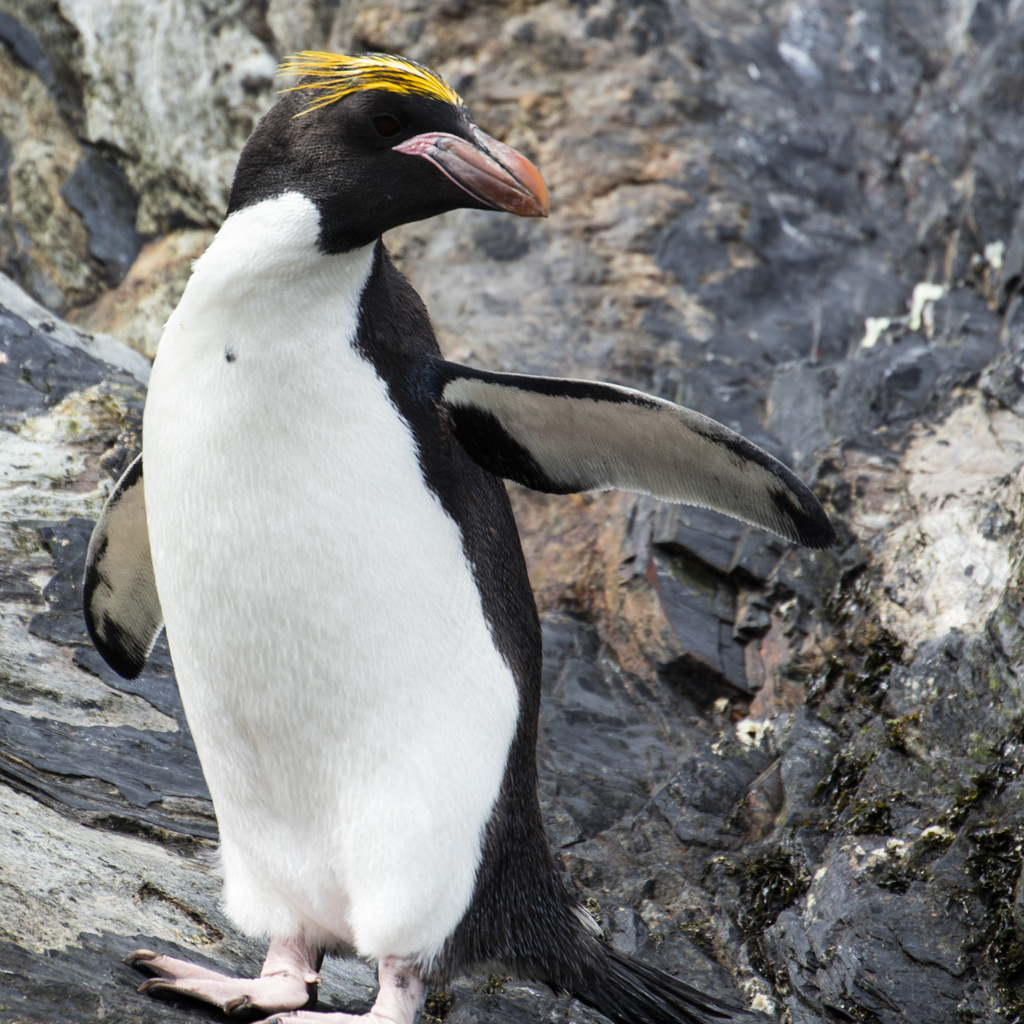} &
    \includegraphics[width=\xwidth, clip=true, trim = 0.1\imagewidth{} 0.3\imagewidth{} 0.1\imagewidth{} 0.3\imagewidth{}]{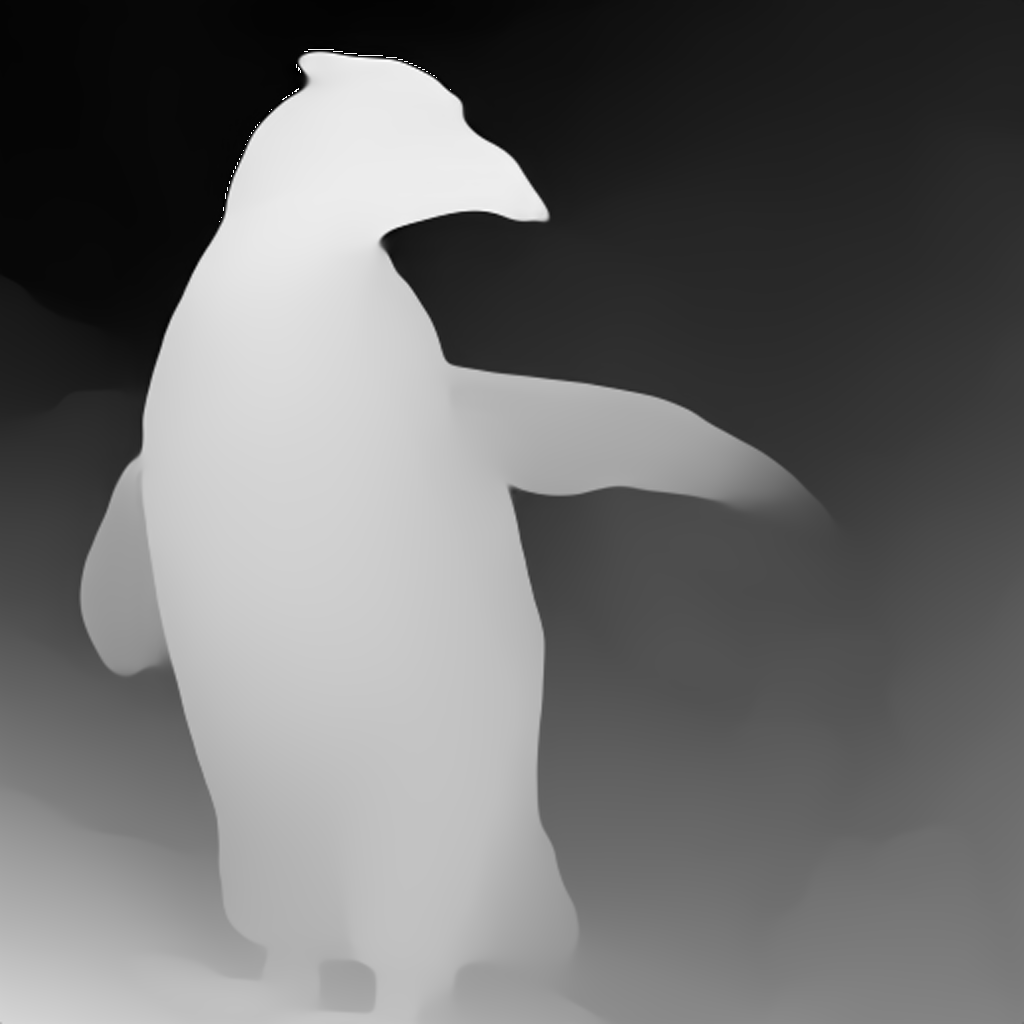} &
    \settowidth{\imagewidth}{\includegraphics{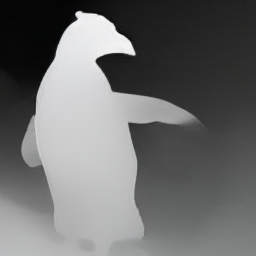}}
    \includegraphics[width=\xwidth, clip=true, trim = 0.1\imagewidth{} 0.3\imagewidth{} 0.1\imagewidth{} 0.3\imagewidth{}]{ibis-results/000000_dep_recon_discrete.png} &
    \settowidth{\imagewidth}{\includegraphics{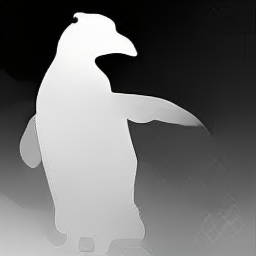}}
    \includegraphics[width=\xwidth, clip=true, trim = 0.1\imagewidth{} 0.3\imagewidth{} 0.1\imagewidth{} 0.3\imagewidth{}]{ibis-results/000000_dep_recon_continous.png}  \\
    \settowidth{\imagewidth}{\includegraphics{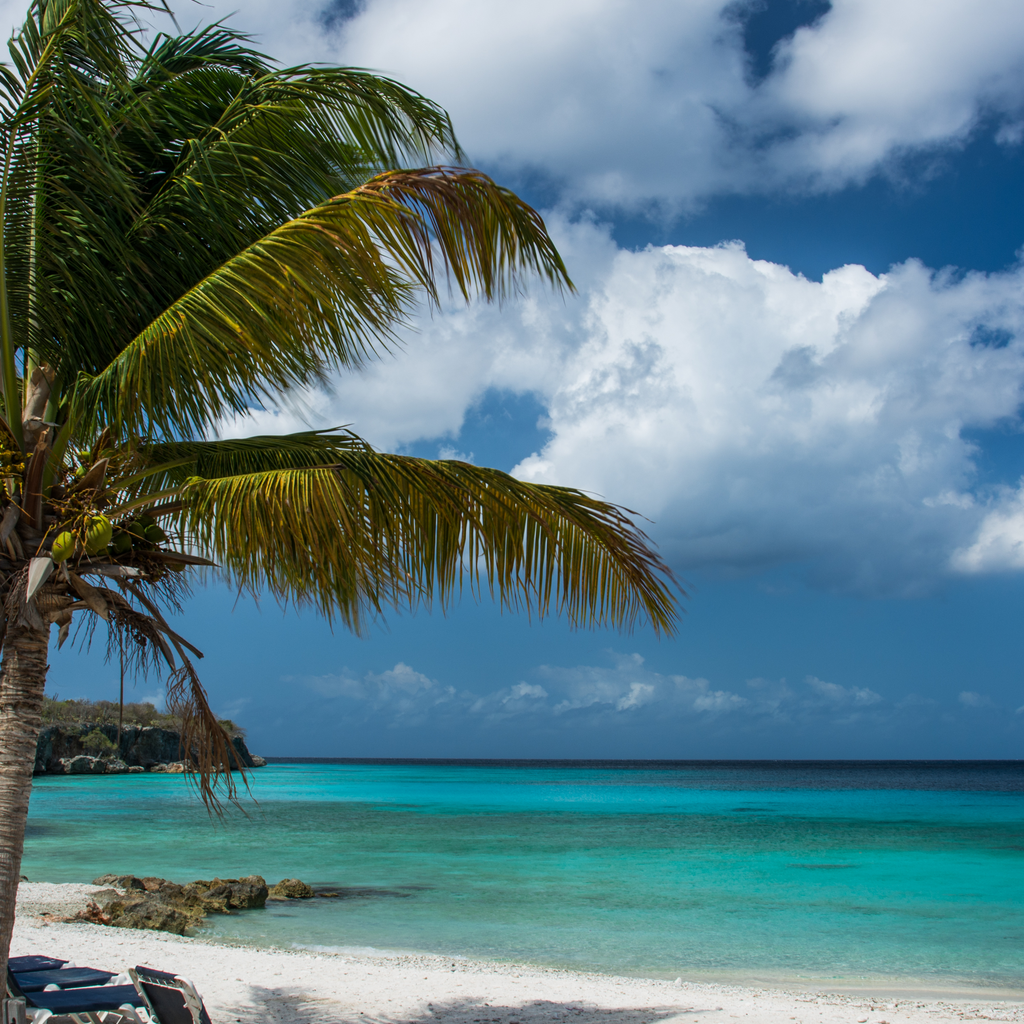}}
    \includegraphics[width=\xwidth, clip=true, trim = 0.1\imagewidth{} 0.2\imagewidth{} 0.1\imagewidth{} 0.4\imagewidth{}]{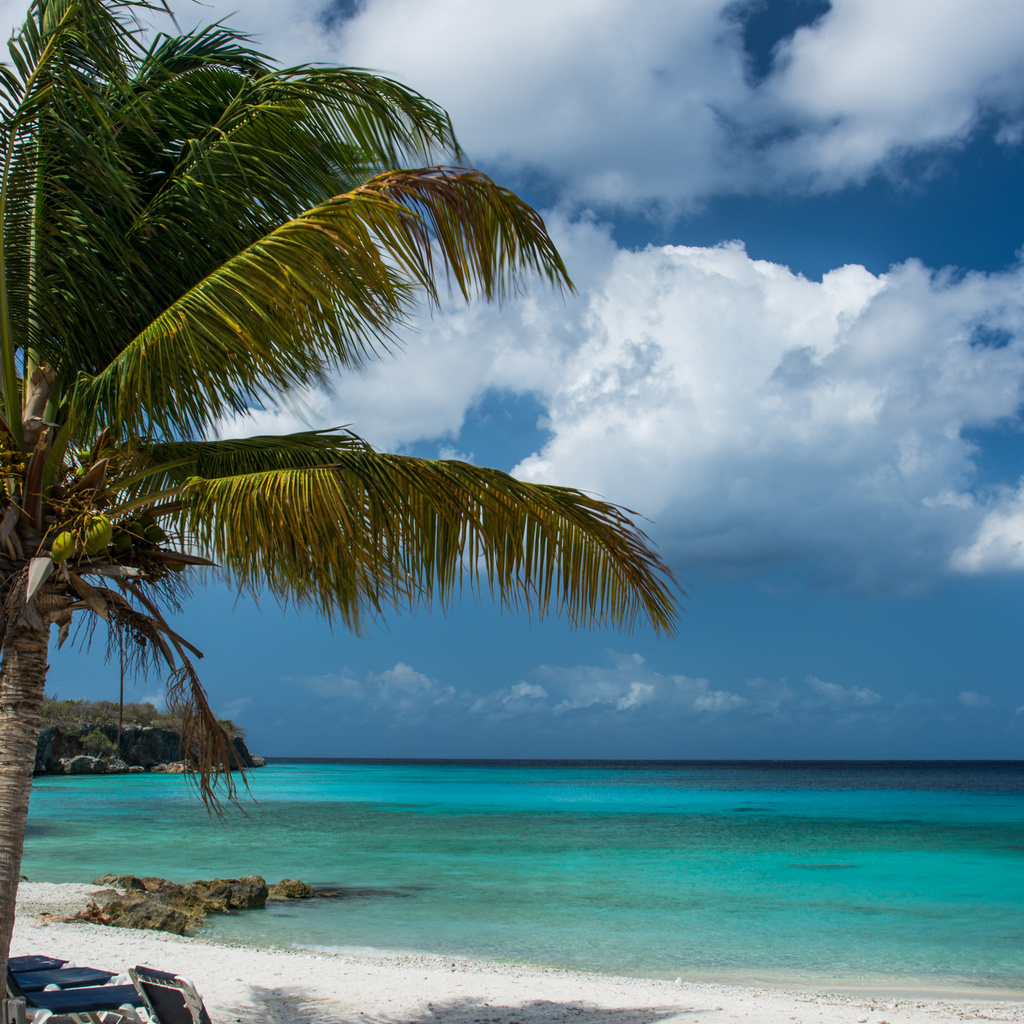} &
    \includegraphics[width=\xwidth, clip=true, trim = 0.1\imagewidth{} 0.2\imagewidth{} 0.1\imagewidth{} 0.4\imagewidth{}]{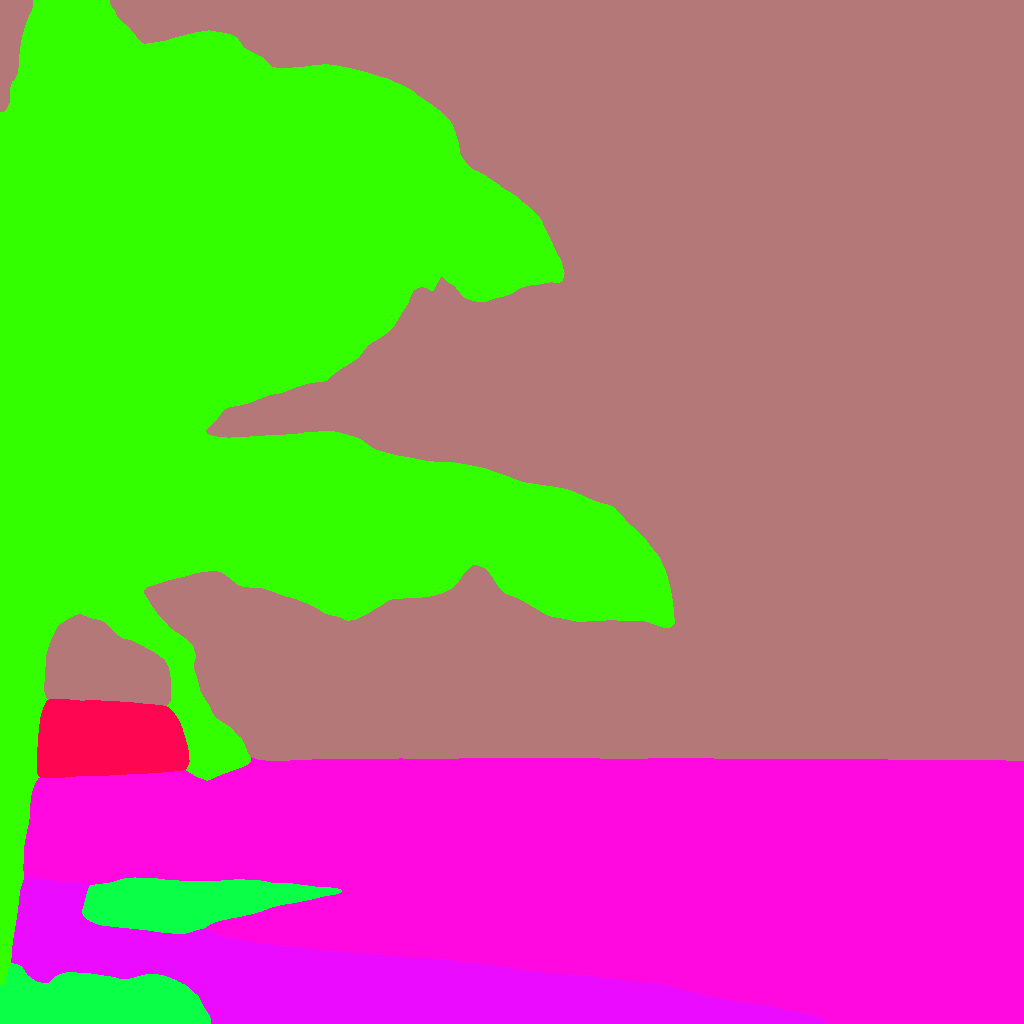} &
    \settowidth{\imagewidth}{\includegraphics{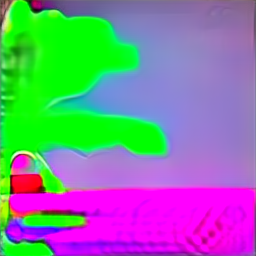}}
    \includegraphics[width=\xwidth, clip=true, trim = 0.1\imagewidth{} 0.2\imagewidth{} 0.1\imagewidth{} 0.4\imagewidth{}]{ibis-results/000005_seg_recon_continous.png} &
    \settowidth{\imagewidth}{\includegraphics{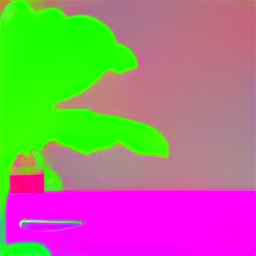}}
    \includegraphics[width=\xwidth, clip=true, trim = 0.1\imagewidth{} 0.2\imagewidth{} 0.1\imagewidth{} 0.4\imagewidth{}]{ibis-results/000005_seg_recon_discrete.png}  \\
    \settowidth{\imagewidth}{\includegraphics{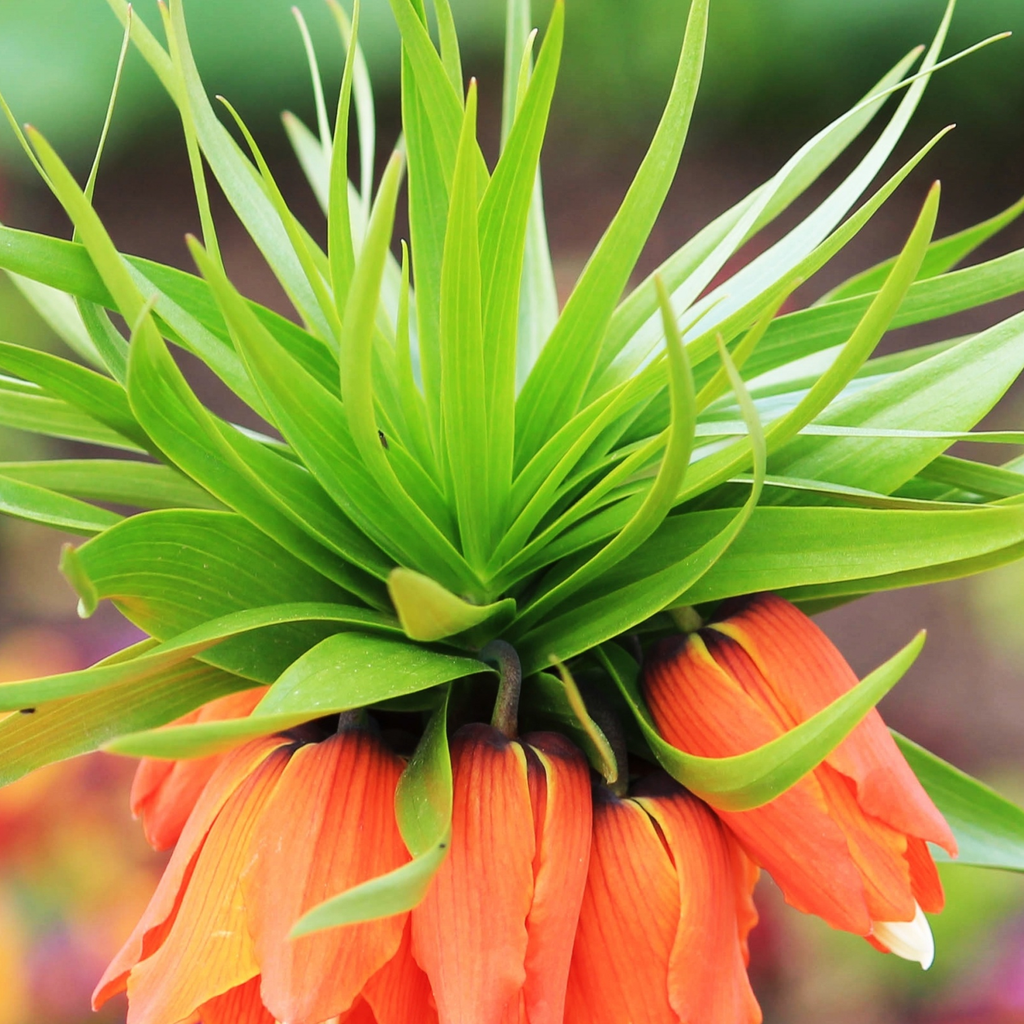}}
    \includegraphics[width=\xwidth, clip=true, trim = 0.1\imagewidth{} 0.3\imagewidth{} 0.1\imagewidth{} 0.3\imagewidth{}]{ibis-results/000002_input_hr.png} &
    \includegraphics[width=\xwidth, clip=true, trim = 0.1\imagewidth{} 0.3\imagewidth{} 0.1\imagewidth{} 0.3\imagewidth{}]{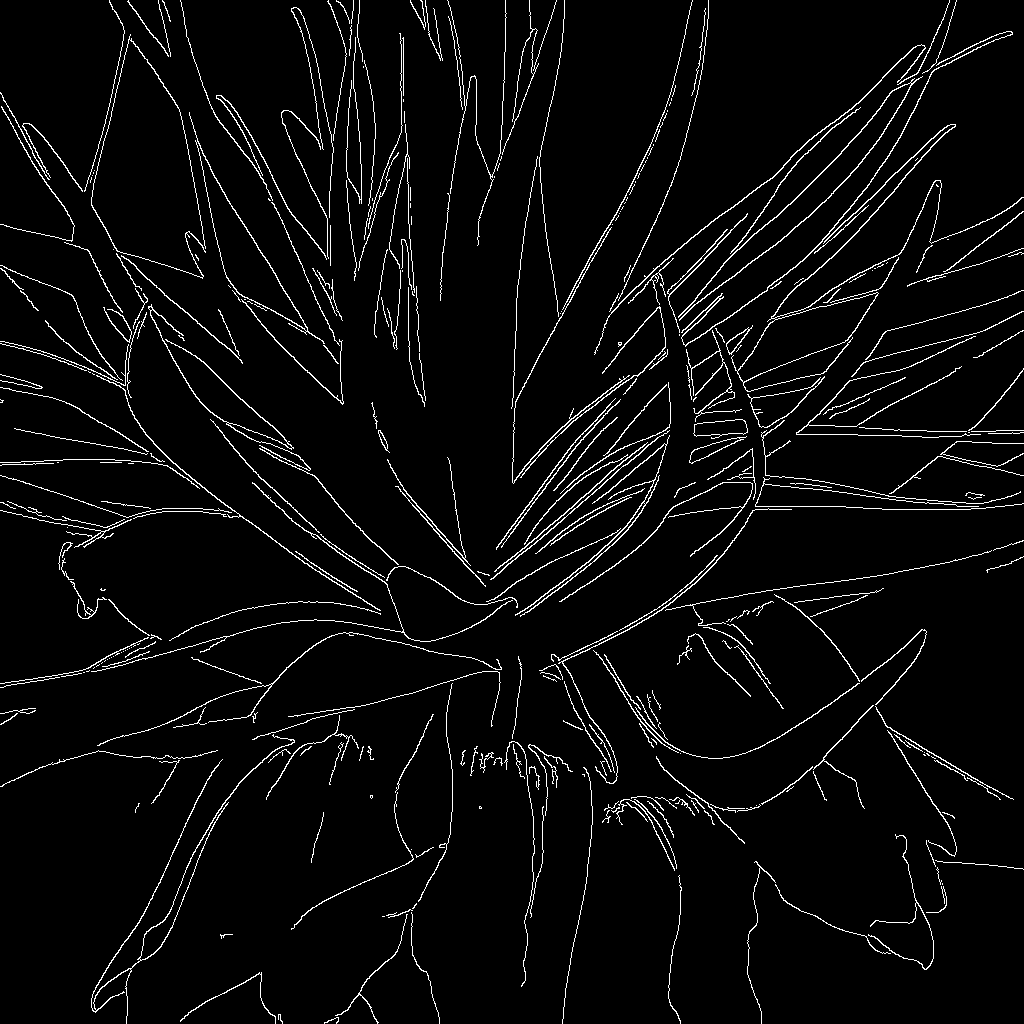} &
    \settowidth{\imagewidth}{\includegraphics{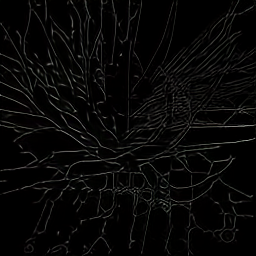}}
    \includegraphics[width=\xwidth, clip=true, trim = 0.1\imagewidth{} 0.3\imagewidth{} 0.1\imagewidth{} 0.3\imagewidth{}]{ibis-results/000002_input_edg_conti.png} &
    \settowidth{\imagewidth}{\includegraphics{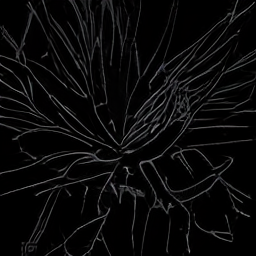}}
    \includegraphics[width=\xwidth, clip=true, trim = 0.1\imagewidth{} 0.3\imagewidth{} 0.1\imagewidth{} 0.3\imagewidth{}]{ibis-results/000002_input_edg_discrete.png}  \\
    \scriptsize Input Image & \scriptsize Input Modality & \scriptsize Continuous Tokens & \scriptsize Discrete Tokens  \\
    \end{tabular}
    \vspace{-.5\baselineskip}
    \caption{
    Using discrete multimodal tokens leads to superior reconstruction of  modalities compared to continuous tokens.}
    \label{fig:multimodalvq}
    \vspace{-1\baselineskip}
\end{figure}

Figure~\ref{fig:multimodalvq} compares the reconstruction quality of multimodal tokens before and after quantization. As the figure illustrates, discrete tokens (post-quantization) better preserve individual modality information, while continuous tokens introduce noticeable artifacts. Consequently, we employ discrete tokens for our super-resolution experiments.

Our multimodal encoding process utilizes the encoder and quantizer of the pre-trained VQGAN. Each $256 \times 256$ input modality is encoded into a $16 \times 16$ multimodal token sequence with a feature dimension of 256. These tokens are then quantized using a codebook of size 1024. To facilitate concatenation with text embeddings, we pad the feature dimension of the multimodal tokens to 1024, resulting in a $(256 \times 3 + 77) \times 1024$ multimodal token sequence condition.

\vspace{.5em}
\noindent \textbf{Multimodal Latent Connector.}
While cross-attention provides a flexible mechanism for conditioning diffusion models on multimodal data, its quadratic complexity with respect to the number of condition tokens introduces a significant computational burden.
To address this, we introduce the Multimodal Latent Connector (MMLC), inspired by recent advances in efficient attention architectures~\cite{lee2019set, wang2020linformer, jaegle2021perceiver}.

As illustrated in Figure~\ref{fig:ibiarch}, the MMLC employs a transformer architecture to efficiently process the multimodal token sequence.
The transformer receives two inputs: a randomly initialized sequence of learnable latent tokens, and the multimodal input sequence.
The output is a token sequence of the same length as the latent token sequence, which serves as conditioning for the diffusion model.
Therefore, the diffusion model conditions on fixed-length latent tokens (128 in our experiments), which are significantly shorter than the original multimodal token sequence input.
The MMLC distills the essential information from the longer multimodal token sequence into the shorter multimodal latent token sequence through cross-attention.
Following cross-attention, several self-attention blocks further process the latent token sequence, allowing the model to fully integrate the distilled information.

This approach significantly reduces the computational cost of cross-attention in the diffusion model. Standard self-attention operates on the full multimodal token sequence ($K, V \in \mathbb{R}^{M \times D}$), resulting in a time complexity of $\mathcal{O}(M^2)$, where $M$ is the length of the sequence and $D$ is the dimensionality of each token. In contrast, the MMLC uses a cross-attention between the latent token sequence (of size $N\times D$) and the multimodal token sequence which reduces this to $\mathcal{O}(MN)$. Here $N$ is the length of the latent sequence and $N \ll M$. This linear complexity with respect to the multimodal sequence length enables efficient processing of high-dimensional multimodal data, effectively capturing essential information for super-resolution.

\begin{figure*}[!ht]
    \centering
    \def\xwidth{0.14\linewidth}
    \def\ywidth{0.12\linewidth}
    \setlength{\tabcolsep}{1pt}
    \renewcommand\arraystretch{0.6}
    \resizebox{\linewidth}{!}{
    \begin{tabular}[t]{c c c c c c c c}\\
    \multicolumn{3}{c}{\cellcolor{tabfirst} \scriptsize Inputs} & & \multicolumn{4}{c}{\cellcolor{tabsecond} \scriptsize Outputs} \\
    
    \cellcolor{tabfirst} \includegraphics[width=\xwidth]{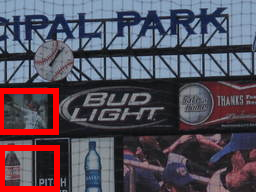}
    & 
    \cellcolor{tabfirst} \cellcolor{tabfirst} \textshape{\xwidth}{The image shows the words "CIPAL PARK" prominently displayed on a metal structure.  Below the main sign are several smaller screens displaying baseball  \dots}
    &
    \cellcolor{tabfirst} \settowidth{\imagewidth}{\includegraphics{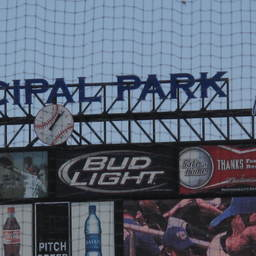}}
    \includegraphics[width=\xwidth, clip=true, trim = 0\imagewidth{} 0.2\imagewidth{} 0.8\imagewidth{} 0.65\imagewidth{}]{figures/realsr/EVAL_1aa13ca74154f6ff_lr.png}
    & &
    \cellcolor{tabsecond}
    \settowidth{\imagewidth}{\includegraphics{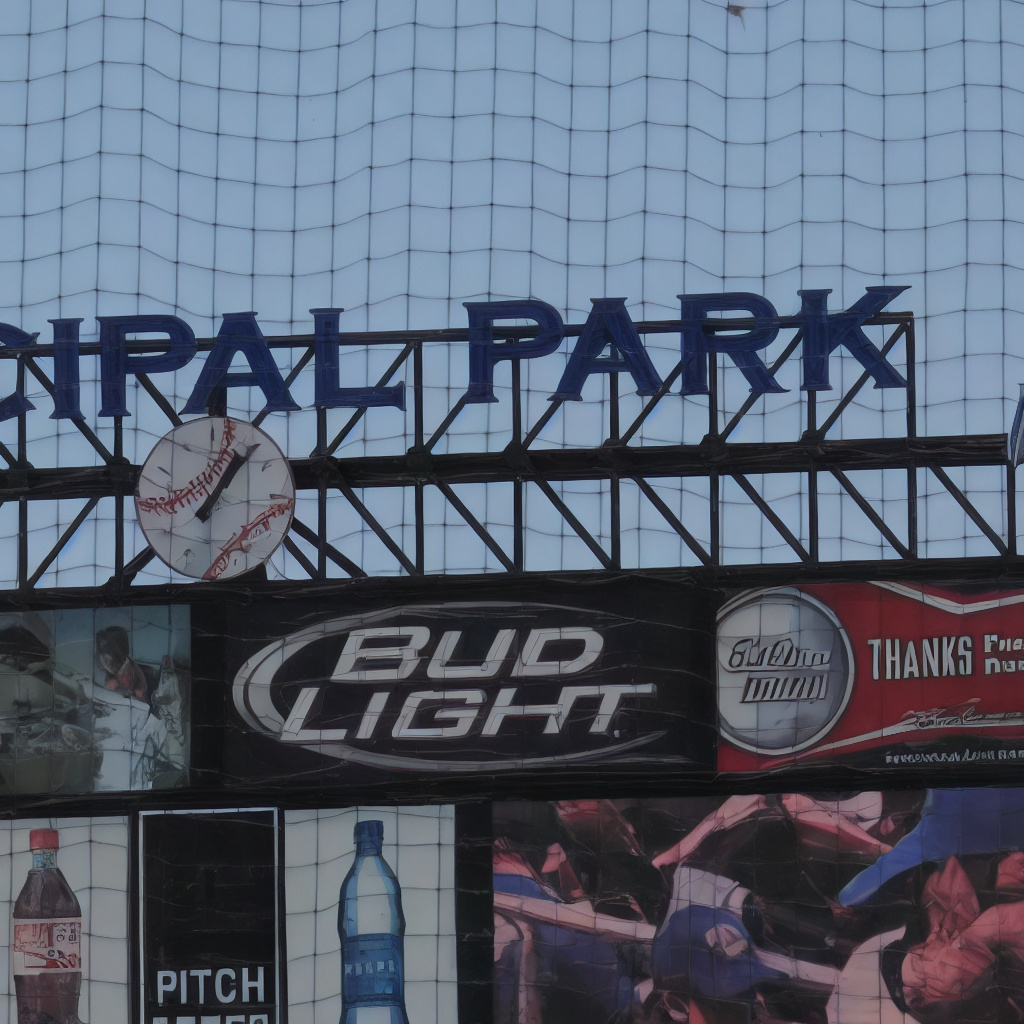}}
    \includegraphics[width=\xwidth, clip=true, trim = 0\imagewidth{} 0.2\imagewidth{} 0.8\imagewidth{} 0.65\imagewidth{}]{figures/realsr/EVAL_1aa13ca74154f6ff_pasd.jpg} 
    &
    \cellcolor{tabsecond} \settowidth{\imagewidth}{\includegraphics{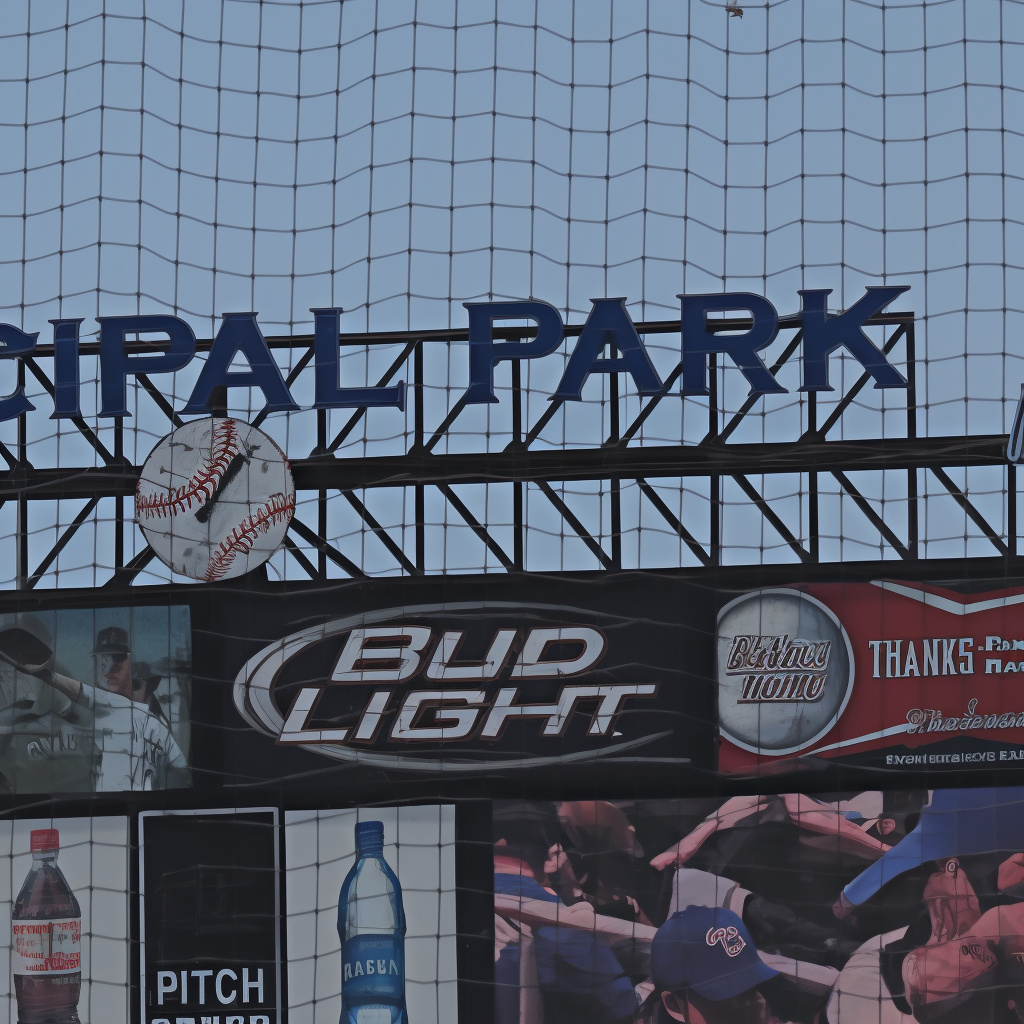}}
    \includegraphics[width=\xwidth, clip=true, trim = 0\imagewidth{} 0.2\imagewidth{} 0.8\imagewidth{} 0.65\imagewidth{}]{figures/realsr/EVAL_1aa13ca74154f6ff_seesr.jpg}
    &
    \cellcolor{tabsecond} \settowidth{\imagewidth}{\includegraphics{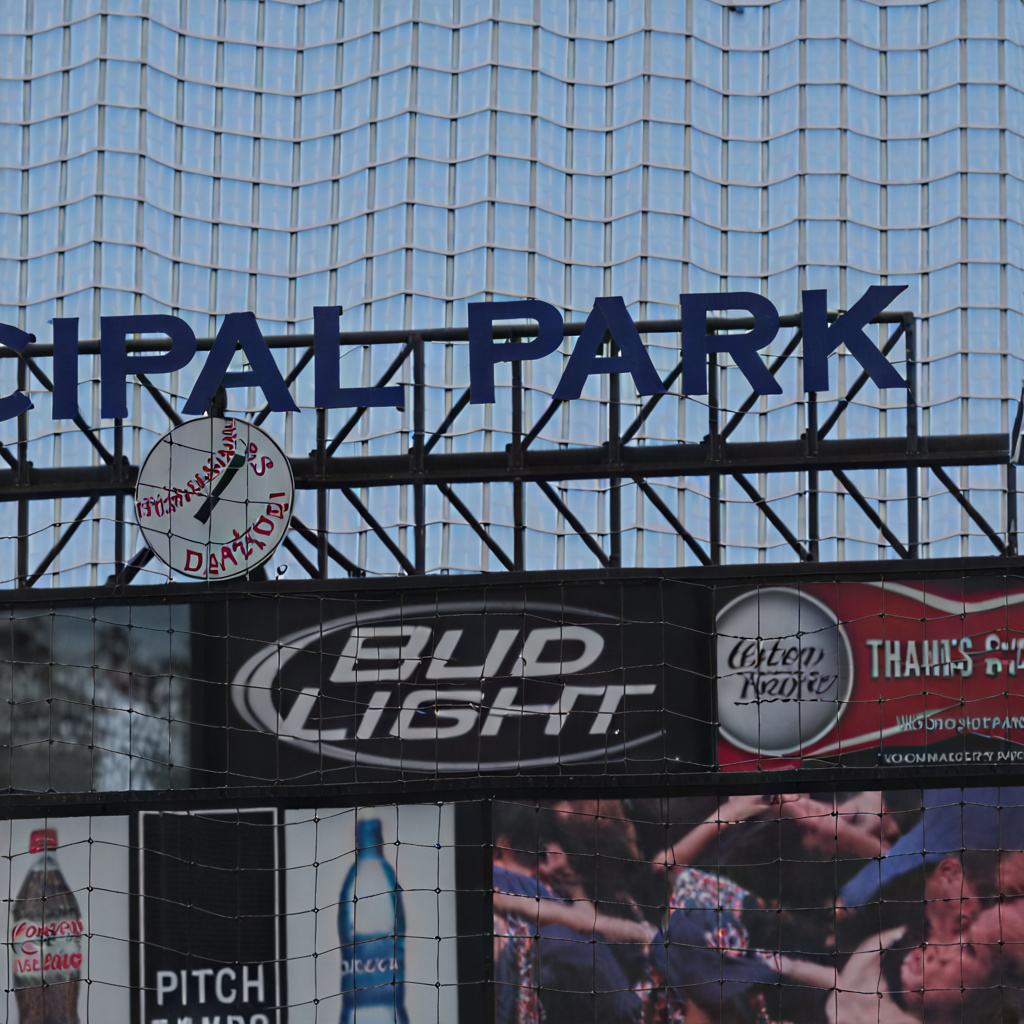}}
    \includegraphics[width=\xwidth, clip=true, trim = 0\imagewidth{} 0.2\imagewidth{} 0.8\imagewidth{} 0.65\imagewidth{}]{figures/realsr/EVAL_1aa13ca74154f6ff_0_supir.jpg}
    &
    \cellcolor{tabsecond} \settowidth{\imagewidth}{\includegraphics{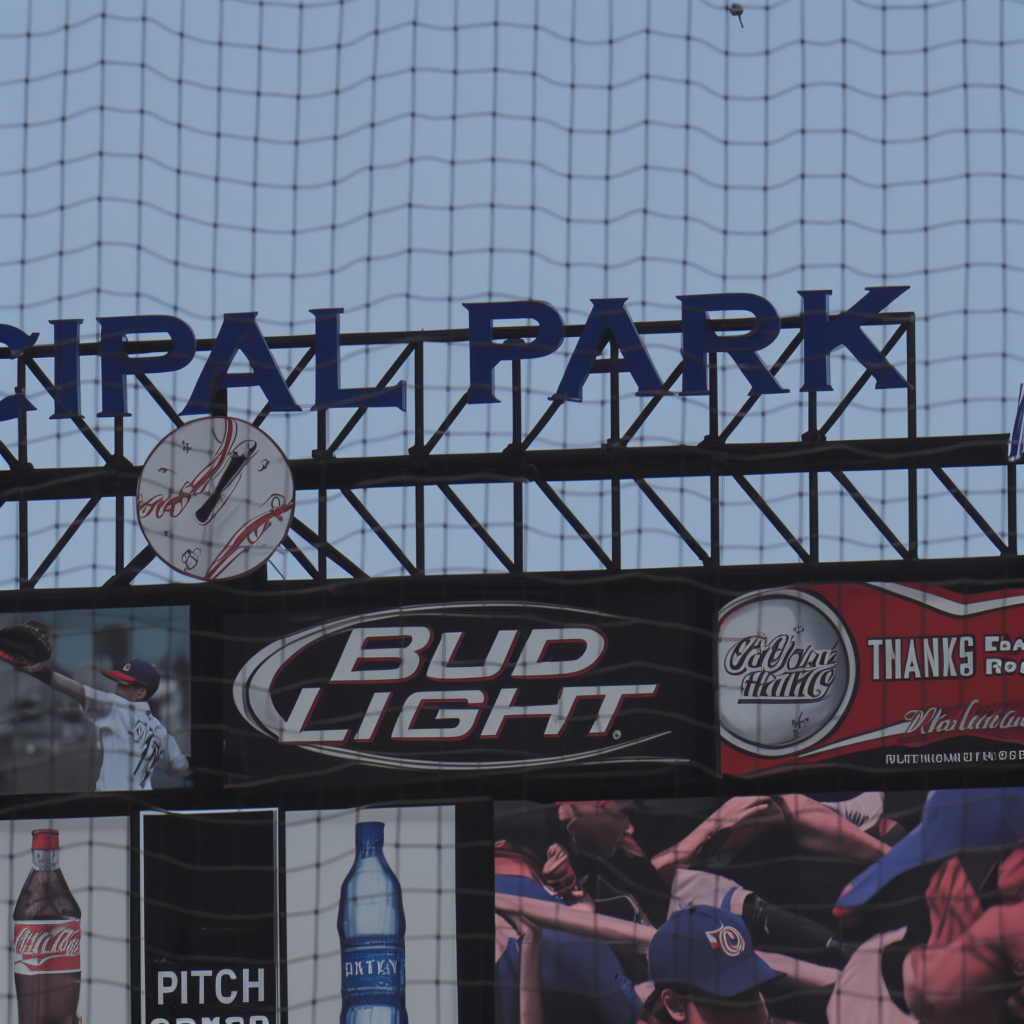}}
    \includegraphics[width=\xwidth, clip=true, trim = 0\imagewidth{} 0.2\imagewidth{} 0.8\imagewidth{} 0.65\imagewidth{}]{figures/realsr/EVAL_1aa13ca74154f6ff_ours.png} \\
    \cellcolor{tabfirst} \scriptsize LR & \cellcolor{tabfirst} \scriptsize Caption & \cellcolor{tabfirst} \scriptsize Patch 1 & & \cellcolor{tabsecond} \scriptsize PASD (Zoomed) & \cellcolor{tabsecond} \scriptsize SeeSR (Zoomed)  & \cellcolor{tabsecond} \scriptsize SUPIR (Zoomed)  &  \cellcolor{tabsecond} \scriptsize MMSR  (Zoomed) \\
    \cellcolor{tabfirst} \includegraphics[width=\xwidth]{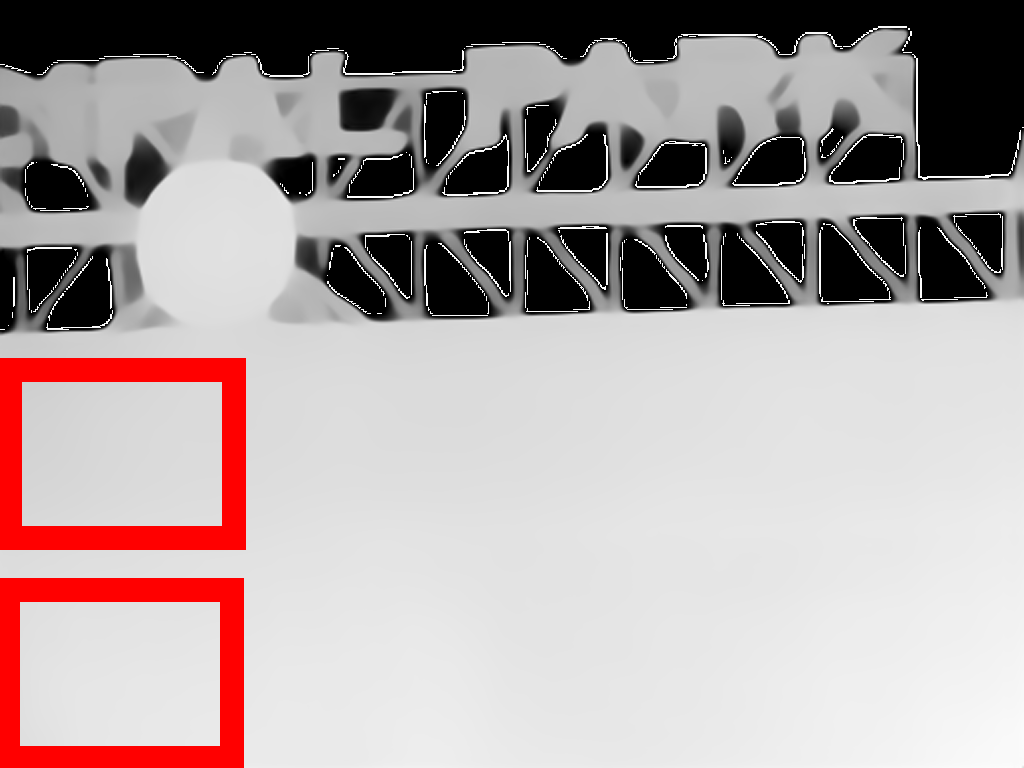} &
    \cellcolor{tabfirst} \includegraphics[width=\xwidth]{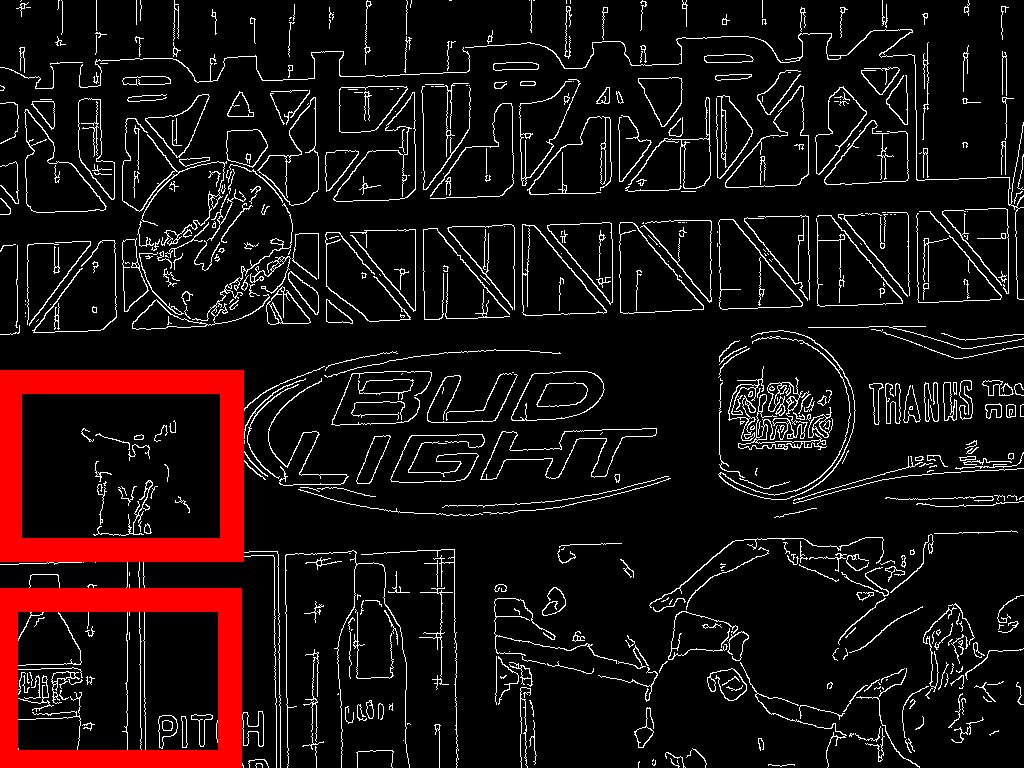} &
    \cellcolor{tabfirst} \settowidth{\imagewidth}{\includegraphics{figures/realsr/EVAL_1aa13ca74154f6ff_lr.png}}
    \includegraphics[width=\xwidth, clip=true, trim = 0.0\imagewidth{} 0.0\imagewidth{} 0.8\imagewidth{} 0.85\imagewidth{}]{figures/realsr/EVAL_1aa13ca74154f6ff_lr.png}
    & & 
    \cellcolor{tabsecond}
    \settowidth{\imagewidth}{\includegraphics{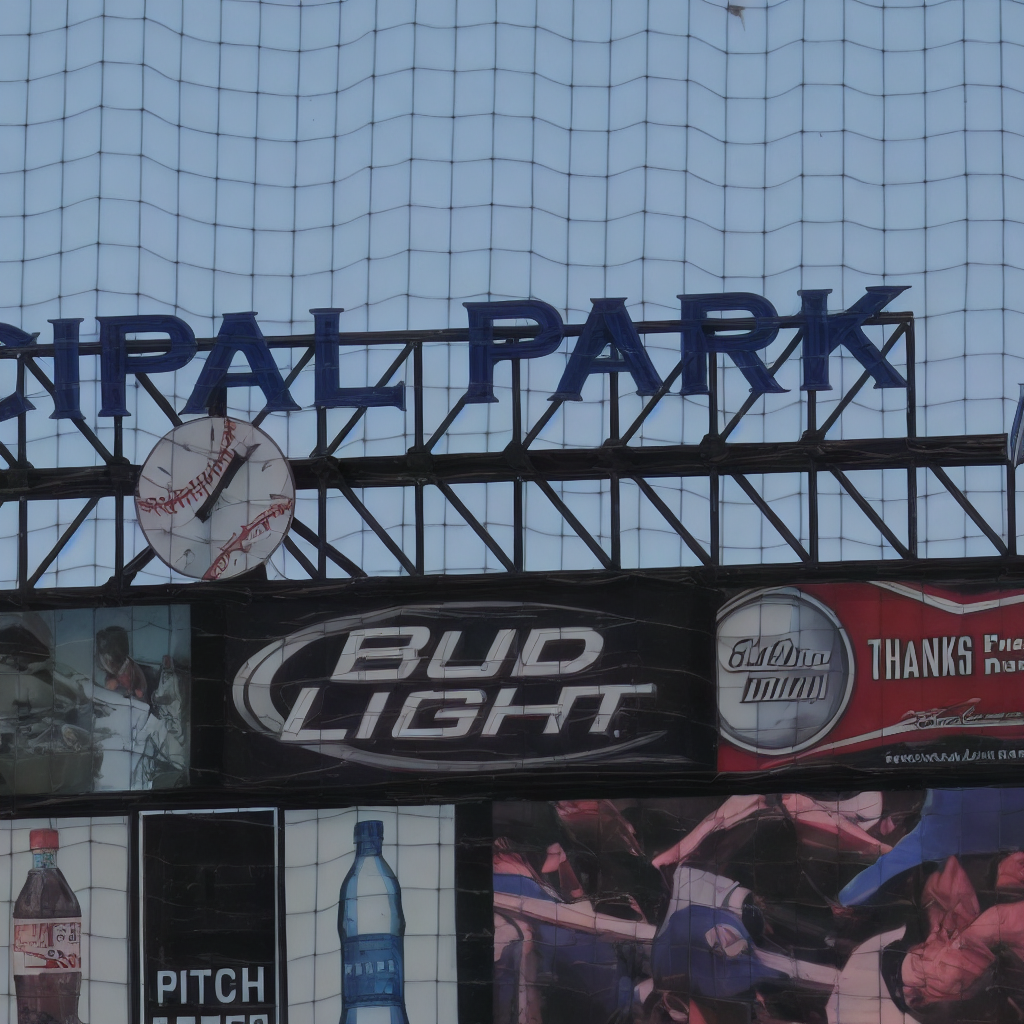}}
    \includegraphics[width=\xwidth, clip=true, trim = 0.0\imagewidth{} 0.0\imagewidth{} 0.8\imagewidth{} 0.85\imagewidth{}]{figures/realsr/EVAL_1aa13ca74154f6ff_pasd.jpg}
    &
    \cellcolor{tabsecond}  
    \settowidth{\imagewidth}{\includegraphics{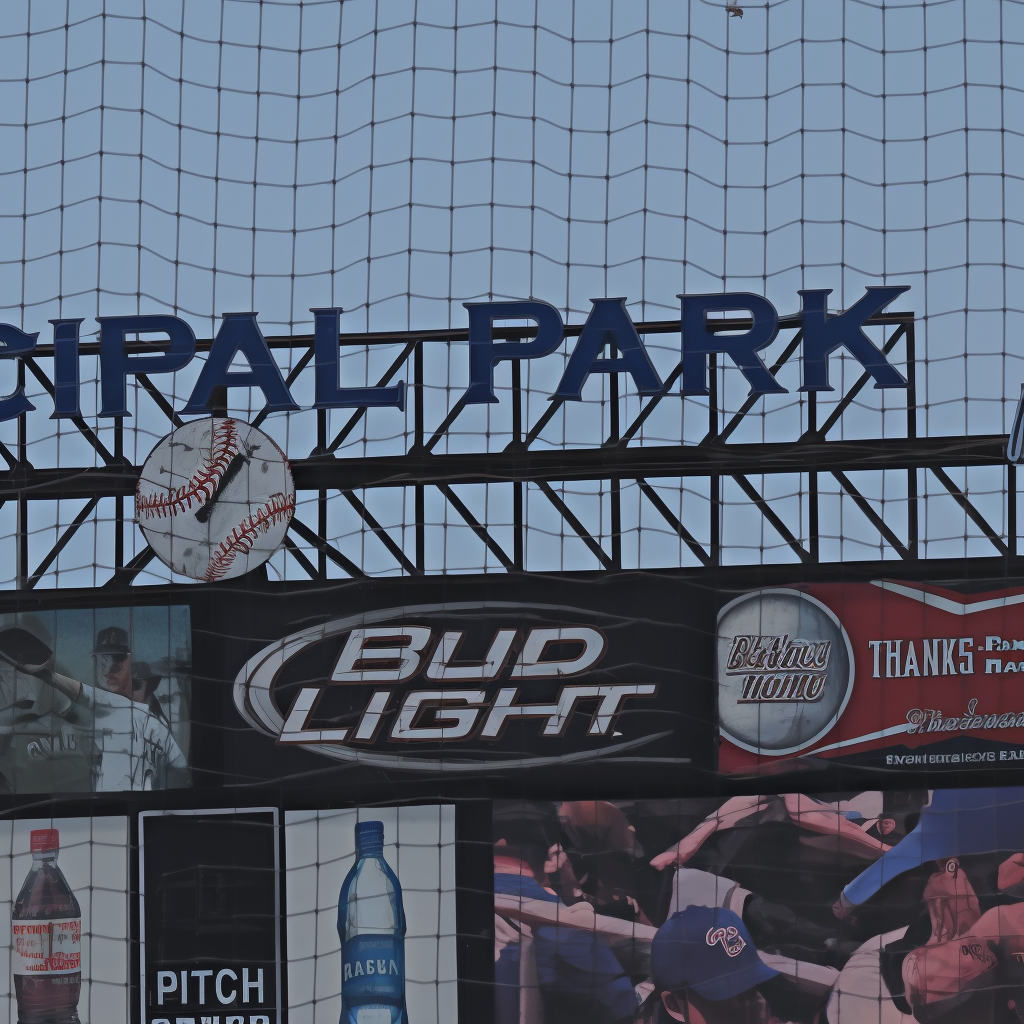}}
    \includegraphics[width=\xwidth, clip=true, trim = 0.0\imagewidth{} 0.0\imagewidth{} 0.8\imagewidth{} 0.85\imagewidth{}]{figures/realsr/EVAL_1aa13ca74154f6ff_seesr.jpg} 
    &
    \cellcolor{tabsecond}  
    \settowidth{\imagewidth}{\includegraphics{figures/realsr/EVAL_1aa13ca74154f6ff_0_supir.jpg}}
    \includegraphics[width=\xwidth, clip=true, trim = 0.0\imagewidth{} 0.0\imagewidth{} 0.8\imagewidth{} 0.85\imagewidth{}]{figures/realsr/EVAL_1aa13ca74154f6ff_0_supir.jpg} 
    &
    \cellcolor{tabsecond}  
    \settowidth{\imagewidth}{\includegraphics{figures/realsr/EVAL_1aa13ca74154f6ff_ours.png}}
    \includegraphics[width=\xwidth, clip=true, trim = 0.0\imagewidth{} 0.0\imagewidth{} 0.8\imagewidth{} 0.85\imagewidth{}]{figures/realsr/EVAL_1aa13ca74154f6ff_ours.png}
     \\
    \cellcolor{tabfirst} \scriptsize Depth & \cellcolor{tabfirst} \scriptsize Edge &  \cellcolor{tabfirst} \scriptsize Patch 2 & & \cellcolor{tabsecond} \scriptsize PASD (Zoomed) & \cellcolor{tabsecond} \scriptsize SeeSR (Zoomed) & \cellcolor{tabsecond} \scriptsize SUPIR (Zoomed) &  \cellcolor{tabsecond} \scriptsize MMSR (Zoomed) \\
    \end{tabular}
    }
    \vspace{-.5\baselineskip}
    \caption{MMSR super-resolution results on real-world images compared with state-of-the-art methods. Zoom in to appreciate the details.
    }
    \label{fig:realresults}
    \vspace{-1\baselineskip}
\end{figure*}

\vspace{.5em}
\noindent \textbf{Flexible Multimodal Input.}
\label{sec:emptoken}
To enhance flexibility and robustness, we enable our method to handle scenarios where certain input modalities are unreliable or unavailable.  
We adopt a learnable embedding approach inspired by DALL-E 2~\cite{ramesh2022hierarchical}.
Specifically, we introduce a special learnable token, $m_\emptyset$,  optimized alongside the diffusion model and MMLC to represent the absence of a modality. During training, we independently randomly replace each modality with 256 $m_\emptyset$ tokens with a probability of 0.1. This encourages the model to learn robust representations that can effectively handle missing information.
During inference, any missing modality is represented by a sequence of 256 $m_\emptyset$ tokens. This approach allows for flexible multimodal input, enabling the model to generate high-quality images even with limited or no auxiliary modalities. As demonstrated in our experiments, this strategy significantly improves performance when dealing with input that includes fewer modalities than were available during training.

\subsection{Multimodal Guidance and Control}
Building upon the success of guidance techniques~\cite{dhariwal2021diffusion,ho2022classifier} in improving sample quality across various image generation tasks~\cite{ramesh2022hierarchical,podell2023sdxl,nichol2021glide}, recent diffusion-based SISR methods have begun incorporating prompt tuning to enhance super-resolution results~\cite{wang2024exploiting, wu2024seesr, yu2024scaling}.
These methods improve the result by using negative prompts with Classifier-free Guidance (CFG)~\cite{ho2022classifier}.
which can be expressed as:
\begin{equation}
    \tilde \bepsilon(\bz_t, c) = (1 + w) \ \bepsilon(\bz_t, c, \mathrm{pos}) - w \ \bepsilon(\bz_t, c, \mathrm{neg}),
\end{equation}
where $\tilde \bepsilon(\bz_t, c)$ represents the guided denoising process, $\bepsilon(\cdot)$ denotes the diffusion model, $\bz_t$ denotes the noisy image latent, $c = \{\bx_{\mathrm{LR}}, t, \dots\}$ denotes the conditioning inputs (including the low-resolution image $\bx_{\mathrm{LR}}$, timestep $t$, and other parameters), $w$ is the guidance scale, and $\mathrm{pos}$ and $\mathrm{neg}$ are the positive and negative prompts.  While increasing $w$ often leads to sharper and more detailed outputs, it can also exacerbate hallucination, resulting in details inconsistent with the low-resolution input.
Such artifacts are widely reported but are  difficult to suppress, even with recent efforts like balancing the training data~\cite{yu2024scaling}.

\vspace{.5em}
\noindent \textbf{Multimodal Classifier-free Guidance.}
To mitigate the issue of excessive hallucination often associated with high guidance scales in CFG, we propose a novel multimodal guidance strategy.
We argue that the artifacts come from the weak guidance in the negative prompting process.
Instead of simply relying on text prompts for guidance, we leverage the rich information encoded in the multimodal latent tokens to strengthen both positive and negative promptings.
Specifically, we condition both the postive and negative generation processes on the multimodal latent token sequence, denoted as $m$. This leads to the following multimodal CFG:
\begin{equation}
\begin{aligned}
    \tilde \bepsilon(\bz_t, c, m) \!=\! (1\! +\! w) \bepsilon(\bz_t, c, \mathrm{pos}, m) - w \ \bepsilon(\bz_t, c, \mathrm{neg}, m).
\end{aligned}
\end{equation}
Sec.~\ref{sec:multimodalcfg} shows that this change in negative generation helps to achieve a better trade-off between perceptual quality and identity preservation, better maintaining the semantic content of the low-resolution input in the upscaled output, compared with the standard CFG in previous text-based methods.

\vspace{.5em}
\noindent \textbf{Scaling Single-modal Guidance.}
Multimodal CFG  effectively controls the overall influence of the prompts but does not offer control over each modality individually.
To address this limitation, we introduce a mechanism to selectively amplify or suppress the contribution of specific modalities.
Specifically, we modify the attention \emph{temperature} $\delta$ of MMLC during cross-conditioning, where $\delta$ scales the attention maps before applying the softmax operation:
\begin{equation}
    \mathrm{Attention}(Q,K, V) = \mathrm{softmax}\left(\frac{QK^T}{\delta}\right)V.
\end{equation}
This parameter controls the sensitivity of the attention mechanism to differences between the feature sequence $Q$ and the multimodal token sequences $K$ and $V$.
A smaller temperature $\delta$ typically leads to a stronger conditioning effect on the attention mechanism.
In the standard scaled dot-product attention, the temperature $\delta$ is empirically set to the square root of the key dimension $\sqrt{d_k}$.
We observe that scaling the standard temperature $\sqrt{d_k}$ (where $d_k$ is the key dimension) within a range of $[0.4, 10]$ can produce high-quality results, with varying levels of fidelity to the input modalities.
By scaling the temperature during sampling, we achieve precise control over the super-resolution process, enabling manageable manipulation of the output with fine-grained control.

\begin{figure*}[!t]
    \centering
    \vspace{-3pt}
    \def\xwidth{0.13\linewidth}
    \def\ywidth{0.12\linewidth}
    \setlength{\tabcolsep}{1pt}
    \renewcommand\arraystretch{0.6}
    \begin{tabular}[t]{c c c c c c c c}\\
    \multicolumn{3}{c}{\cellcolor{tabfirst} \scriptsize Inputs} & & \multicolumn{3}{c}{\cellcolor{tabsecond} \scriptsize Outputs} & \cellcolor{tabthird} \scriptsize Reference \\

    \cellcolor{tabfirst} \redrectangleee{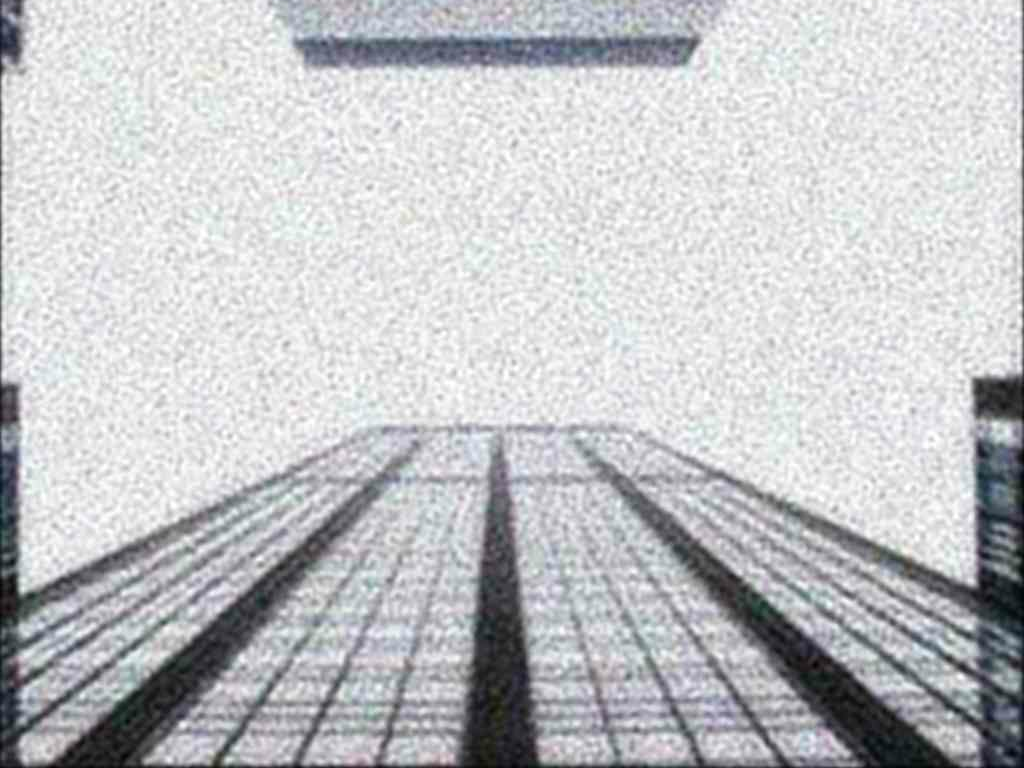}{\xwidth}{0.3cm}{0.4}{-0.43} &
    \cellcolor{tabfirst} \textshape{\xwidth}{Two skyscrapers converge towards the center, creating a symmetrical composition.  Perspective is from ground level  \dots} &
    \cellcolor{tabfirst} \redrectangleee{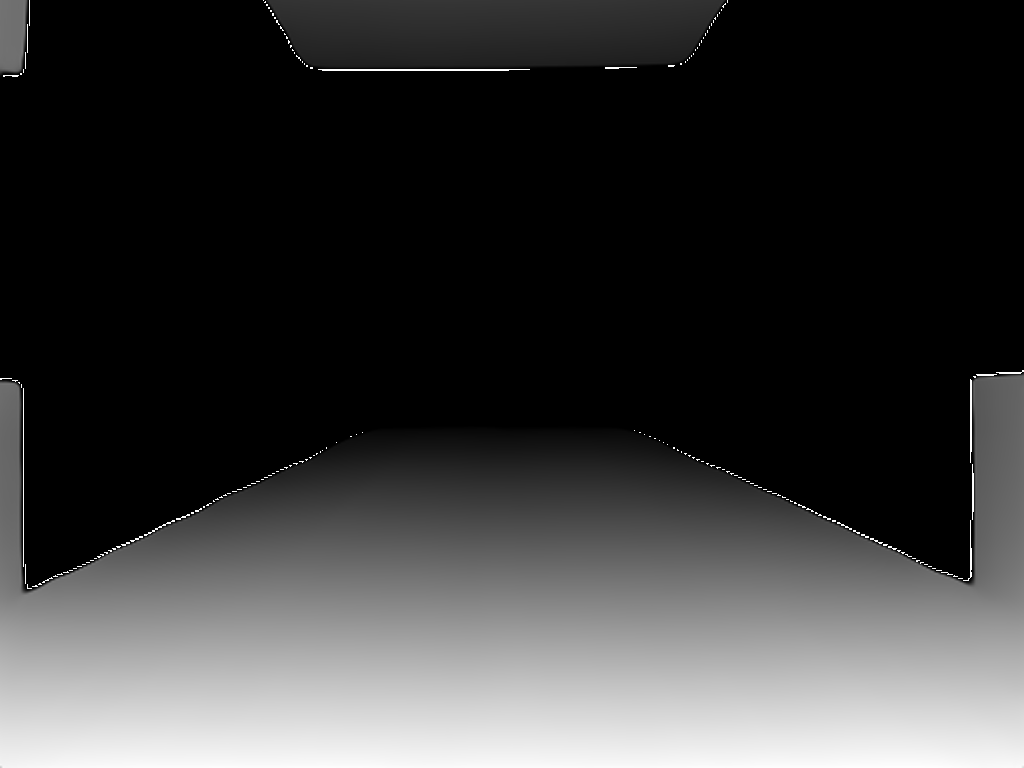}{\xwidth}{0.3cm}{0.4}{-0.43} & &
    \cellcolor{tabsecond} \redrectangleee{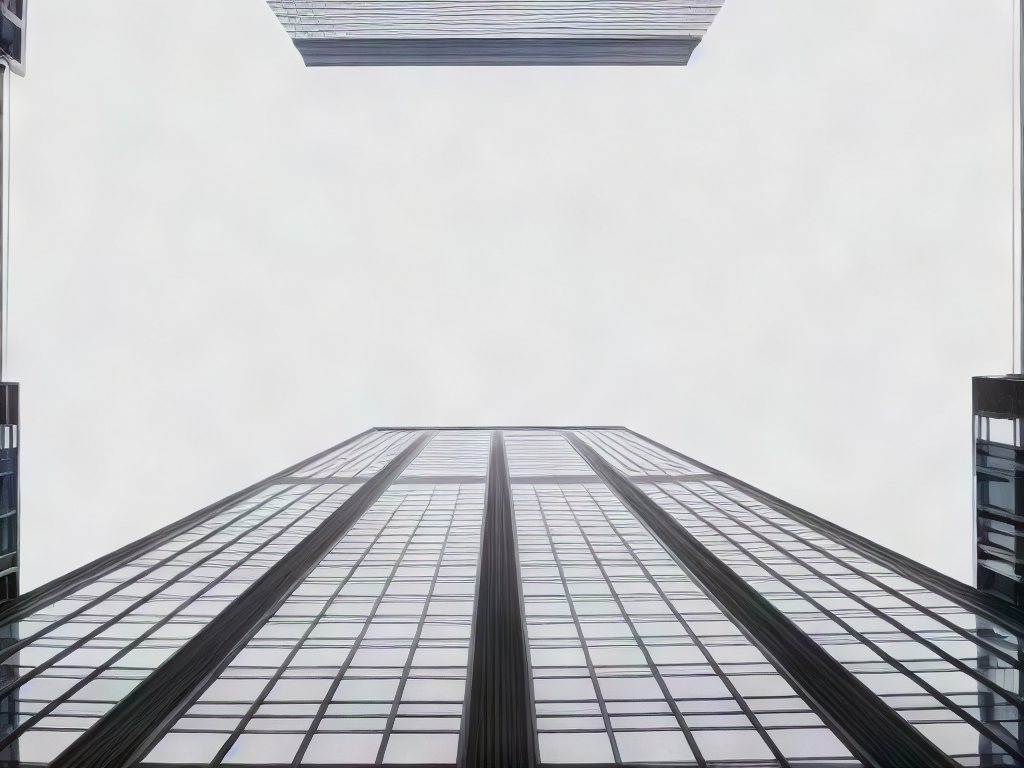}{\xwidth}{0.3cm}{0.4}{-0.43} &
    \cellcolor{tabsecond} \redrectangleee{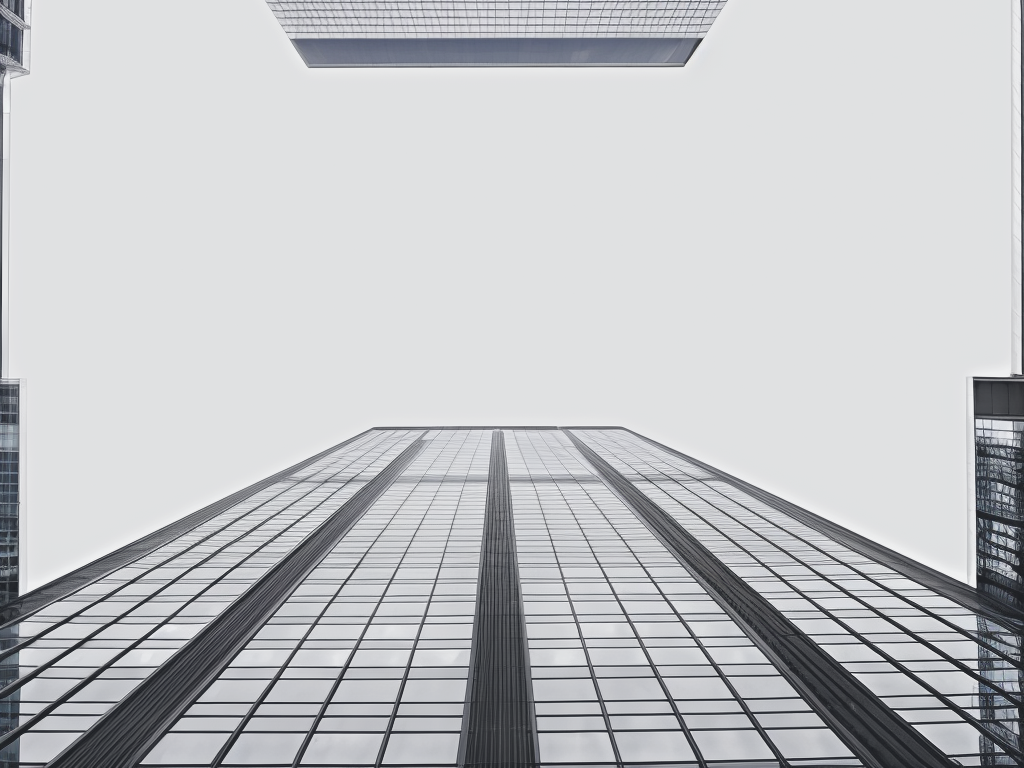}{\xwidth}{0.3cm}{0.4}{-0.43} &
    \cellcolor{tabsecond} \redrectangleee{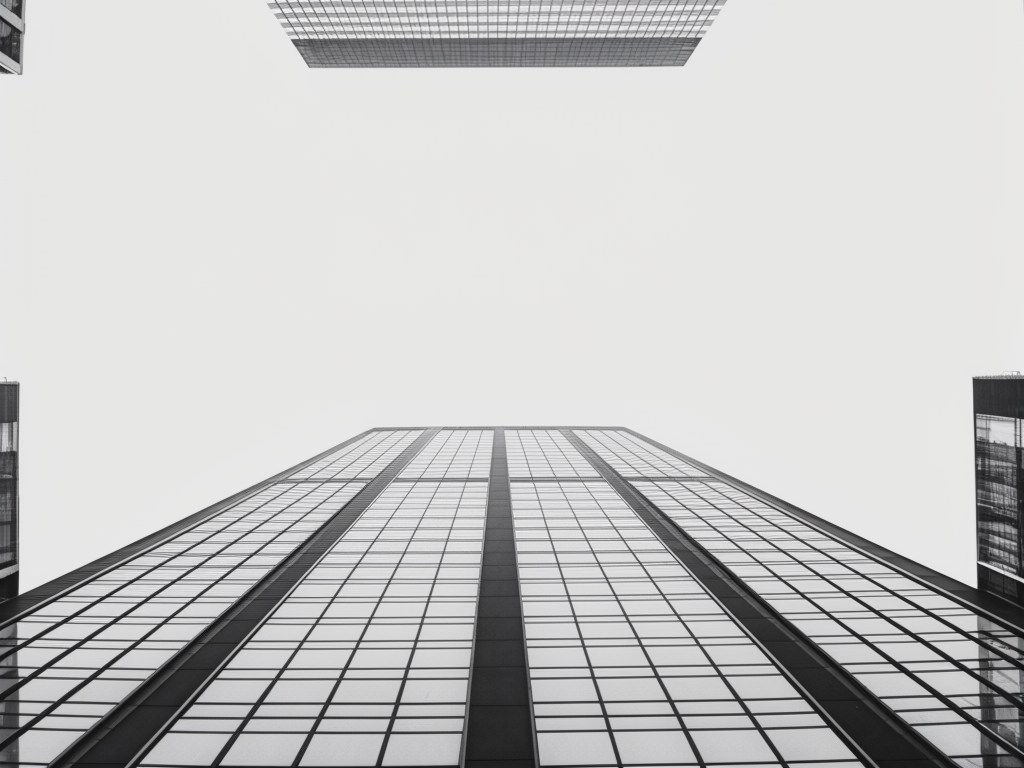}{\xwidth}{0.3cm}{0.4}{-0.43} &
    \cellcolor{tabthird} \redrectangleee{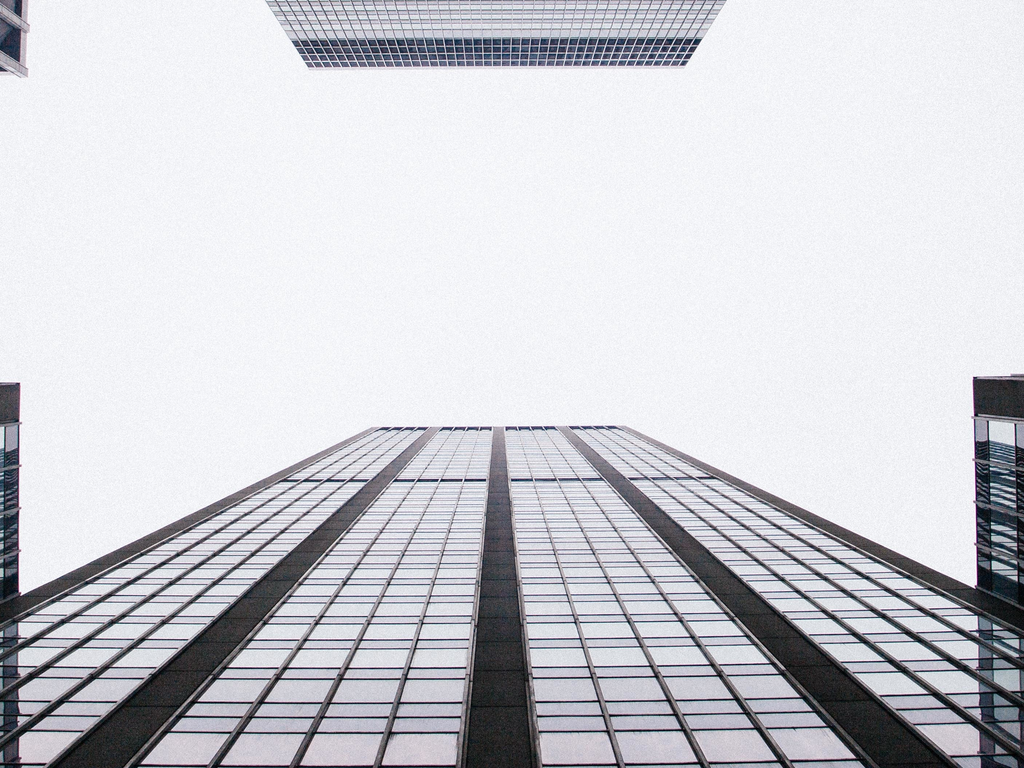}{\xwidth}{0.3cm}{0.4}{-0.43} \\
    \cellcolor{tabfirst} \scriptsize LR & \cellcolor{tabfirst} \scriptsize Caption & \cellcolor{tabfirst} \scriptsize Depth & & \cellcolor{tabsecond} \scriptsize PASD & \cellcolor{tabsecond} \scriptsize SeeSR  &  \cellcolor{tabsecond} \scriptsize MMSR (Ours) & \cellcolor{tabthird} \scriptsize HR \\
    \cellcolor{tabfirst} \settowidth{\imagewidth}{\includegraphics{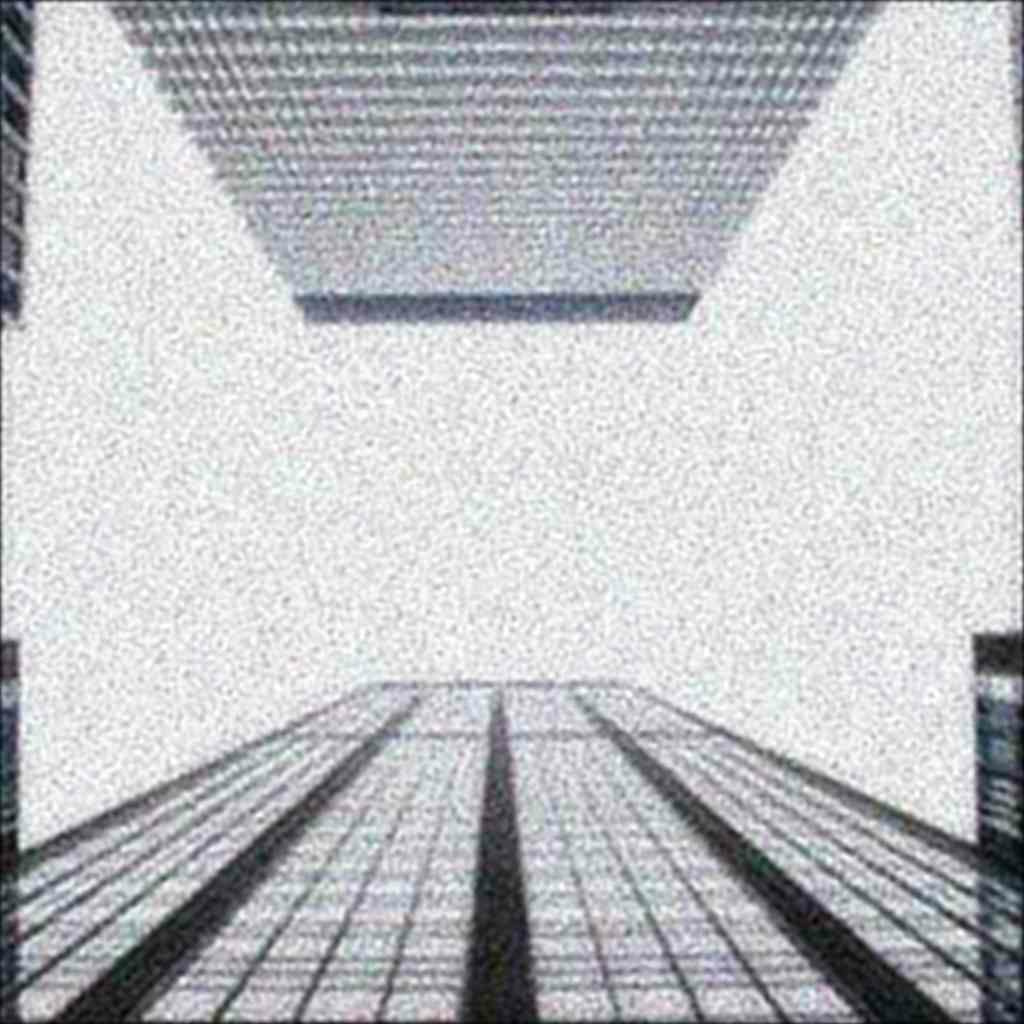}}
    \includegraphics[width=\xwidth, clip=true, trim = 0.4\imagewidth{} 0.2\imagewidth{} 0.44\imagewidth{} 0.68\imagewidth{}]{ibis-results/000244_lr.jpg} &
    \cellcolor{tabfirst}
    \redrectangleee{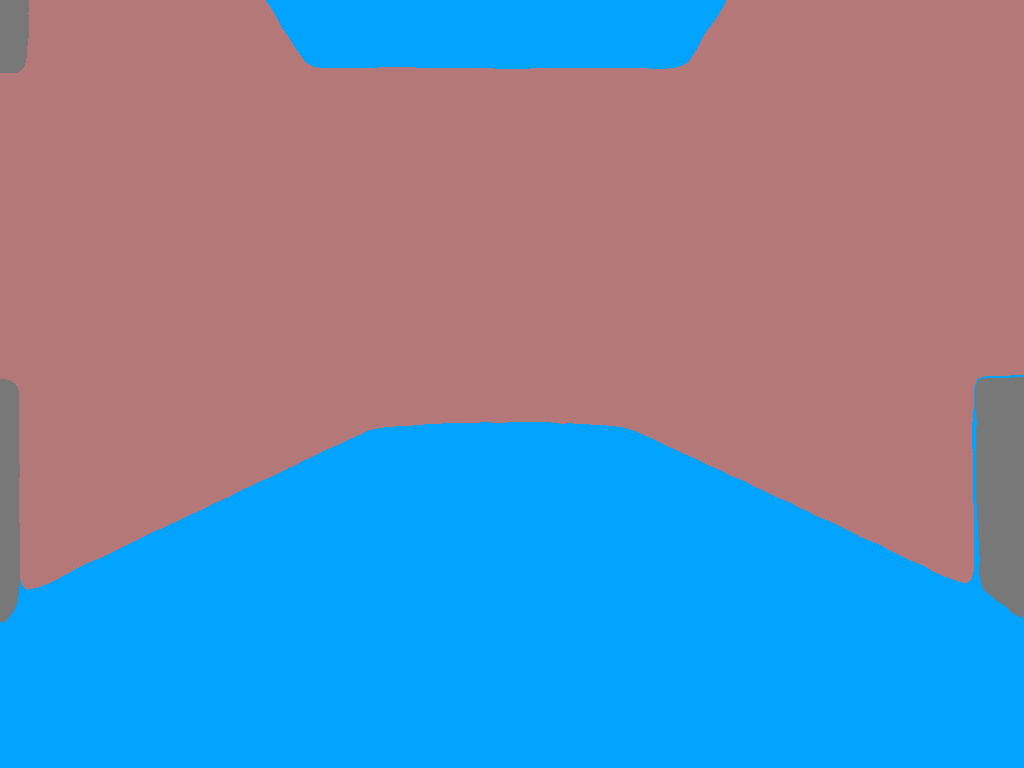}{\xwidth}{0.3cm}{0.4}{-0.43} &
    \cellcolor{tabfirst}
    \redrectangleee{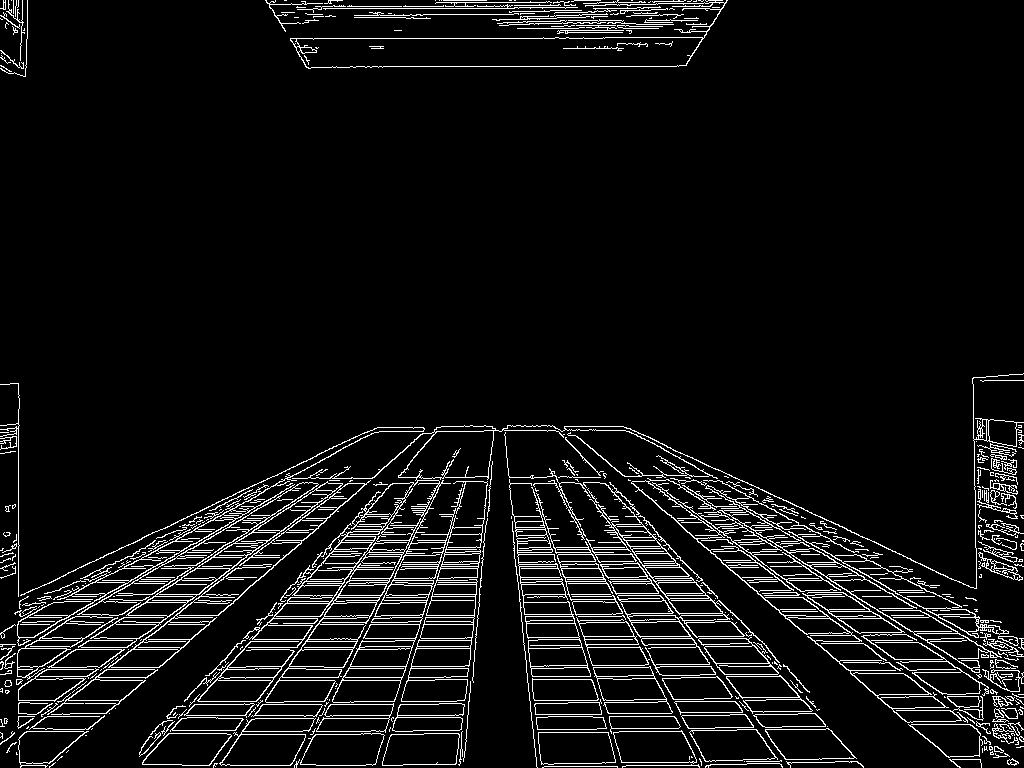}{\xwidth}{0.3cm}{0.4}{-0.43} & & 
    \cellcolor{tabsecond} \settowidth{\imagewidth}{\includegraphics{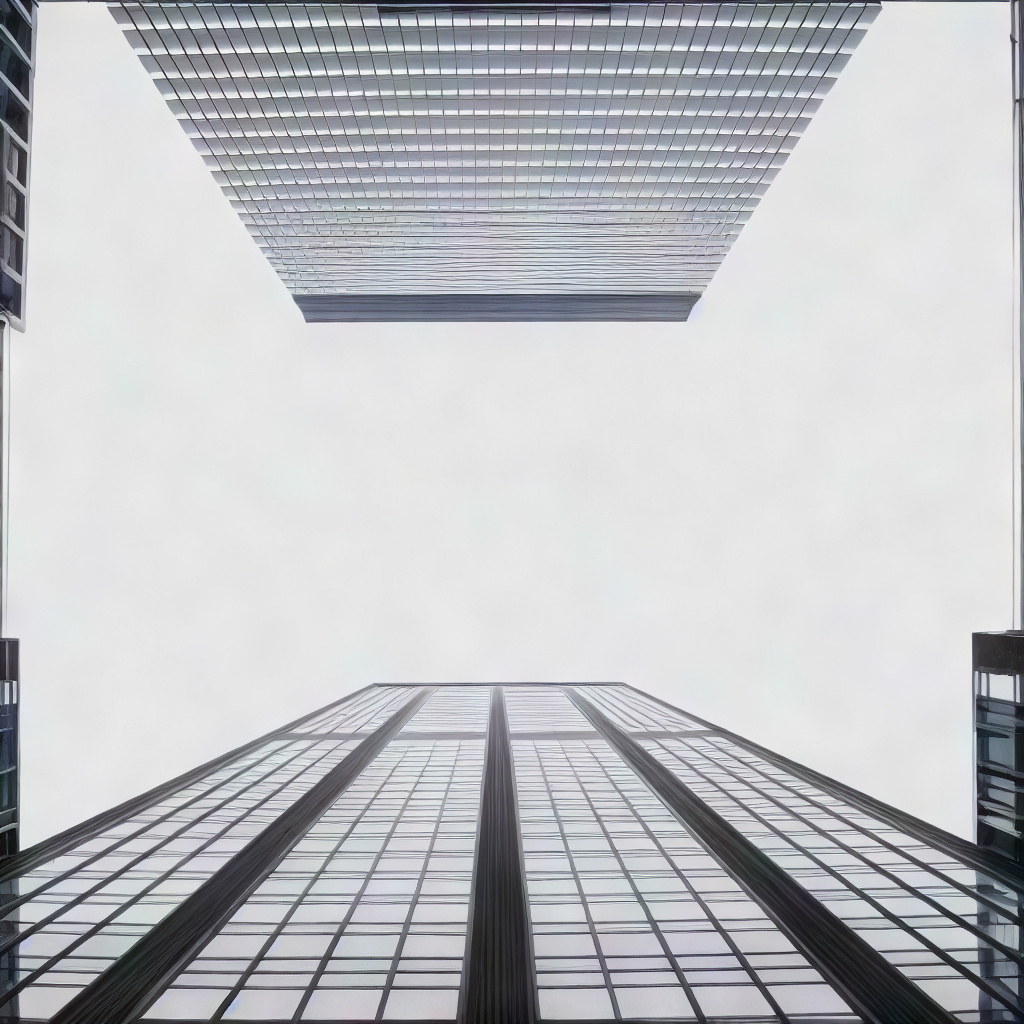}} \includegraphics[width=\xwidth, clip=true, trim = 0.4\imagewidth{} 0.2\imagewidth{} 0.44\imagewidth{} 0.68\imagewidth{}]{ibis-results/000244_pasd.jpg} &
    \cellcolor{tabsecond}  
    \settowidth{\imagewidth}{\includegraphics{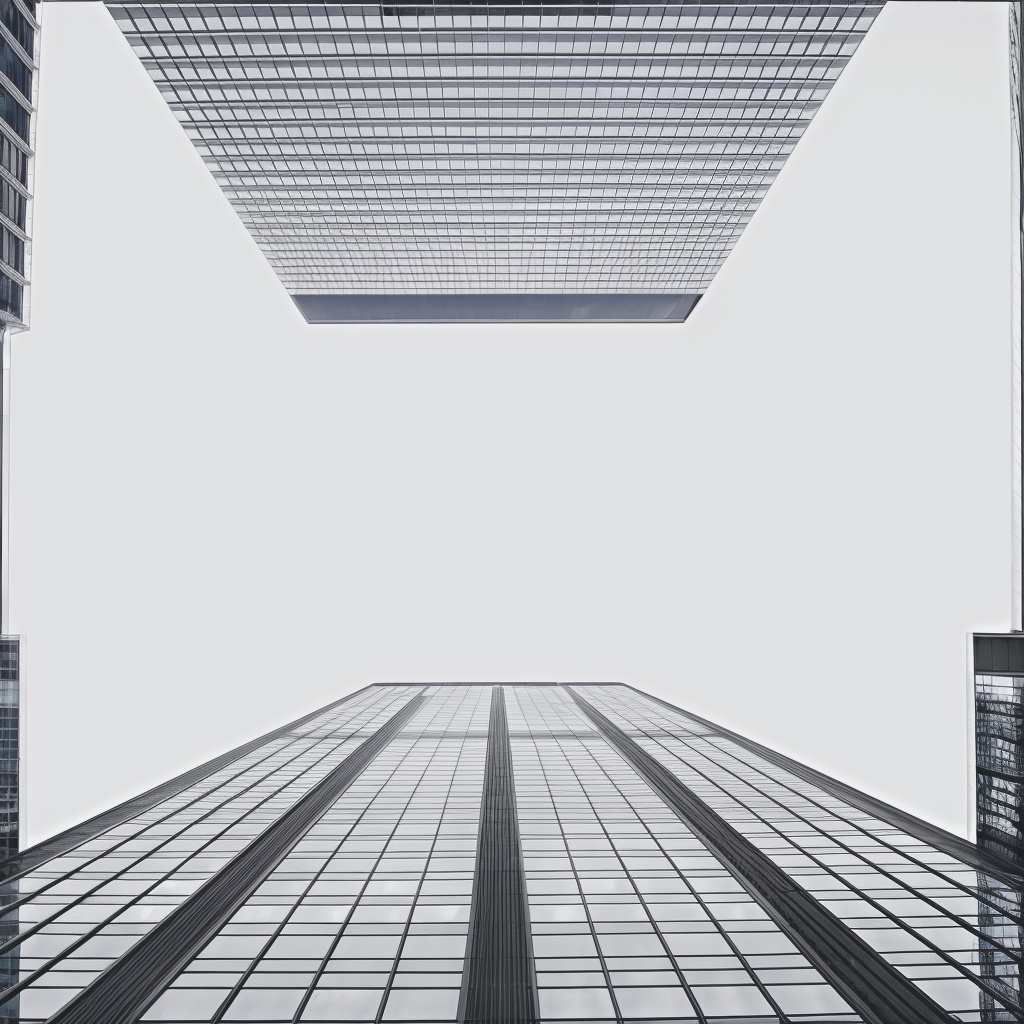}}
    \includegraphics[width=\xwidth, clip=true, trim = 0.4\imagewidth{} 0.2\imagewidth{} 0.44\imagewidth{} 0.68\imagewidth{}]{ibis-results/000244_seesr.jpg} &
    \cellcolor{tabsecond}  
    \settowidth{\imagewidth}{\includegraphics{ibis-results/000244_pasd.jpg}}
    \includegraphics[width=\xwidth, clip=true, trim = 0.4\imagewidth{} 0.2\imagewidth{} 0.44\imagewidth{} 0.68\imagewidth{}]{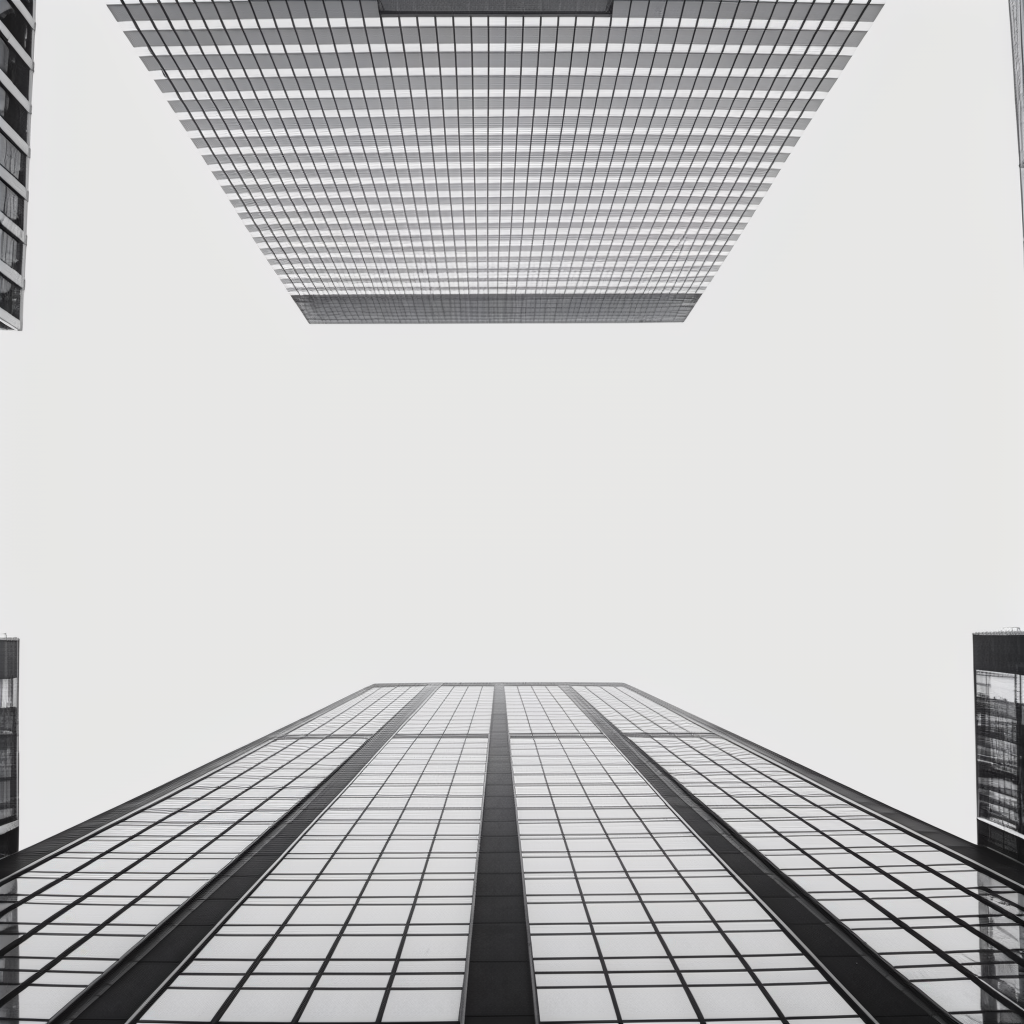} &
    \cellcolor{tabthird}  
    \settowidth{\imagewidth}{\includegraphics{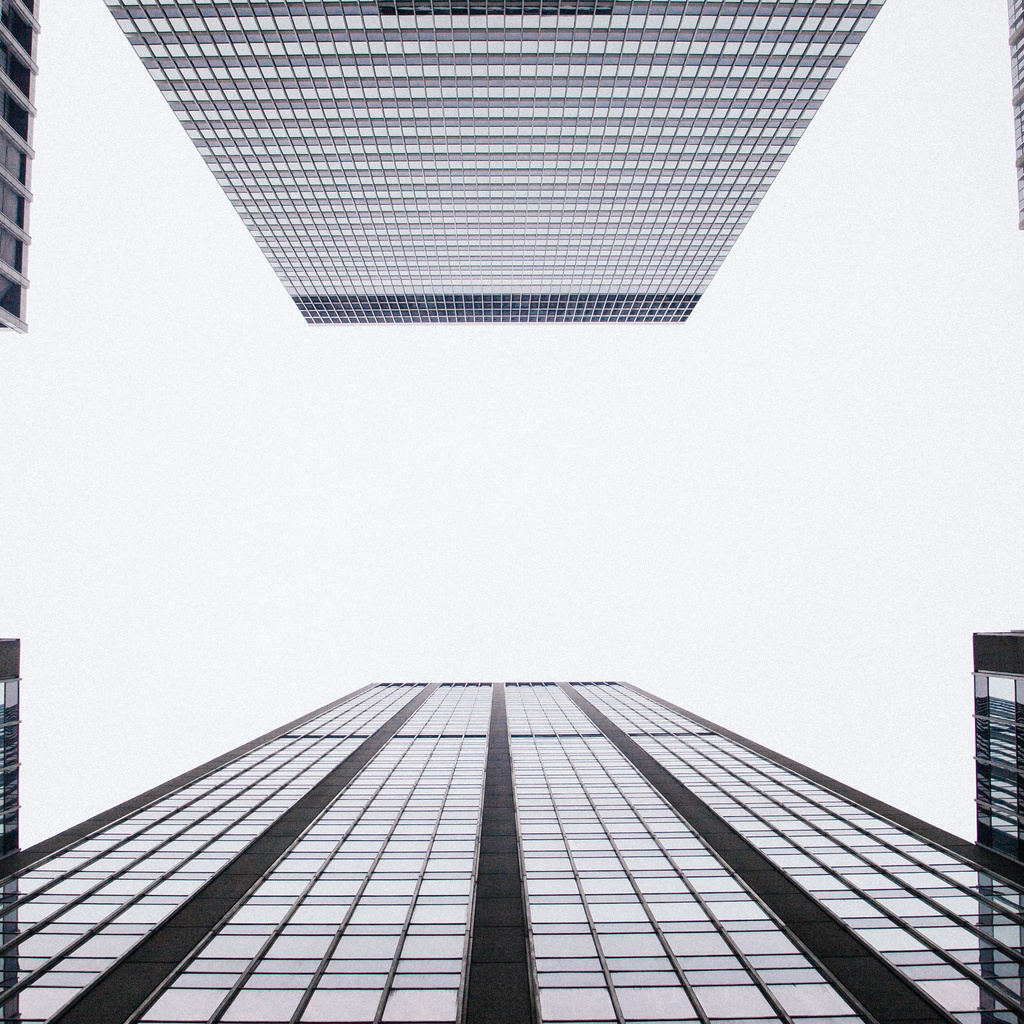}}
    \includegraphics[width=\xwidth, clip=true, trim = 0.4\imagewidth{} 0.2\imagewidth{} 0.44\imagewidth{} 0.68\imagewidth{}]{ibis-results/000244_hr.jpg}
     \\
    \cellcolor{tabfirst} \scriptsize LR (Zoomed) & \cellcolor{tabfirst} \scriptsize Segmentation &  \cellcolor{tabfirst} \scriptsize Edge & & \cellcolor{tabsecond} \scriptsize PASD (Zoomed) & \cellcolor{tabsecond} \scriptsize SeeSR (Zoomed) &  \cellcolor{tabsecond} \scriptsize MMSR (Zoomed) & \cellcolor{tabthird} \scriptsize HR (Zoomed) \\
    \\
    
    \cellcolor{tabfirst} \redrectangleee{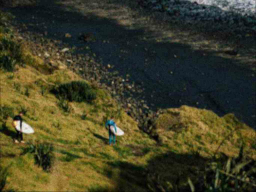}{\xwidth}{0.4cm}{0.32}{-0.43} &
    \cellcolor{tabfirst} \textshape{\xwidth}{A person wearing a blue hooded jacket and white pants carries a white surfboard down a grassy hill towards a dark sandy beach \dots} &
    \cellcolor{tabfirst} \redrectangleee{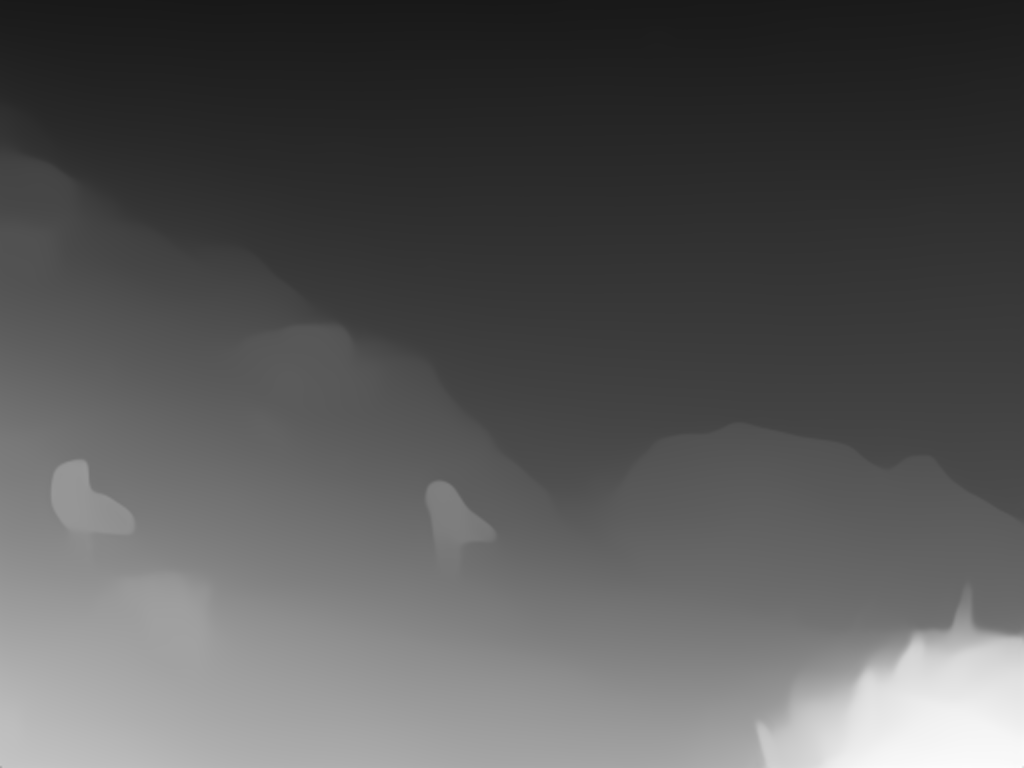}{\xwidth}{0.4cm}{0.32}{-0.43} & &
    \cellcolor{tabsecond} \redrectangleee{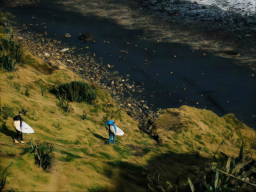}{\xwidth}{0.4cm}{0.32}{-0.43} &
    \cellcolor{tabsecond} \redrectangleee{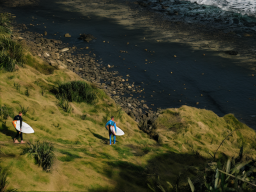}{\xwidth}{0.4cm}{0.32}{-0.43} &
    \cellcolor{tabsecond} \redrectangleee{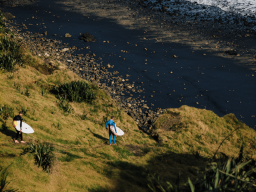}{\xwidth}{0.4cm}{0.32}{-0.43} &
    \cellcolor{tabthird} \redrectangleee{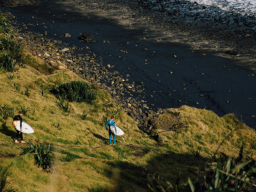}{\xwidth}{0.4cm}{0.32}{-0.43} \\
    \cellcolor{tabfirst} \scriptsize LR & \cellcolor{tabfirst} \scriptsize Caption & \cellcolor{tabfirst} \scriptsize Depth & & \cellcolor{tabsecond} \scriptsize PASD & \cellcolor{tabsecond} \scriptsize SeeSR  &  \cellcolor{tabsecond} \scriptsize MMSR (Ours) & \cellcolor{tabthird} \scriptsize HR \\
    \cellcolor{tabfirst} \settowidth{\imagewidth}{\includegraphics{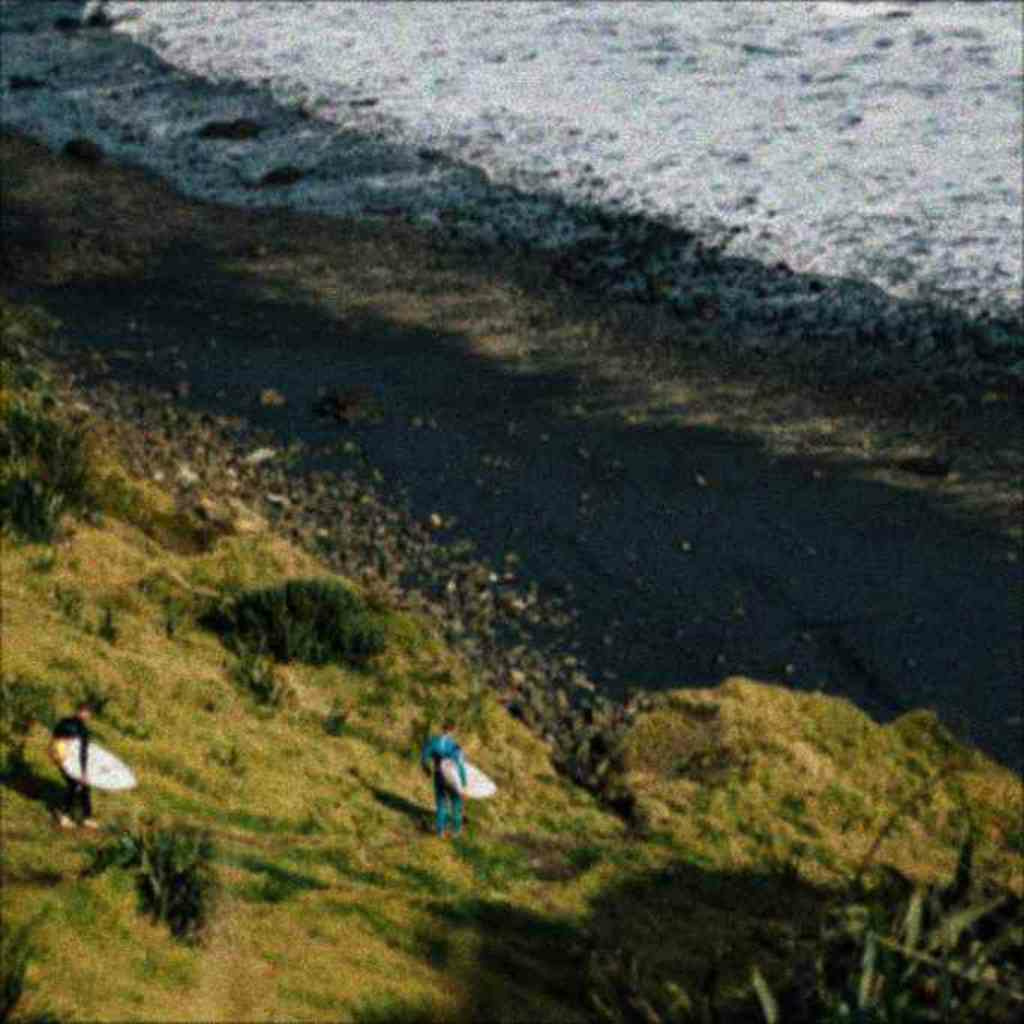}} 
    \includegraphics[width=\xwidth, clip=true, trim = 0.32\imagewidth{} 0.16\imagewidth{} 0.44\imagewidth{} 0.66\imagewidth{}]{ibis-results/000710_lr.jpg} &
    \cellcolor{tabfirst} \whiterectangleee{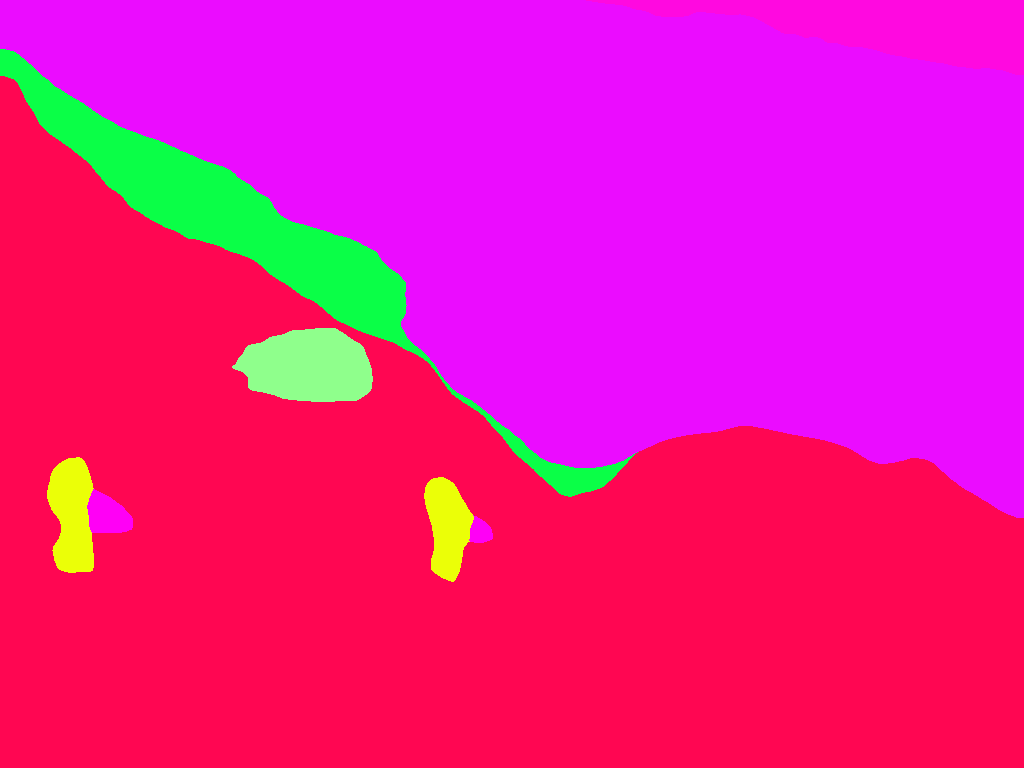}{\xwidth}{0.4cm}{0.32}{-0.43} &
    \cellcolor{tabfirst} \redrectangleee{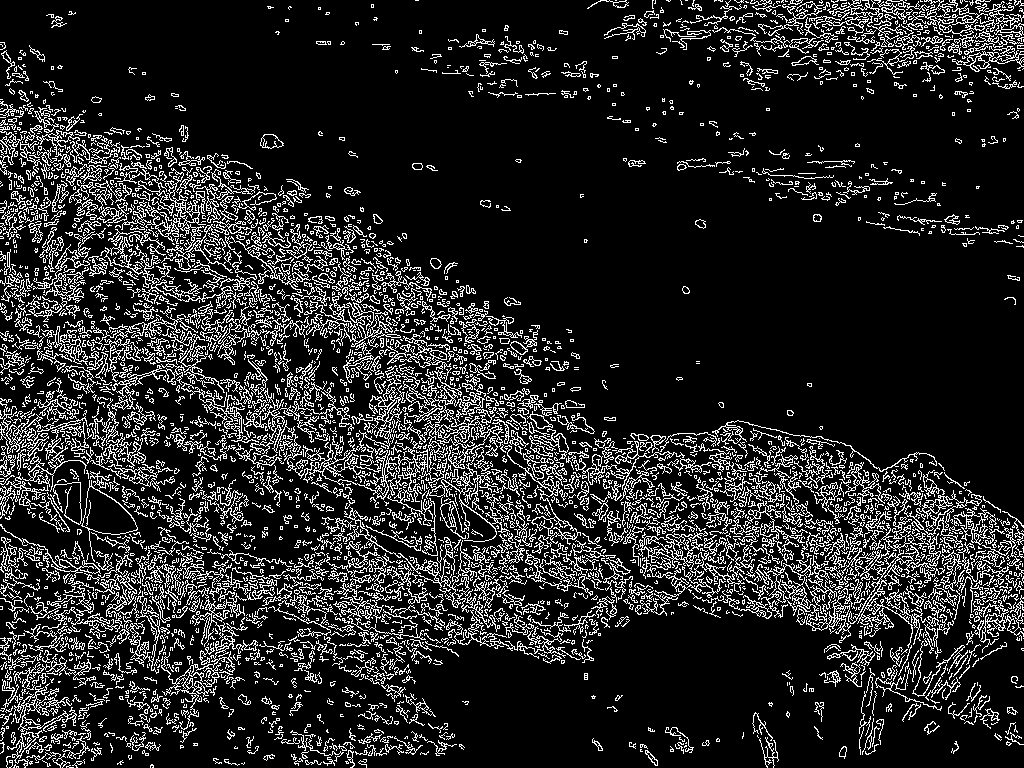}{\xwidth}{0.4cm}{0.32}{-0.43} & & 
    \cellcolor{tabsecond} \settowidth{\imagewidth}{\includegraphics{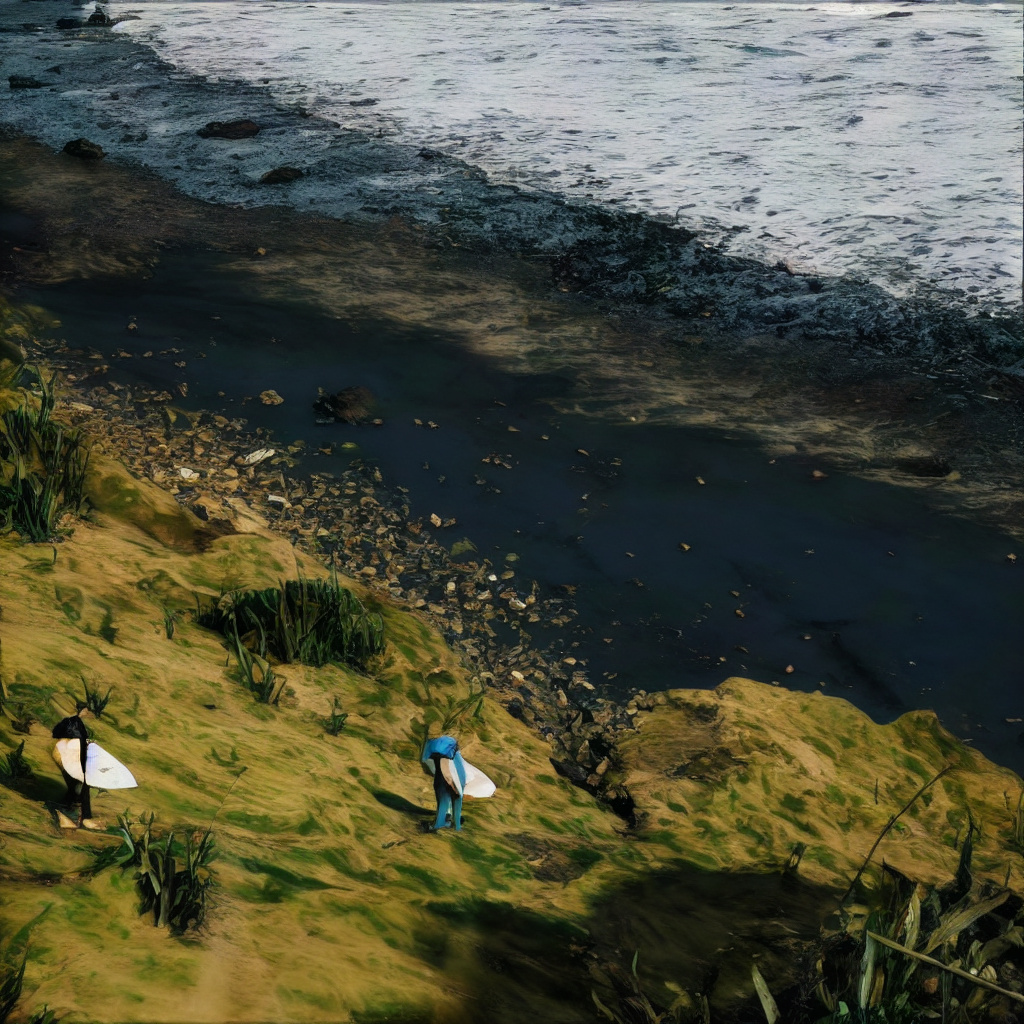}} \includegraphics[width=\xwidth, clip=true, trim = 0.32\imagewidth{} 0.16\imagewidth{} 0.44\imagewidth{} 0.66\imagewidth{}]{ibis-results/000710_pasd.jpg} &
    \cellcolor{tabsecond}  
    \settowidth{\imagewidth}{\includegraphics{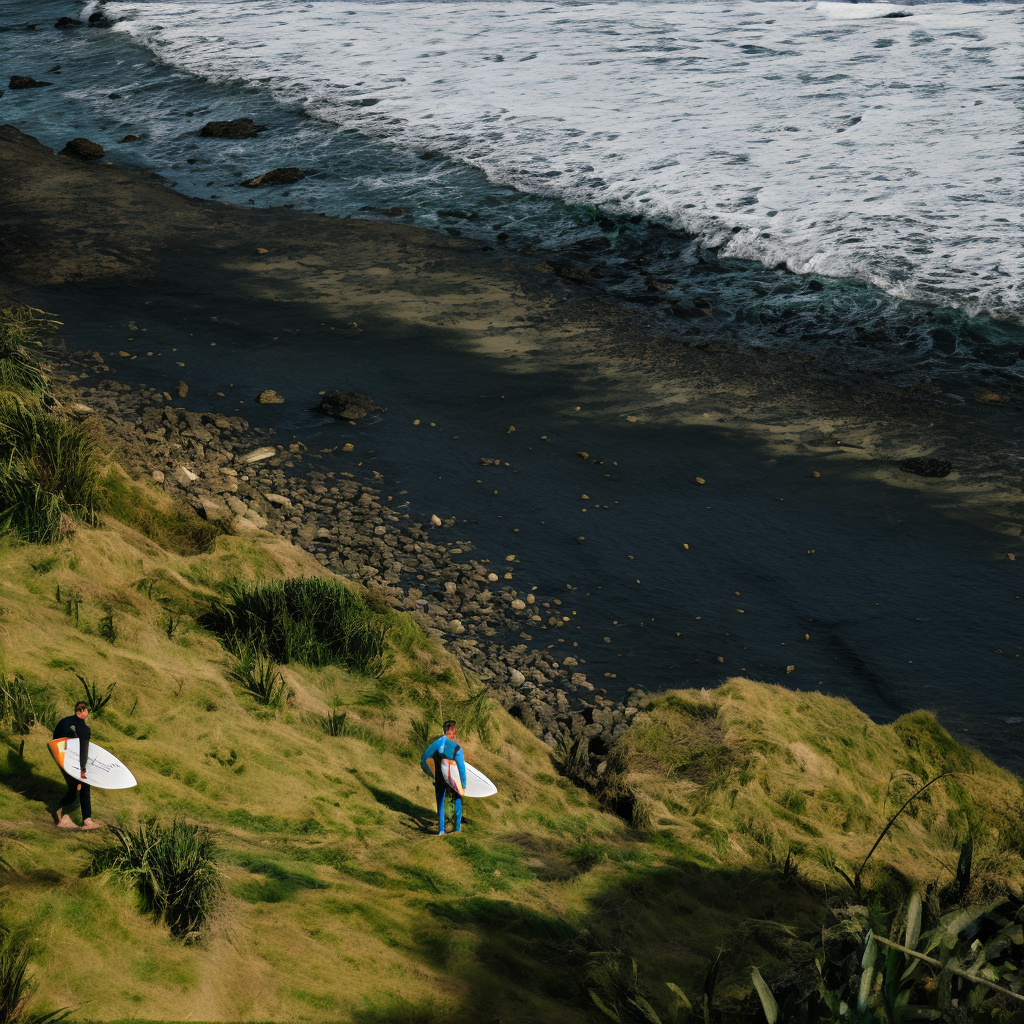}}
    \includegraphics[width=\xwidth, clip=true, trim = 0.32\imagewidth{} 0.16\imagewidth{} 0.44\imagewidth{} 0.66\imagewidth{}]{ibis-results/000710_seesr.jpg} &
    \cellcolor{tabsecond}  
    \settowidth{\imagewidth}{\includegraphics{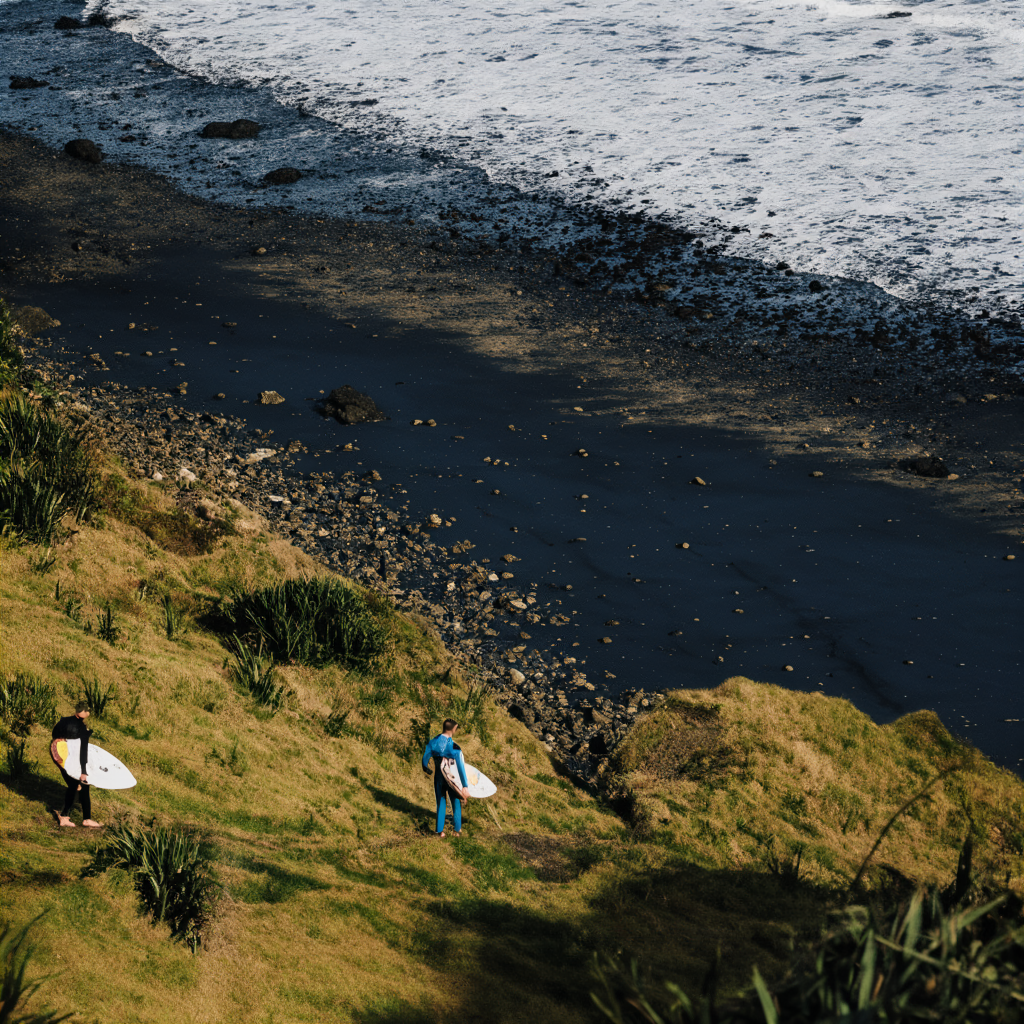}}
    \includegraphics[width=\xwidth, clip=true, trim = 0.32\imagewidth{} 0.16\imagewidth{} 0.44\imagewidth{} 0.66\imagewidth{}]{ibis-results/000710_ours.png} &
    \cellcolor{tabthird}  
    \settowidth{\imagewidth}{\includegraphics{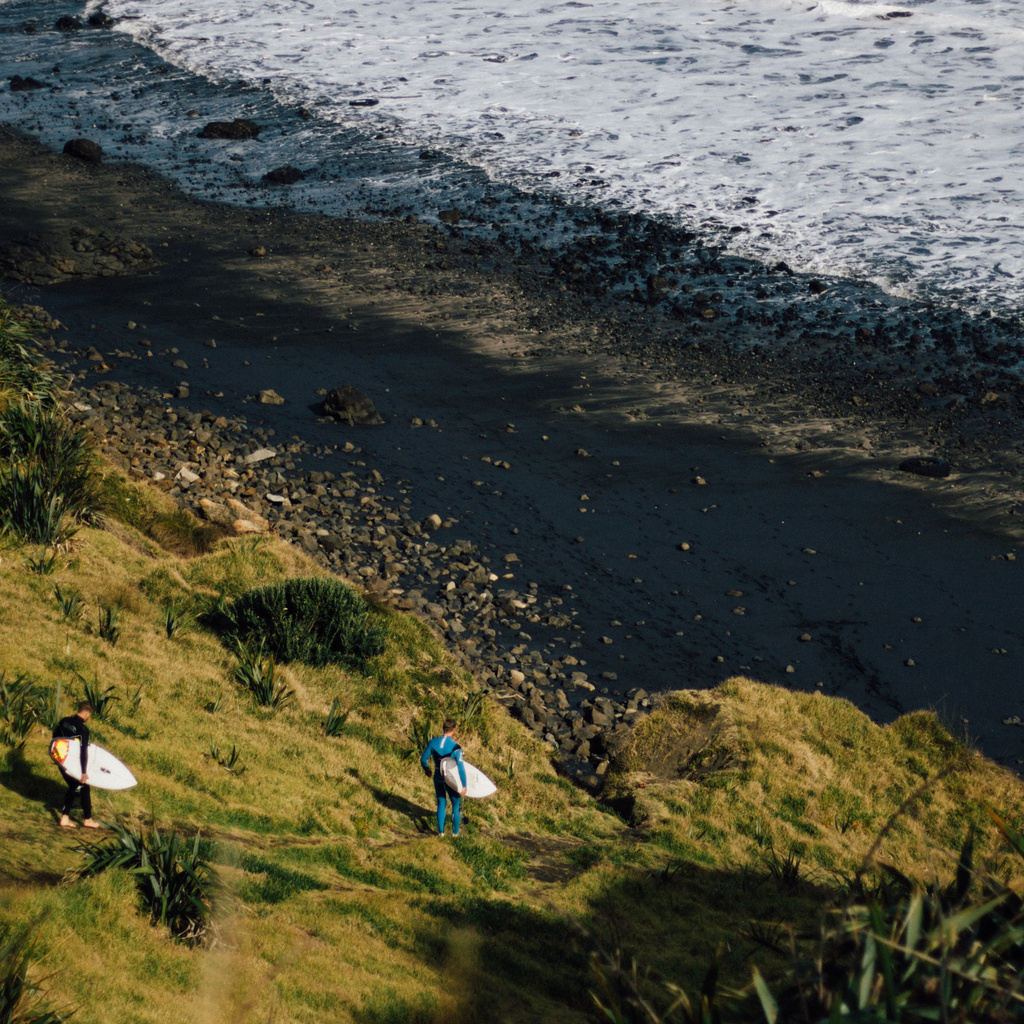}}
    \includegraphics[width=\xwidth, clip=true, trim = 0.32\imagewidth{} 0.16\imagewidth{} 0.44\imagewidth{} 0.66\imagewidth{}]{ibis-results/000710_hr.jpg}
     \\
    \cellcolor{tabfirst} \scriptsize LR (Zoomed) & \cellcolor{tabfirst} \scriptsize Segmentation &  \cellcolor{tabfirst} \scriptsize Edge & & \cellcolor{tabsecond} \scriptsize PASD (Zoomed) & \cellcolor{tabsecond} \scriptsize SeeSR (Zoomed) &  \cellcolor{tabsecond} \scriptsize MMSR (Zoomed) & \cellcolor{tabthird} \scriptsize HR (Zoomed) \\
    \\
    
    \end{tabular}
    \vspace{-1\baselineskip}
    \caption{
    Super-resolution results on common benchmarks, comparing MMSR with state-of-the-art techniques. Zoom in for detail.
    }
    \label{fig:srresults}
    \vspace{-1\baselineskip}
\end{figure*}

\subsection{MMSR Implementation}
Figure~\ref{fig:ibiarch} illustrates our multimodal guided super-resolution pipeline. During inference, we first extract four modalities from the low-resolution (LR) image: a text caption generated by Gemini Flash~\cite{team2023gemini}, a depth map estimated by Depth Anything~\cite{yang2024depth}, a semantic segmentation mask produced by Mask2Former~\cite{cheng2022masked}, and edge information extracted with a Canny edge detector. The depth map, segmentation mask, and edge information are encoded into token sequences using a pretrained VQGAN, while the text caption is processed by a pretrained CLIP encoder to obtain a text embedding.

Following a similar conditioning strategy to Instruct-Pix2Pix~\cite{brooks2023instructpix2pix}, the LR image is concatenated with a noisy latent vector sampled from the diffusion model, providing additional conditioning.

\section{Experiments}

\vspace{.5em}
\noindent \textbf{Training Details.}
Our super-resolution model is initialized with the weights of a pretrained text-to-image model, with the same architecture and size as Stable Diffusion v2~\cite{rombach2022high}.
The super-resolution training dataset consists of randomly degraded (using RealESRGAN degradation~\cite{wang2022realesrgan}) high-resolution images from the combined LSDIR and DIV2K datasets, with corresponding $512\times512$ high-resolution images as ground truth. 
We set the batch size to 1024 and the learning rate to 1e-4, which we empirically found maximized compute efficiency on TPUv5 pods.
During testing, all benchmarks use a model checkpoint finetuned for 160k iterations. We use 50-step DDIM sampling, consistent with previous methods.
We use a guidance rate of 4 as the default.

\vspace{.5em}
\noindent \textbf{Evaluation Details.}
We use reference-based metrics like LPIPS~\cite{zhang2018unreasonable} and DISTS~\cite{ding2021locally}, and non-reference-based metrics like NIQE~\cite{mittal2012making}, MANIQA~\cite{yang2022maniqa}, MUSIQ~\cite{ke2021musiq}, and CLIPIQA~\cite{wang2022exploring}, as these have been shown to align well with humans' aesthetic preferences.
Our compared baselines include recent diffusion-based super-resolution methods~\cite{wu2024seesr, yang2023pixel, lin2023diffbir, yue2024resshift, wang2024exploiting, rombach2022high}
and the representative methods~\cite{wang2022realesrgan, zhang2021designing,liang2022details,liang2022efficient,chen2022real}.
We collected their results either by running their official code or from published results.
In a few cases where the multimodal prediction failed (two flat images and one pencil drawing from DRealSR), we manually replaced the prediction with $m_\emptyset$ (see supplemental material).

\subsection{Quantitative Results}

\begin{table}[!t]
\centering
\small
\setlength{\tabcolsep}{2pt}
\renewcommand{\arraystretch}{1.2}
\resizebox{\linewidth}{!}
{%
\begin{tabular}{lcccccccc}
& \multicolumn{8}{l}{\emph{DIV2K-Val-3k} 512 $\times$ 512} \\
\cline{2-9}
Methods & PSNR & SSIM & LPIPS $\downarrow$ & DISTS $\downarrow$ & NIQE $\downarrow$ & FID $\downarrow$ & MUSIQ & CLIPIQA \\
\midrule
BSRGAN & \textcolor{blue}{\underline{21.87}} & 0.5539 & 0.4136 & 0.2737 & 4.7615 & 64.28 & 59.11 & 0.5183 \\
 R-ESRGAN & \textcolor{red}{\textbf{21.94}} & \textcolor{red}{\textbf{0.5736}} & 0.3868 & 0.2601 & 4.9209 & 53.46 & 58.64 & 0.5424 \\
LDL & 21.52 & 0.5690 & 0.3995 & 0.2688 & 5.0249 & 58.94 & 57.90 & 0.5313 \\
 DASR & 21.72 & 0.5536 & 0.4266 & 0.2688 & 4.8596 & 67.22 & 54.22 & 0.5241 \\
FeMASR & 20.85 & 0.5163 & 0.3973 & 0.2428 & \tbd{4.5726} & 53.70 & 58.10 & 0.5597 \\
 LDM & 21.26 & 0.5239 & 0.4154 & 0.2500 & 6.4667 & 41.93 & 56.52 & 0.5695 \\
StableSR & 20.84 & 0.4887 & 0.4055 & 0.2542 & 4.6551 & 36.57 & 62.95 & 0.6486 \\
 ResShift & 21.75 & 0.5422 & 0.4284 & 0.2606 & 6.9731 & 55.77 & 58.23 & 0.5948 \\
PASD & 20.77 & 0.4958 & 0.4410 & 0.2538 & 4.8328 & 40.77 & 66.85 & 0.6799 \\
 DiffBIR & 20.94 & 0.4938 & 0.4270 & 0.2471 & 4.7211 & 40.42 & 65.23 & 0.6664 \\
SeeSR & 21.19 & 0.5386 & \tbd{0.3843} & \tbd{0.2257} & 4.9275 & \tbd{31.93} & \tbd{68.33} & \tbd{0.6946} \\
 \rowcolor{lightgray2} \textbf{MMSR} & 21.74 & \tbd{0.5693} & \trf{0.3707} & \trf{0.2071} & \trf{4.2532} & \trf{29.35}  & \trf{70.06} & \trf{0.7164} \\
\midrule
\end{tabular}
}

\hfill

\vspace{.2\baselineskip}
\resizebox{\linewidth}{!}
{%
\begin{tabular}{lcccccccc}
& \multicolumn{8}{l}{\emph{DIV2K-Val-100} 1024 $\times$ 1024} \\
\cline{2-9}
Methods & PSNR & SSIM & LPIPS $\downarrow$ & DISTS $\downarrow$ & NIQE $\downarrow$ & MANIQA & MUSIQ & CLIPIQA \\
\midrule
 R-ESRGAN & \tbd{21.77} & \trf{0.5813} & 0.3624 & 0.1990 & 3.6573 & 0.4046 & 47.54 & 0.5358 \\
StableSR & 20.07 & 0.3947 & 0.5097 & 0.2427 & 3.6260 & 0.4113 & 65.39 & 0.6938  \\
PASD & 21.27 & 0.5369 & 0.3473 & 0.1753 & \tbd{3.6321} & 0.4708 & 69.69 & 0.6914 \\
SUPIR & 20.65 & 0.5350 & 0.3849 & 0.1814 & 3.6458 & 0.4051 & 65.88 & 0.5697  \\
SeeSR & 21.31 & \tbd{0.5578} & \tbd{0.3273} & \tbd{0.1620} & 4.0215 & \trf{0.5439} & \tbd{69.79} & \tbd{0.6941}  \\
\rowcolor{lightgray2} \textbf{MMSR} & \trf{21.87} & 0.5565 & \trf{0.2810} & \trf{0.1492} & \trf{3.4243} & \tbd{0.4885} & \trf{72.31} & \trf{0.7294}   \\
\midrule
\end{tabular}
}

\hfill

\vspace{.2\baselineskip}
\resizebox{\linewidth}{!}
{%
\begin{tabular}{lcccccccc}
& \multicolumn{8}{l}{\emph{RealSR} 512 $\times$ 512} \\
\cline{2-9}
Methods & PSNR & SSIM & LPIPS $\downarrow$ & LIQE & NIMA & MANIQA & MUSIQ & CLIPIQA \\
\midrule
 R-ESRGAN & \trf{25.69} & \trf{0.7616} & \trf{0.2727} & 3.3574 & 4.6548 & 0.5487 & 60.18 & 0.4449 \\
StableSR & 24.70 & 0.7085 & 0.3018 & 3.6106 & 4.8150 & 0.6221 & 65.78 & 0.6178 \\
PASD & 24.29 & 0.6630 & 0.3435 & 3.5749 & 4.8554 & 0.6493 & 68.69 & 0.6590 \\
SUPIR & 22.97 & 0.6298 & 0.3750 & 3.5682 & 4.5757 & 0.5745 & 61.49 & 0.6434  \\
SeeSR & \tbd{25.18} & \tbd{0.7216} & 0.3009 & \tbd{4.1360} & \tbd{4.9193} & \tbd{0.6442} & \tbd{69.77} & \tbd{0.6612} \\
\rowcolor{lightgray2} \textbf{MMSR} & 24.83 & 0.7003 & \tbd{0.2952} & \trf{4.3468} & \trf{5.1094} & \trf{0.6578} & \trf{71.33} & \trf{0.6717} \\
\midrule
\end{tabular}
}

\vspace{.2\baselineskip}
\resizebox{\linewidth}{!}
{%
\begin{tabular}{lcccccccc}
& \multicolumn{8}{l}{\emph{DrealSR} 512 $\times$ 512} \\
\cline{2-9}
Methods & PSNR & SSIM & LPIPS $\downarrow$ & LIQE & NIMA  & MANIQA & MUSIQ & CLIPIQA \\
\midrule
 R-ESRGAN & \trf{28.64} & \trf{0.8053} & \trf{0.2847} & 2.9255 & 4.3258 & 0.4907 & 54.18 & 0.4422 \\
StableSR & 26.71 & 0.7224 & 0.3284 & 3.2425 & 4.4861 & 0.5594 & 58.51 & 0.6357 \\
PASD & 27.00 & 0.7084 & 0.3931 & 3.5908 & 4.6618 & 0.5850 & 64.81 & 0.6773 \\
 SUPIR & 24.61 & 0.6123 & 0.4294 & 3.4710 & 4.3815 & 0.5381 & 57.32 & 0.6758 \\
 SeeSR & 26.75 & 0.7405 & \tbd{0.3174} & \tbd{4.1270} & \tbd{4.6942} & \tbd{0.6052} & \tbd{65.09} & \tbd{0.6908} \\
\rowcolor{lightgray2} \textbf{MMSR} & \tbd{27.28} & \tbd{0.7456} & 0.3249 & \trf{4.5023} & \trf{5.0558} & \trf{0.6301} & \trf{68.93} & \trf{0.6999} \\
\bottomrule
\end{tabular}
}
\vspace{-.5\baselineskip}
\caption{
Severely degraded, low-resolution (LR) images can produce inaccurate multi-modal information, manifesting as distorted edges, misidentified objects, and other artifacts. %
}
\label{tab:results}
\vspace{-1\baselineskip}
\end{table}

Table~\ref{tab:results} shows the quantitative result comparisons of ours with the other baselines, compared on the synthetic benchmark and real-world super-resolution benchmarks.
Several facts are worth noting:
(1) Our method achieves the best LPIPS, DISTS, NIQE, and FID scores on the \emph{DIV2K-Val} and \emph{RealSR} dataset, significantly outperforming previous state-of-the-art methods.
This superiority demonstrates that our method can generate more perceptually identical details from the guidance of multimodal context.
(2) Our method also achieves the best performance in the non-reference visual quality metrics including NIQE, MANIQA, CLIPIQA, and MUSIQ scores.
This advantage demonstrates our method can fully utilize the photorealistic priors encapsulated in the pretrained diffusion model.
Overall, these observations fully demonstrate our advantages in using multimodal guidance compared with the past single-modality methods.

\vspace{.4em}
\noindent \textbf{Effects of Multimodal CFG Guidance.}
\label{sec:multimodalcfg}
Table~\ref{tab:cfg} demonstrates the effects of our proposed multimodal CFG on the 1MP \emph{DIV2K-Val} benchmark, which replaces the negative score guided by empty language token $\bepsilon(\bz_t, c, \mathrm{neg})$, shortened as \emph{cfg}, with multimodal guided negative score ($\bepsilon(\bz_t, c, m, \mathrm{neg})$), shortened as \emph{$m$-cfg}.
In order to exclude the influence of architectural differences, we also compare our method with the empty multimodal token $\bepsilon(\bz_t, c, m_\emptyset, \mathrm{neg})$, shortened as \emph{$m_\emptyset$-cfg}.
Note that the positive score for all three compared methods is the same $\bepsilon(\bz_t, c, m, \mathrm{pos})$.
The results show that our method mitigates the degradation reflected in the LPIPS and NIQE scores when using guidance rates of 10 and 14 in both cfg and $m_\emptyset$-cfg, while maintaining comparable performance.
The visual results in Table~\ref{tab:cfg}  further demonstrate our superiority, where ours suppresses the color change and incorrect texture of high CFG guidance rate, leading to higher CLIPIQA score.

\vspace{.4em}
\noindent \textbf{Effects of Latent Connector.}
To demonstrate the effectiveness of the proposed multimodal latent connector, we conducted an ablation study by training a new multimodal super-resolution model without the latent connector, which directly uses the longer multimodal token sequence as input.
The quantitative performance is shown in Table~\ref{tab:abmmlc}.
Our \emph{w. MMLC} outperforms the \emph{w/o. MMLC} in all reference and non-reference metrics.
The visual results in Figure~\ref{fig:mmlcabresult} show that the model without the MMLC is more likely to exacerbate hallucination.
Its result shows bokeh on the branches but a clear trunk, even though the depth input clearly shows they are at the same depth.

\begin{table}[!ht]
    \centering
    \small
    \resizebox{\linewidth}{!}{
    \setlength{\tabcolsep}{8pt}
    \begin{tabular}{lccccc}
    \toprule
        & MUSIQ & NIQE $\downarrow$ & DISTS $\downarrow$ & LPIPS $\downarrow$ & Throughput \\
    \midrule
        \emph{w/o. MMLC} & \tbd{69.69} & \tbd{3.4845} & \tbd{0.1781}  & \tbd{ 0.3929} & 3.48 img/s   \\
        \emph{w. MMLC} & \trf{72.31} & \trf{3.4243}  & \trf{0.1492}  & \trf{0.2810} & 3.32 img/s  \\
    \bottomrule
    \end{tabular}
    }
    \vspace{-.5\baselineskip}
    \caption{Ablation of the impact of the Multimodal Latent Connector module (MMLC) on \emph{DIV2K-Val-100} 1024p.}
    \vspace{-1\baselineskip}
    \label{tab:abmmlc}
\end{table}

\begin{table}[t]
    \centering
    \small
    \setlength{\tabcolsep}{6pt}
    \renewcommand{\arraystretch}{1.3}
    \begin{subtable}[t]{.49\linewidth}
    \resizebox{\linewidth}{!}{
    \begin{tabular}{rccc}
    guidance & 2 & 10 & 14 \\
    \midrule
    \emph{cfg} & 0.3239 & \tbd{0.4491} & \tbd{0.5064} \\
    \emph{$m_\emptyset$-cfg} & \tbd{0.2815} & 0.4803 & 0.5493 \\
    \rowcolor{lightgray2} \emph{$m$-cfg} & \trf{0.2810} & \trf{0.3471} & \trf{0.3772}  \\
    \bottomrule
    \end{tabular}
    }
    \caption{LPIPS score}
    \end{subtable}
    \small
    \begin{subtable}[t]{.49\linewidth}
    \resizebox{\linewidth}{!}{
    \begin{tabular}{rccc}
    guidance & 2 & 10 & 14 \\
    \midrule
    \emph{cfg} & \tbd{3.577} & \tbd{4.6179} & \tbd{5.1886}  \\
    \emph{$m_\emptyset$-cfg} & 3.6261 & 5.1081 & 5.9175 \\
    \rowcolor{lightgray2} \emph{$m$-cfg} & \trf{3.4679} & \trf{3.7419} & \trf{3.9815}  \\
    \bottomrule
    \end{tabular}
    }
    \caption{NIQE score}
    \end{subtable}
    \hfill
    
    \begin{subtable}[t]{\linewidth}
    \def\xwidth{0.38\linewidth}
    \def\xxwidth{0.6\linewidth}
    \def\ywidth{0.12\linewidth}
    \setlength{\tabcolsep}{1pt}
    \renewcommand\arraystretch{0.6}
    \settowidth{\imagewidth}{\includegraphics{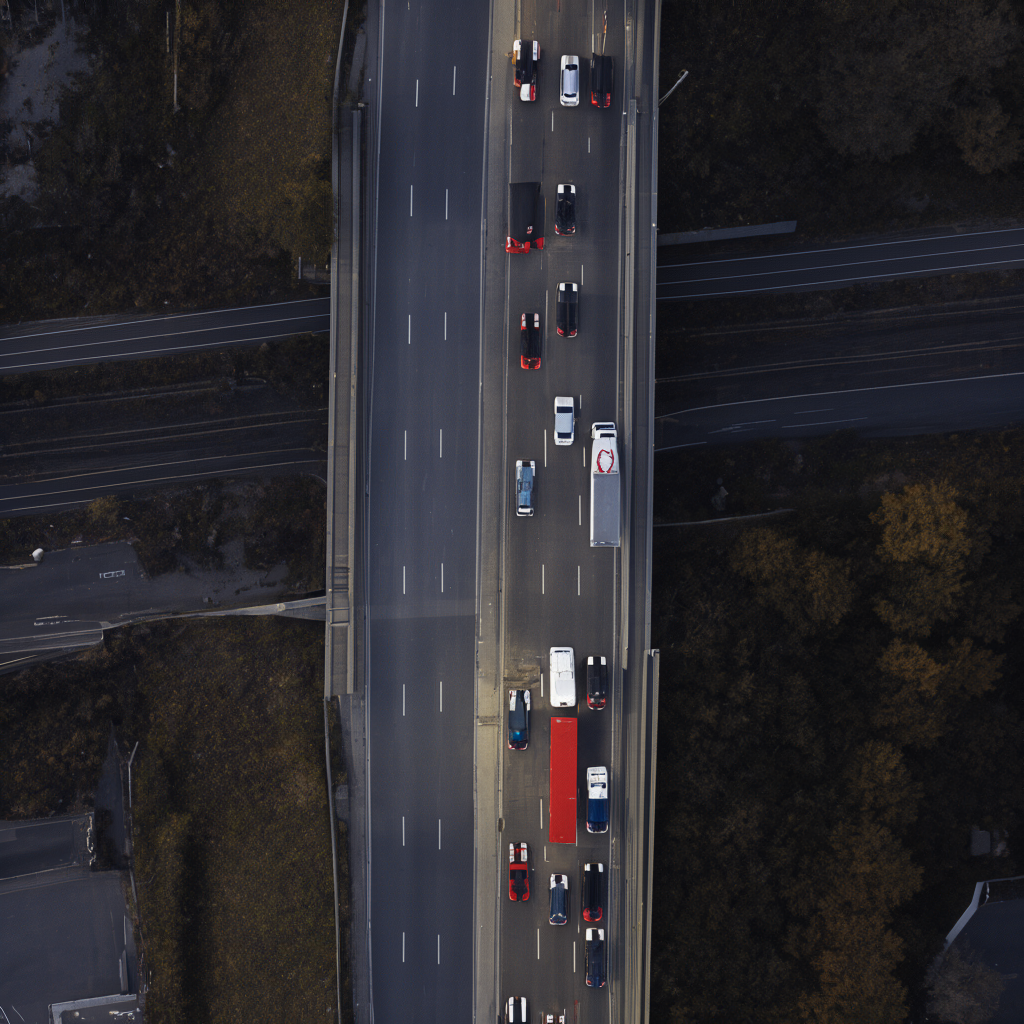}}
    \resizebox{\linewidth}{!}{
    \begin{tabular}[t]{c c}
    & \settowidth{\imagewidth}{\includegraphics{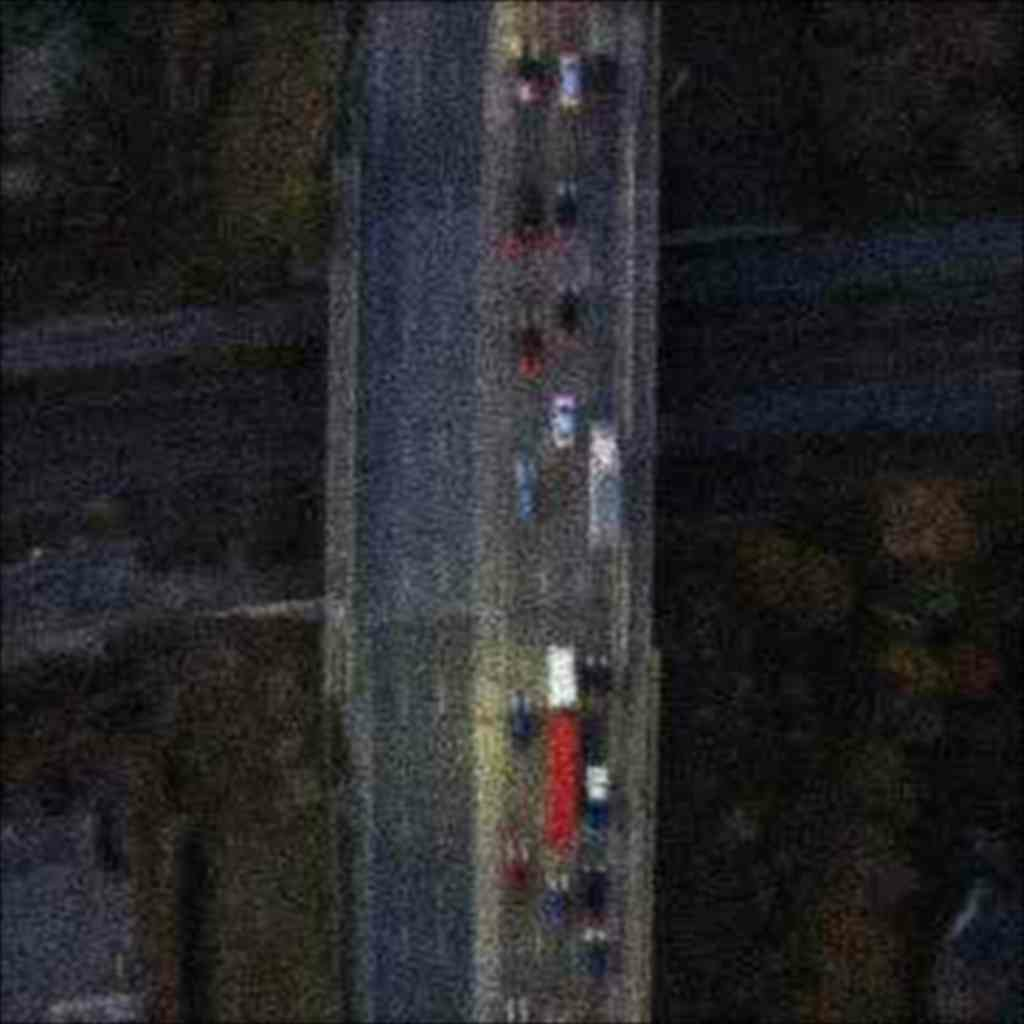}}
    \includegraphics[width=\xwidth, clip=true, trim = 0.1\imagewidth{} 0.25\imagewidth{} 0.45\imagewidth{} 0.65\imagewidth{}]{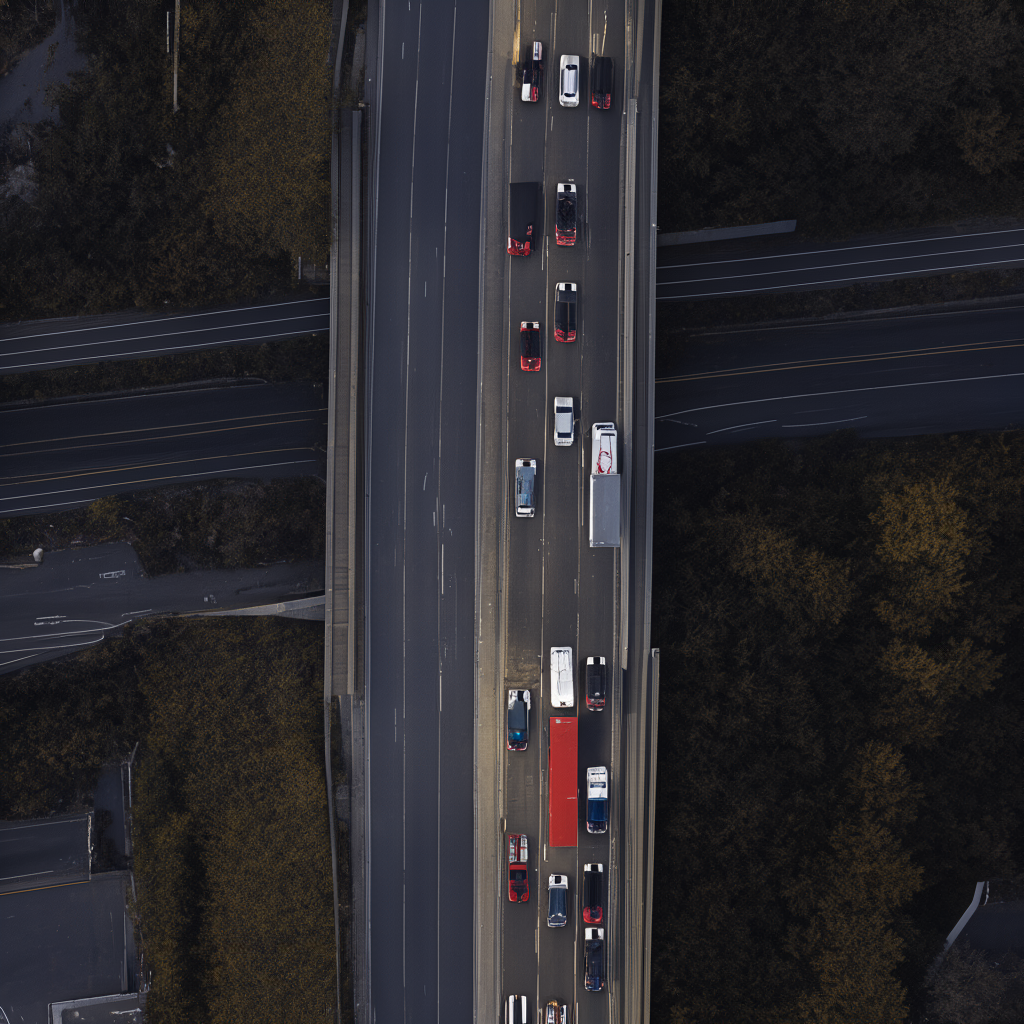} \\
    & \scriptsize \emph{cfg}=2.0 CLIPIQA=0.4481 \\
    & \includegraphics[width=\xwidth, clip=true, trim = 0.1\imagewidth{} 0.25\imagewidth{} 0.45\imagewidth{} 0.65\imagewidth{}]{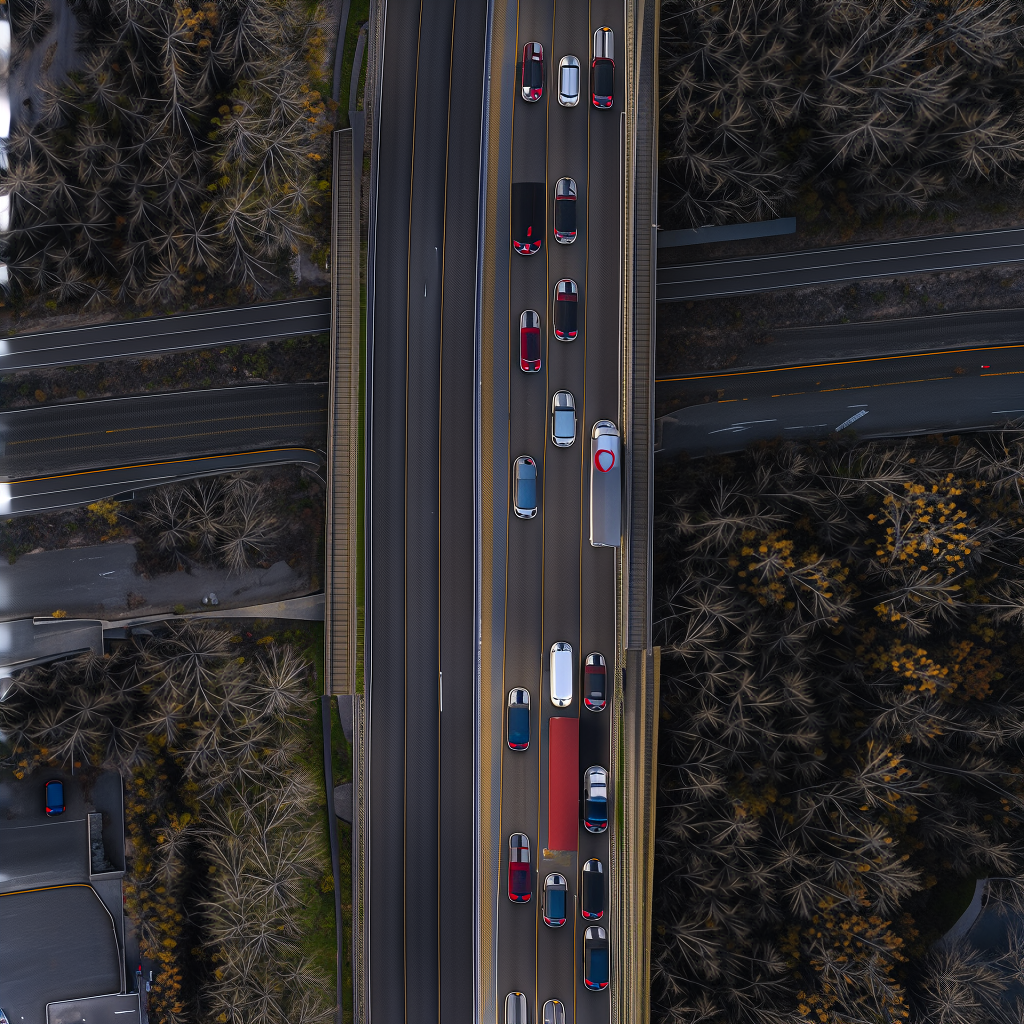} \\
    & \scriptsize \emph{cfg}=14.0 CLIPIQA=0.7091 \\
    & \includegraphics[width=\xwidth, clip=true, trim = 0.1\imagewidth{} 0.25\imagewidth{} 0.45\imagewidth{} 0.65\imagewidth{}]{figures/effect_cfg/mmsr_mmcfg_2_000026.png} \\
    & \scriptsize \emph{$m$-cfg}=2.0 CLIPIQA=0.4122 \\
    \multirow[t]{7}{*}{\redrectangle{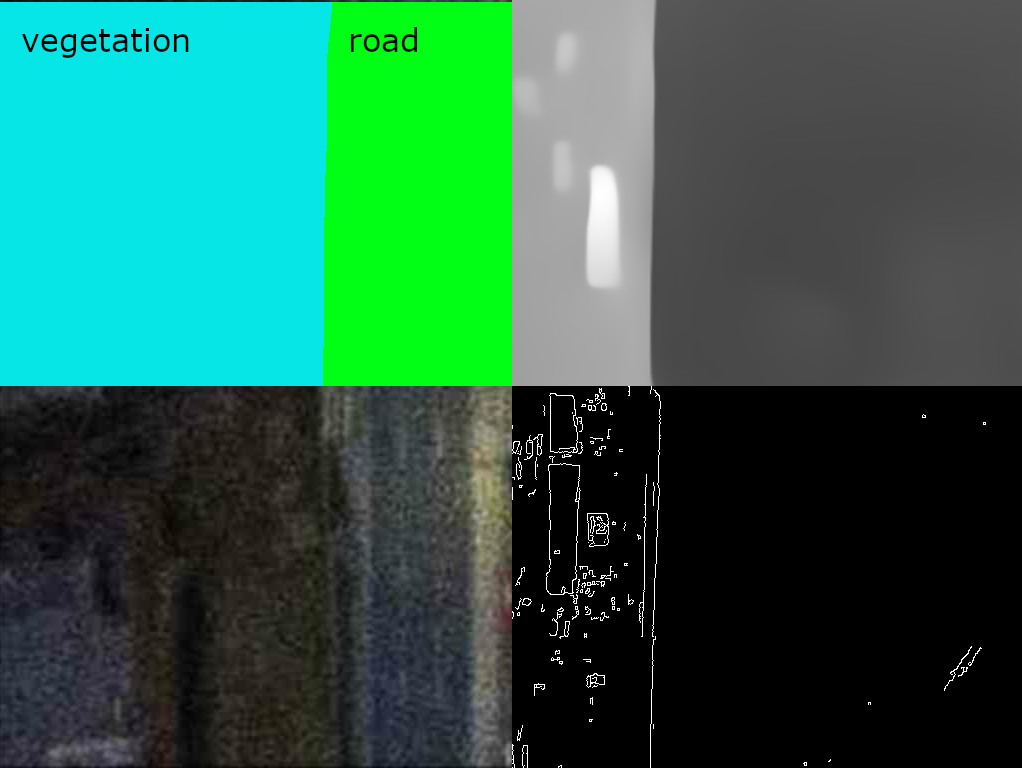}{\xxwidth}{0.2cm}{0.05}{-0.45}}
    & \includegraphics[width=\xwidth, clip=true, trim = 0.1\imagewidth{} 0.25\imagewidth{} 0.45\imagewidth{} 0.65\imagewidth{}]{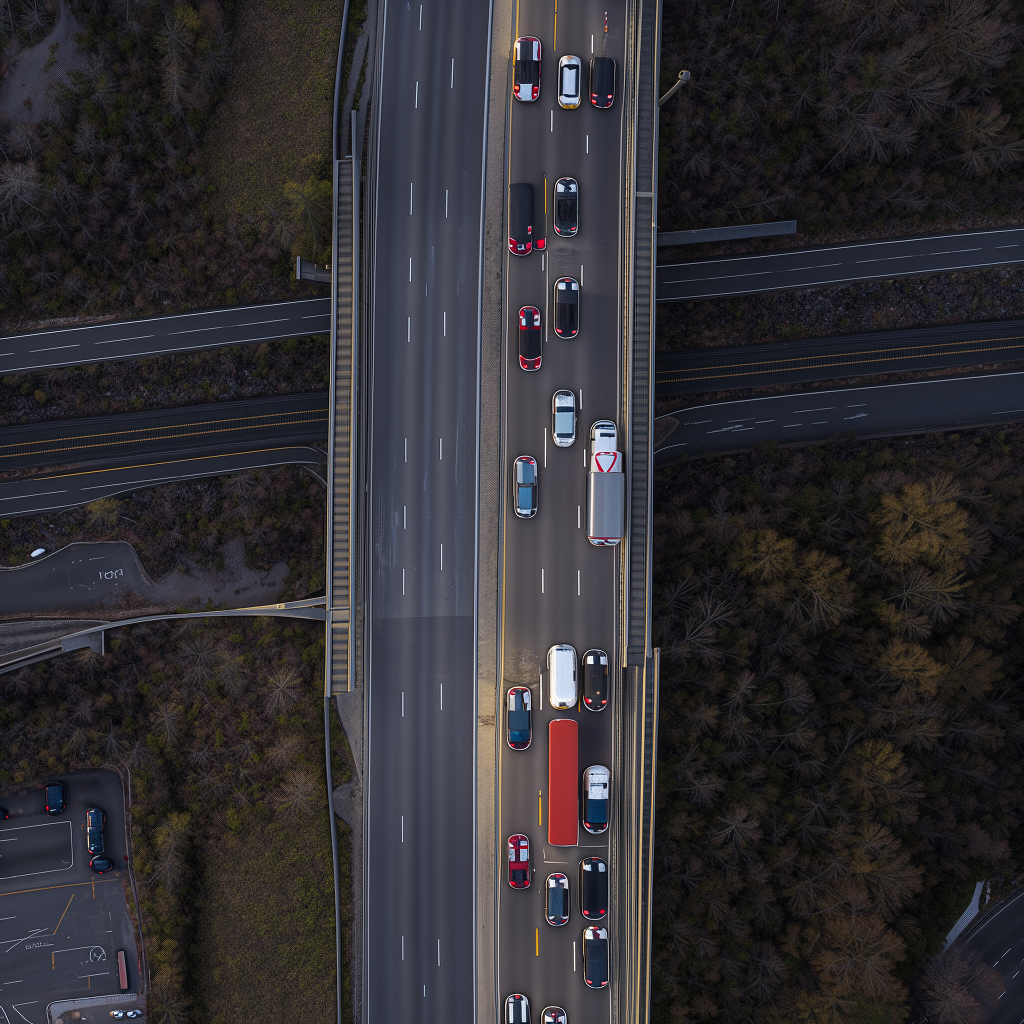} \\
    \scriptsize Segment, Depth, LR, and Edge Input & \scriptsize \emph{$m$-cfg}=14.0 CLIPIQA=0.7352 \\
    \end{tabular}
    }
    \caption{Visual comaprisons}
    \end{subtable}
    \vspace{-1\baselineskip}
    \caption{
    Ablation of guidance on \emph{DIV2K-Val-100} 1024p.
    Our multimodal CFG (\emph{$m$-cfg}) mitigates artifacts often present when using high guidance rates. This leads to an improved balance between visual fidelity and preservation of key identifying features.
    }
    \label{tab:cfg}
    \vspace{-1\baselineskip}
\end{table}

\begin{figure}[ht]
    \centering
    \def\xwidth{0.44\linewidth}
    \cellcolor{tabfirst} \settowidth{\imagewidth}{\includegraphics{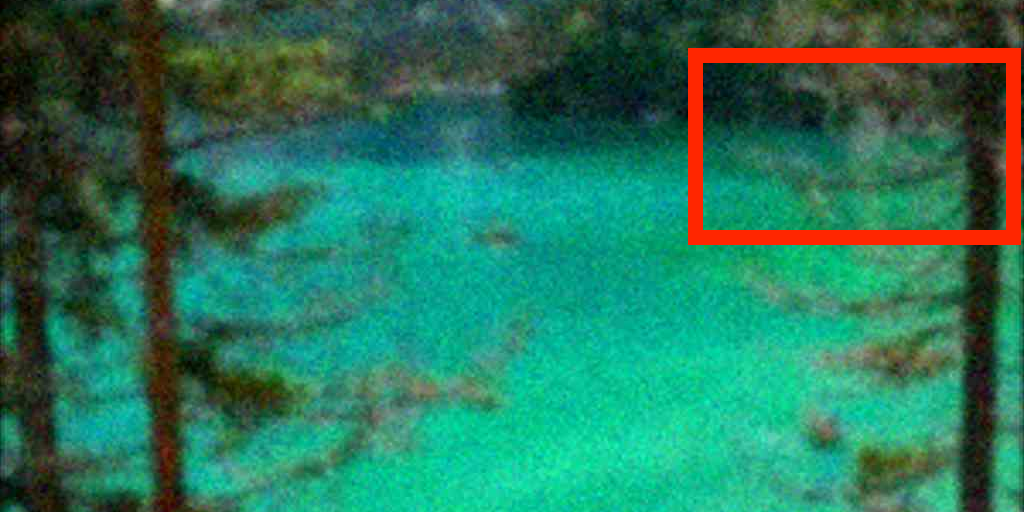}} 
    \setlength{\tabcolsep}{1pt}
    \renewcommand\arraystretch{0.6}
    \resizebox{\linewidth}{!}{
    \begin{tabular}{@{}c c@{}}
    \includegraphics[width=\xwidth, clip=true, trim = 0\imagewidth{} 0.1\imagewidth{} 0\imagewidth{} 0.0\imagewidth{}]{figures/latent_connector_ab/000006_input.png} &
    \settowidth{\imagewidth}{\includegraphics{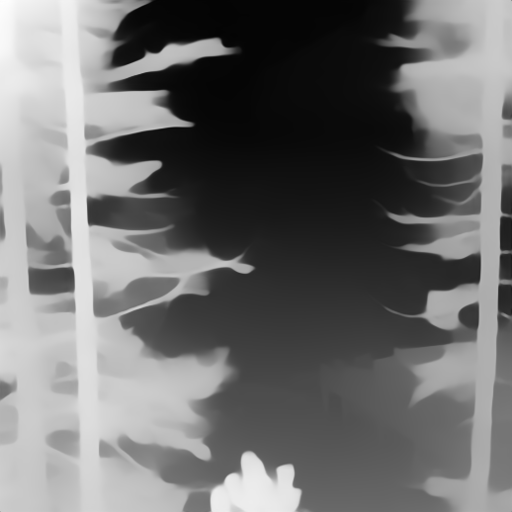}} 
    \includegraphics[width=\xwidth, clip=true, trim = 0.7\imagewidth{} 0.65\imagewidth{} 0\imagewidth{} 0.23\imagewidth{}]{figures/latent_connector_ab/000006_depth.png} \\
    \scriptsize LR Input & \scriptsize Depth Input \\
    \includegraphics[width=\xwidth, clip=true, trim = 0.7\imagewidth{} 0.65\imagewidth{} 0\imagewidth{} 0.23\imagewidth{}]{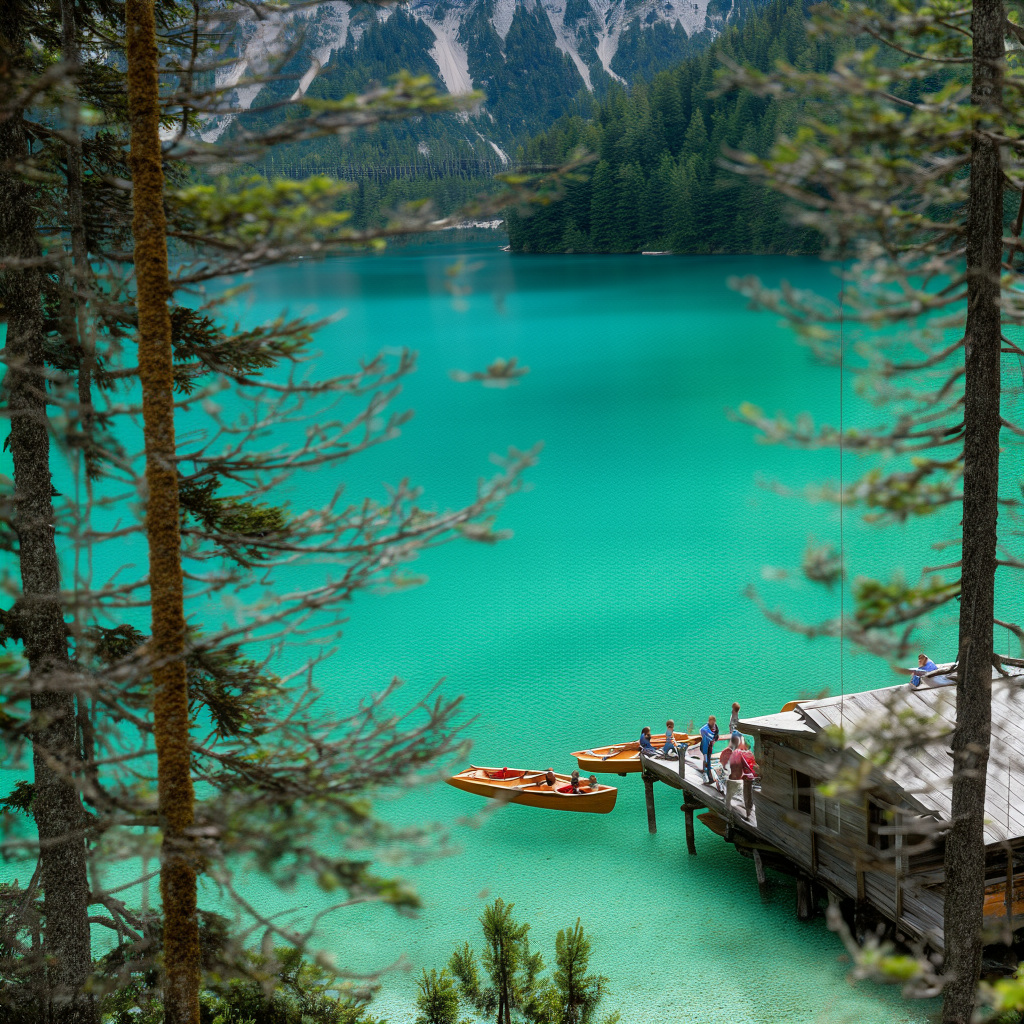} &
    \includegraphics[width=\xwidth, clip=true, trim = 0.7\imagewidth{} 0.65\imagewidth{} 0\imagewidth{} 0.23\imagewidth{}]{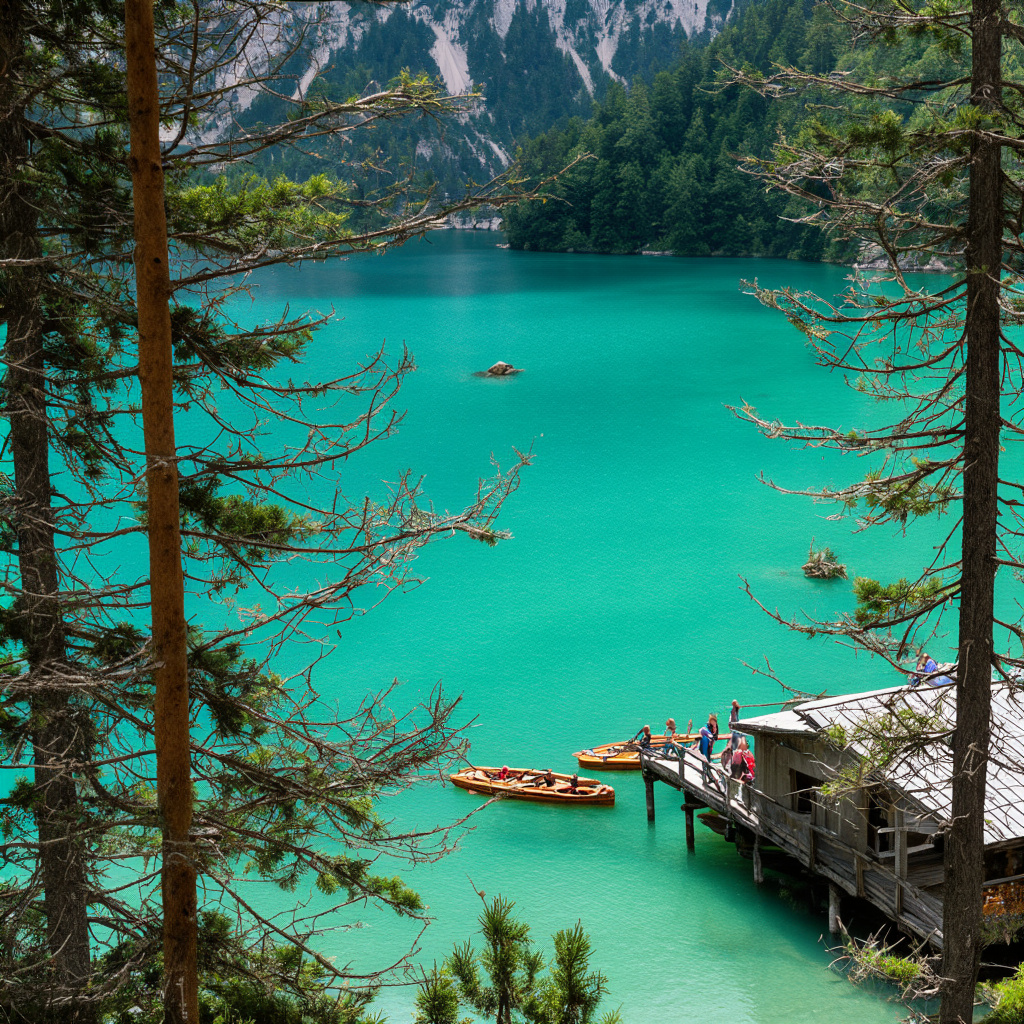} \\
    \scriptsize \emph{w/o. MMLC} Result & \scriptsize \emph{w. MMLC} Result
    \end{tabular}
    }
    \vspace{-.7\baselineskip}
    \caption{Visual results when (not) using the MMLC module.}
    \label{fig:mmlcabresult}
    \vspace{-.5\baselineskip}
\end{figure}

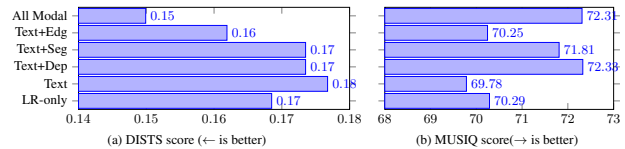
\begin{figure}[t]
\centering
\small
\setlength{\tabcolsep}{1pt}
\resizebox{\linewidth}{!}{
\begin{tabular}{c c}
\begin{tikzpicture}
\begin{axis}[
xbar,
xmin=0.14, xmax=0.18,
symbolic y coords={All Modal, Text+Edg, Text+Seg, Text+Dep, Text, LR-only},
ytick=data,
nodes near coords,
nodes near coords align={horizontal},
y dir=reverse,
bar shift auto,
height=4cm,
width=8cm,
]
\addplot coordinates {
(0.1685,LR-only)
(0.17675175561507542,Text)
(0.17350511318445205,Text+Dep)
(0.17350511318445205,Text+Seg)
(0.16190479417641957,Text+Edg)
(0.14991367663939795,All Modal)
};
\end{axis}
\end{tikzpicture}
&
\begin{tikzpicture}
\begin{axis}[
xbar,
xmin=68, xmax=73,
symbolic y coords={All Modal, Text+Edg, Text+Seg, Text+Dep, Text, LR-only},
ytick=data,
nodes near coords,
nodes near coords align={horizontal},
y dir=reverse,
yticklabels={}, %
bar shift auto,
height=4cm,
width=7cm,
]
\addplot coordinates {
(70.2882,LR-only)
(69.78388242149353,Text)
(72.32654594612121,Text+Dep)
(71.80904658508301,Text+Seg)
(70.2450787709554,Text+Edg)
(72.31,All Modal)
};
\end{axis}
\end{tikzpicture} \\
(a) DISTS score ($\leftarrow$ is better)  & (b) MUSIQ score($\rightarrow$ is better) \\
\end{tabular}
}
\vspace{-.7\baselineskip}
\caption{
Ablation of modalities on \emph{DIV2K-Val-100} 1024p.
Incorporating different modalities enhances various aspects of the super-resolution results compared to the text-guided baseline.}
\label{tab:eachmodal}
\vspace{-1\baselineskip}
\end{figure}

\begin{figure*}[t]
    \centering
    \includegraphics[width=\linewidth]{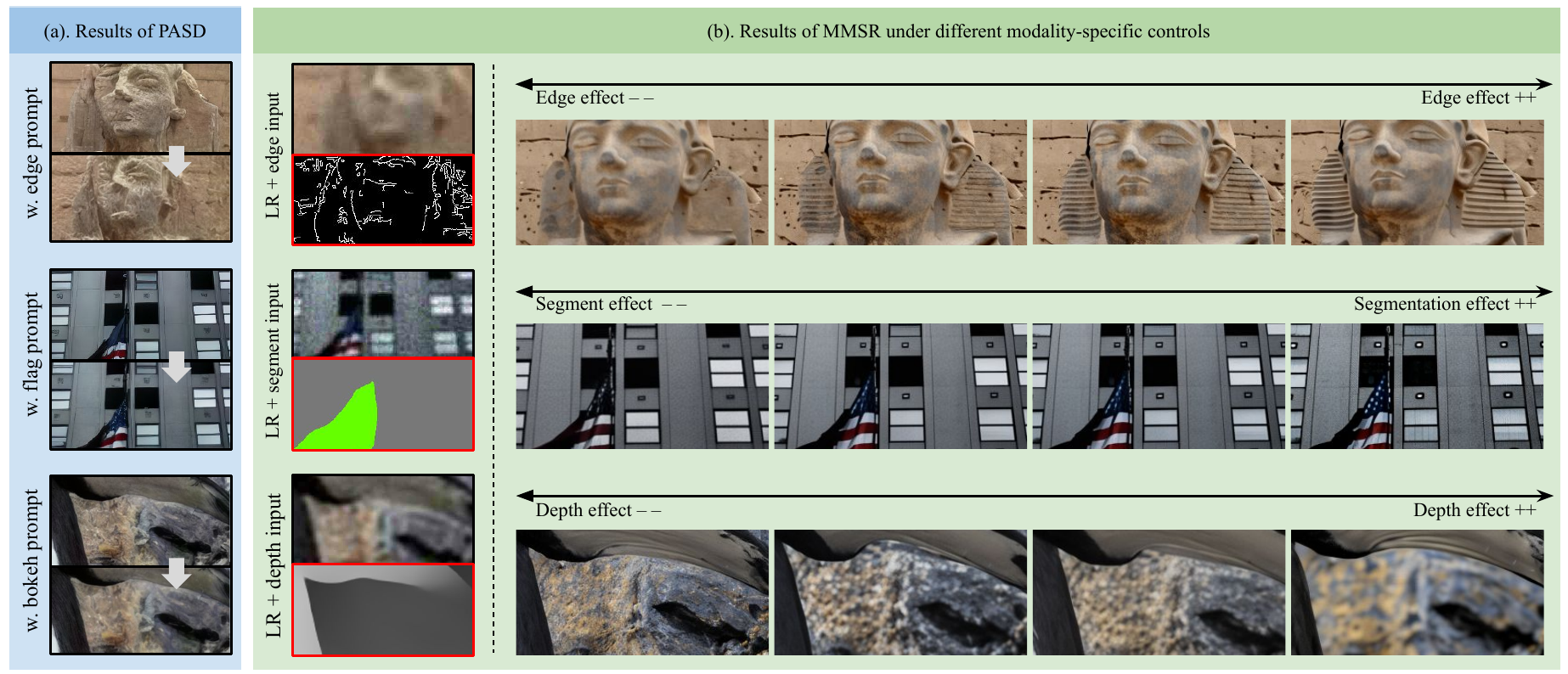}
    \vspace{-1\baselineskip}
    \vspace{-.2em}
    \caption{Our method allows for fine-grained control over super-resolution results by adjusting the influence of each input modality. For example, reducing the edge temperature enhances edge sharpness (first row). Lowering the segmentation temperature emphasizes distinct features, such as the star pattern on the flag (second row). Decreasing the depth temperature accentuates depth-of-field effects, like the bokeh between the foreground and background (third row). In contrast, PASD~\cite{yang2023pixel} exhibits limited control over such fine-grained details.} 

    \label{fig:control}
\end{figure*}

\vspace{.1em}
\noindent \textbf{Contributions of Each Modality.}
Figure~\ref{tab:eachmodal} shows the contributions of each modality by masking out input modalities during testing.
Benefiting from the improved flexibility of using multimodal token $m_\emptyset$, discussed in Sec.~\ref{sec:emptoken}, our method is robust in scenarios with only fewer modalities input during testing.
We would like to note several observations:
(1) Our default multimodal setting achieves the best trade-off  between the non-reference metric MUSIQ and the reference-based metric DISTS.
(2) Depth information primarily enhances perceptual quality (MUSIQ), while other modalities contribute more significantly to preserving identity (DISTS).  
These results highlight the diverse contributions of each modality and emphasize the advantage of our multimodal approach in effectively combining their strengths for optimal super-resolution performance.

\subsection{Qualitative Results}
Figures~\ref{fig:realresults} and \ref{fig:srresults} provide a visual comparison of our method against state-of-the-art approaches, namely SeeSR~\cite{wu2024seesr}, PASD~\cite{yang2023pixel}, and SUPIR~\cite{yu2024scaling}. Our method demonstrates several key advantages.  \emph{Enhanced Realism:} In the first example of Figure~\ref{fig:srresults}, our method produces fewer artifacts and remains more faithful to the high-resolution image compared to other methods. This highlights our ability to maintain realism without introducing spurious details. \emph{Robustness to Challenging Conditions:} The second example in Figure~\ref{fig:srresults}, a long-range shot affected by noise and turbulence, shows our method's ability to generate clear human details where other methods struggle. This is attributed to the effective use of semantic segmentation for accurate human localization, further enriched by the textual caption. These examples illustrate the effectiveness of our multimodal approach in generating high-quality images across diverse scenarios. 
See supplemental material for additional results and analysis.

\subsection{Controllability Comparisons}

While recent works leverage text prompts for controlling super-resolution~\cite{yang2023pixel,wu2024seesr, yu2024scaling}, these often lack fine-grained control and can yield inconsistent results.  Our multimodal approach introduces modality-specific temperature weights to scale attention scores, enabling precise manipulation of SR outputs by amplifying or diminishing the influence of each modality (e.g., depth, segmentation). Figure~\ref{fig:control} contrasts our approach with PASD~\cite{yang2023pixel}, which relies solely on text prompts.
Specifically, decreasing the temperature of the edge modality changes the richness of details in our result, while changing PASD's prompt with ``more edge'' doesn't change their result.
Decreasing the temperature of semantic segmentation map and depth map also lead to manageable changes such as more visible star parttern and stronger bokeh.
In contrast, directly add corresponding prompt in PASD's prompt barely changes its result.
Moreover, our approach exhibits smooth transitions in image characteristics as temperatures are adjusted, providing interpretable control and insights into the role of each modality.

\section{Conclusion}
This work introduces a novel diffusion-based framework for image super-resolution that seamlessly integrates diverse modalities—including text descriptions, depth maps, edges, and segmentation maps—within a single, unified model. By leveraging pretrained text-to-image models, our method achieves enhanced realism and accurate reconstruction, outperforming existing text-guided super-resolution methods both qualitatively and quantitatively.  Furthermore, a learnable multimodal token and modality-specific contribution controls provide fine-grained control over the super-resolution process, enabling adjustable perception-distortion tradeoffs and robust performance even with imperfect or missing modalities in most cases.

\vspace{.2em}
\noindent \textbf{Limitations and Future Work.} While adding multimodal information significantly enhances SR performance, it introduces computational overhead. For instance, using Gemini Flash for image captioning results in a throughput of 0.34 images per second, which is slower than depth at 1.99 img/s, semantic segmentation at 2.09 img/s, and DDIM sampling at 0.54 img/s. However, cross-modal predictions can be parallelized, and our method achieves speeds comparable to other text-driven SR models~\cite{yang2023pixel, yu2024scaling}. Future work will explore optimizing the vision-language component for faster inference and investigate more robust modules for extracting modality-specific information, potentially enhancing performance even with noisy or incomplete inputs.

\section{Acknowledgments}
Vishal M. Patel was supported by NSF CAREER award 2045489.
We are grateful to Keren Ye and Mo Zhou for their valuable feedback. We also extend our gratitude to Shlomi Fruchter, Kevin Murphy, Mohammad Babaeizadeh, and Han Zhang for their instrumental
contributions in facilitating the initial implementation of the latent diffusion models.

{
    \small
    \bibliographystyle{ieeenat_fullname}
    \bibliography{old}
}

\end{document}